\documentclass[]{entalpic}
\usepackage[utf8]{inputenc} %
\usepackage[T1]{fontenc}    %
\usepackage{hyperref}       %
\usepackage{url}            %
\usepackage{booktabs}       %
\usepackage{amsfonts}       %
\usepackage{nicefrac}       %
\usepackage{microtype}      %
\usepackage{xcolor}         %
\usepackage[most]{tcolorbox}
\usepackage{graphicx}
\usepackage{amsmath}
\usepackage{multirow}
\usepackage{multicol}
\usepackage[numbers,super,sort&compress]{natbib}
\usepackage{doi}
\usepackage{subcaption}
\usepackage{tcolorbox}
\usepackage{totcount}
\usepackage{booktabs}
\usepackage{adjustbox}
\usepackage{cleveref}
\usepackage{nameref}
\usepackage{xspace}
\usepackage{bm}
\usepackage{tablefootnote}
\usepackage{footmisc}
\usepackage{placeins}
\DeclareUnicodeCharacter{2212}{-}
\makeatletter
\def\ttl@Hy@steplink#1{%
  \Hy@MakeCurrentHrefAuto{#1*}%
  \edef\ttl@Hy@saveanchor{%
    \noexpand\Hy@raisedlink{%
      \noexpand\hyper@anchorstart{\@currentHref}%
      \noexpand\hyper@anchorend
      \def\noexpand\ttl@Hy@SavedCurrentHref{\@currentHref}%
      \noexpand\ttl@Hy@PatchSaveWrite
    }%
  }%
}
\def\ttl@Hy@PatchSaveWrite{%
  \begingroup
    \toks@\expandafter{\ttl@savewrite}%
    \edef\x{\endgroup
      \def\noexpand\ttl@savewrite{%
        \let\noexpand\@currentHref
            \noexpand\ttl@Hy@SavedCurrentHref
        \the\toks@
      }%
    }%
  \x
}
\def\ttl@Hy@refstepcounter#1{%
  \let\ttl@b\Hy@raisedlink
  \def\Hy@raisedlink##1{%
    \def\ttl@Hy@saveanchor{\Hy@raisedlink{##1}}%
  }%
  \refstepcounter{#1}%
  \let\Hy@raisedlink\ttl@b
}
\makeatother

\usepackage{enumitem}
\setlist{topsep=0pt, partopsep=0pt, itemsep=5pt, parsep=2pt}

\newcommand{\meanstd}[2]{%
  \ensuremath{#1\,{\mathchoice{\scriptstyle \pm\,#2}{\scriptstyle \pm\,#2}{\scriptscriptstyle \pm\,#2}{\scriptscriptstyle \pm\,#2}}}%
}

\newcommand{\lematgen}{\textsc{LeMat-GenBench}\xspace}
\newcommand{\gloss}[1]{#1\hyperlink{box:crystallography}{\textsuperscript{\dag}}}
\newcommand{\secref}[1]{Section ``\nameref{#1}''}

\makeatletter
\renewcommand{\title}[1]{\gdef\@title{#1}\newcommand{\titlelist}{%
  {\LARGE\bfseries\rmfamily\raggedright\linespread{1.1}\selectfont #1\par}%
  \vspace{0.4cm}%
}}
\makeatother

\title{\lematgen: A Unified Evaluation Framework for Crystal Generative Models} 

\author[1,*]{Siddharth~Betala}
\author[1]{Samuel~P.~Gleason}
\author[1]{Ali~Ramlaoui}
\author[2]{Andy~Xu}
\author[3]{Georgia~Channing}
\author[4, 5]{Daniel~Levy}
\author[3]{Clementine~Fourrier}
\author[6]{Nikita~Kazeev}
\author[7]{Chaitanya~K.~Joshi}
\author[4, 5]{Sekou-Oumar~Kaba}
\author[5]{Felix~Therrien}
\author[8, 5]{Alex~Hernandez-Garcia}
\author[9]{Rocío~Mercado}
\author[10]{N.~M.~Anoop~Krishnan}
\author[1,*]{Alexandre~Duval}

\affiliation[1]{Entalpic, Paris, France}
\affiliation[2]{Harvey Mudd College, Claremont, USA}
\affiliation[3]{Hugging Face, Paris, France}
\affiliation[4]{McGill University, Montreal, Canada}
\affiliation[5]{Mila, Montreal, Canada}
\affiliation[6]{National University of Singapore, Singapore}
\affiliation[7]{University of Cambridge, Cambridge, United Kingdom}
\affiliation[8]{Université de Montréal, Montreal, Canada}
\affiliation[9]{Chalmers University of Technology, Gothenburg, Sweden}
\affiliation[10]{Indian Institute of Technology Delhi, New Delhi, India}
\contribution[*]{Corresponding authors}

\abstract{
Generative machine learning (ML) models hold great promise for accelerating materials discovery through the inverse design of inorganic crystals, enabling an unprecedented exploration of chemical space. Yet, the lack of standardized evaluation frameworks makes it challenging to evaluate, compare, and further develop these ML models meaningfully. In this work, we introduce \lematgen, a unified benchmark for generative models of crystalline materials, supported by a set of evaluation metrics designed to better inform model development and downstream applications. We release both an open-source evaluation suite and a public leaderboard on Hugging Face, and benchmark 12 recent generative models. Results reveal that an increase in stability is accompanied by lower novelty and diversity on average, with no model excelling across all dimensions. Altogether, \lematgen establishes a reproducible and extensible foundation for fair model comparison and aims to guide the development of more reliable, discovery-oriented generative models for crystalline materials.
}
\metadata[Code]{\url{https://github.com/LeMaterial/lemat-genbench}}
\metadata[Leaderboard]{\url{https://huggingface.co/spaces/LeMaterial/LeMat-GenBench}}
\correspondence{\email{siddharth.betala@entalpic.ai}, \email{alexandre.duval@entalpic.ai}}

\begin{document}

\maketitle

\section{Introduction}
 
Discovery of inorganic crystalline materials has traditionally relied on an Edisonian cycle of expert intuition and experimental validation \citep{schmidt2019recent}, occasionally aided by quantum simulations such as density functional theory (DFT) \citep{kohn1996density, Sholl2009}. While such quantum simulations provide valuable insights into structure and stability, they remain computationally expensive and require a fully specified atomic configuration (\gloss{crystal lattices} and atomic positions), information that is rarely available when proposing novel materials. Machine learning (ML) models \citep{deringer2019machine, unke2021machine}, particularly those based on geometric graph neural networks \citep{duval2023hitchhiker}, offer fast proxies for DFT and enable scalable evaluation of candidate structures. Yet, similar to DFT, these models are limited to predicting the properties (such as energy and forces) for given atomic structure and are not designed to explore or propose new crystal geometries. This has motivated a growing wave of generative ML models for materials discovery \citep{milaai4science2023crystalgfn, levysymmcd, kazeev2025wyckoff,zeni2025generative}. 

These models, ranging from variational autoencoders (VAEs) \citep{kingma2013autoencodingvariationalbayes} and diffusion models \citep{song2021scorebased} to GFlowNets \citep{bengio2023gflownetfoundations} and large language models (LLMs) \citep{brown2020language}, aim to learn the distribution of valid crystal structures and sample from it, guided by target properties when applicable. This \textit{inverse design} paradigm promises to unlock previously inaccessible regions of chemical space and accelerate the discovery of practically relevant materials.

The rapid development of generative models for crystal structures has revealed a critical challenge: the absence of standardized evaluation protocols. Studies vary widely in how they assess stability, define novelty, or validate structures. For instance, researchers use different reference datasets, fingerprinting methods, energy estimators \citep{batatia2023foundation, unke2021machine}, relaxation procedures, energy thresholds, and more, making the direct comparison of the performance of these models challenging. Without shared benchmarks designed for the fair comparison of generative models, it remains difficult to disentangle genuine model improvements from differences in evaluation design. While standardized benchmarks for predictive modeling of materials, such as Matbench \citep{dunn2020benchmarking}, have enabled progress in property prediction by providing systematic model comparison, no analogous established framework exists for the evaluation of generative models of crystal structures. This gap makes progress difficult to quantify and leaves the community without a shared reference for scientific advancement in data-driven materials design.

\begin{figure}
\centering
\includegraphics[width=\textwidth]{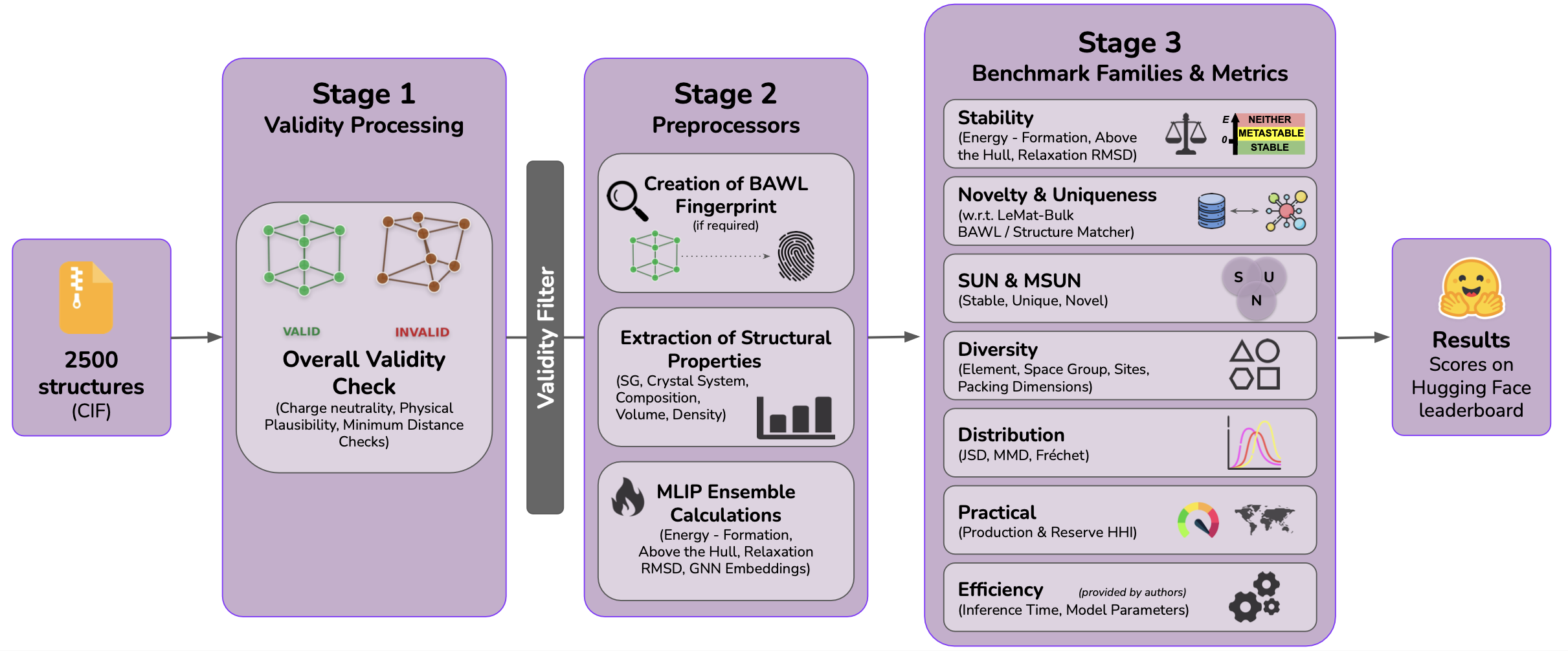}
\caption{\textbf{LeMat-GenBench pipeline from raw model outputs to comprehensive evaluation.} The framework begins by filtering generated crystals through rigorous validity checks. Valid structures are enriched with structural fingerprints, crystallographic descriptors, and MLIP-based energetic properties. Then, LeMat-GenBench computes a unified set of metrics capturing stability, novelty, uniqueness, diversity, distributional alignment, practical synthesizability considerations, and model efficiency. The resulting metric suite provides a standardized, leaderboard-ready assessment of generative models for inorganic crystal design.}
\label{app:fig:pipeline}
\end{figure}

To this end, we introduce \lematgen, a benchmarking framework (\Cref{app:fig:pipeline}) aimed at standardizing the evaluation of generative models for inorganic crystal structures with the following key aspects:
\begin{itemize}
    \item \textbf{Standardized evaluation protocol:} unified metrics suite centered on (Meta)Stable, Unique, Novel ((M.)S.U.N.) rate alongside validity, diversity, and efficiency;
    \item \textbf{Principled evaluation methodology:} carefully designed choices for stability estimation, validity filtering, and reference dataset construction that improve reliability and reduce common sources of error;
    \item \textbf{Open-source toolkit:} a public Python suite for reproducible metric computation and model evaluation;
    \item \textbf{Comprehensive benchmark analysis:} evaluation of 12 contemporary models, revealing clear trade-offs across metrics made available through a leaderboard to establish performances and provide a fillip to the development of better models.
\end{itemize}

By establishing shared protocols and rigorous evaluation standards, \lematgen aims to enable more systematic, fair, and transparent model comparisons. The framework is not set in stone; it is designed to evolve with methodological advancements while supporting both research and practical applications in AI-driven materials.

The field of generative modeling for inorganic crystals has rapidly grown, with an increasing diversity of proposed architectures, training strategies, representations, and target outputs. As the challenge of materials discovery is addressed by an expanding array of methodologies, this heterogeneity complicates the task of fair comparison.
In what follows, we provide an overview of current modeling approaches, followed by an analysis of existing evaluation practices and their limitations. Together, these point to the need for standardized benchmarks and motivate the framework we introduce in this work.

\textbf{\textit{Generative modeling for inorganic crystals}} has emerged as a promising strategy for inverse design, enabling the proposal of candidate structures with desired stability, \gloss{symmetry}, or functional properties. Most of these ML models are trained on large datasets of relaxed crystal structures to generate new candidates, either by sampling from a learned distribution of valid crystals, or conditioned on specific target properties.
Over the past few years, an extensive range of architectural paradigms have been explored. We briefly discuss the prominent families of models below.
A more comprehensive survey of the different approaches to generative modeling crystals can be found in \Cref{app:sec:survey}, with a taxonomy presented in \Cref{app:fig:taxonomy}.

Among the earliest approaches were latent-variable models. \textbf{Variational AutoEncoders} (VAEs)\citep{noh_inverse_2019, hoffmann_data-driven_2019, court_3-d_2020} encode crystal representations into a continuous latent space to enable interpolation and sampling, but struggle with decoding to valid atomic configurations ~\citep{zhao2023physics}. \textbf{GAN}-based methods, on the other hand, attempt to generate crystal representations via adversarial training \citep{kim2020}. 
Despite being widely used years ago, their use has been drastically reduced due to training instabilities, limited diversity, and poor domain adaptation to 3D periodic systems~\citep{zhao2021}.

\textbf{Diffusion models} have become the dominant paradigm, learning to gradually transform Gaussian noise into 3D representations of periodic atomic structures \citep{xie2021crystal, jiao2023crystal, zeni2025generative}.
The diffusion process is able to incorporate geometric and physical inductive biases, resulting in physically plausible crystals. However, they suffer from slow inference times due to the increased number of steps required during the denoising process. \textbf{Flow-based} models offer a related but computationally faster alternative than diffusion, and learn velocity fields that map between a base distribution and the data distribution \citep{miller2024flowmm}.
Recent variants of these models have incorporated explicit symmetry conditioning \citep{jiao2024space, levysymmcd, chang2025space, puny2025space}, textual conditioning from pretrained language models \citep{das2025periodic, park2025exploration, sriram2024flowllm}, and joint training on non-crystal datasets \citep{joshi2025all}.

An alternative paradigm uses \textbf{sequential generation} strategies, where the task of crystal generation is decomposed into a series of discrete actions or tokens.
Autoregressive transformers have been used to learn stepwise generation paths of crystals using tokenizations based on \gloss{Wyckoff positions} \citep{cao2024space, kazeev2025wyckoff}.
\textbf{Large language models} (LLMs) have recently been adapted to crystal generation by tokenizing \gloss{CIF} formats.
They can be fine-tuned for specific tasks, and due to their pretraining, can flexibly leverage textual prompts \citep{gruver2024fine, xu2025plaid, mishra2024foundational} and incorporate multimodal conditioning \citep{johansen2025decifer, moro2025multimodal}.
\textbf{Reinforcement learning} (RL) \citep{zamaraeva2023reinforcement, govindarajan2024learning} and \textbf{Generative Flow Networks} (GFlowNets) \citep{milaai4science2023crystalgfn, cipcigan2024reticulargflownet,podina2025catalyst} decompose generation into stepwise actions guided by a reward. 
They differ from the previously described generative models: rather than training on a dataset of known crystals and sampling from this learned distribution, these methods are instead directly trained to generate samples that optimize a reward, such as crystal stability and validity, or some target property.

Taken together, these methods reflect an increasingly diverse and dynamic modeling landscape for crystal generation.
In designing these methods, researchers have chosen different priorities and tradeoffs, balancing aspects such as physical plausibility, efficiency, diversity, and the ability to condition on properties.
This inherent variability leaves the evaluation and comparison of crystal generative models as a major open challenge---due to the lack of standardized tasks, metrics, and data splits.

\textbf{\textit{Evaluating generative models for crystal structures}} is inherently more complex than supervised tasks such as property prediction, where objective metrics such as Root Mean Square Error (RMSE) or Mean Absolute Error (MAE) are well established. In generative settings, one must assess both structural plausibility (i.e., whether the generated atomic arrangement could exist as a real material), and functional relevance, often in the absence of direct ground-truth targets. As a result, current evaluation protocols vary considerably, hindering systematic comparisons across the literature.

Most studies focus on generating structures that are \emph{valid} and \emph{stable}. Validity is typically defined via heuristic filters such as charge neutrality (requiring that the sum of oxidation states across all atoms equals zero) and interatomic distances (being greater than a target physical value) \citep{xie2021crystal}. Some approaches also measure validity by comparing the space group number distribution of generated structures against reference datasets through distributional distance metrics~\citep{baird2024matbench}. However, these checks vary across works, serve different purposes, either as hard pre-filters or as indicative sanity checks, and are not always consistent with real-world data. For example, a notable number of experimentally observed structures in Materials Project \citep{jain2013materials} entries fail charge neutrality tests due to phenomena such as partial oxidation, which can be challenging to quantify (see \secref{sec:validity-metric-logic}).

Thermodynamic stability is usually evaluated through \gloss{energy above the convex hull} ($E_{\text{hull}}$), which quantifies how far a structure lies above the lowest-energy combination of competing phases \citep{batatia2023foundation}. Studies differ in how they define stability thresholds (some requiring structures on the hull, others allowing up to 0.1 eV/atom above it), in their \gloss{relaxation strategies}, and in their choice of reference datasets. Besides, recent studies often favor Machine Learning Interatomic Potentials (MLIPs) to predict energies, since they scale better than DFT. However, MLIPs tend to incorporate biases related to the distribution of their training datasets \citep{fu2022forces}. To date, few works quantify this uncertainty or reconcile MLIP-based scores with DFT-based convex hulls.

Building upon stability, the \emph{S.U.N.} framework (Stable, Unique, Novel) \citep{zeni2025generative} rewards models for generating thermodynamically viable structures (Stable) that are both distinct from one another (Unique) and absent from known databases (Novel). This metric has emerged as a composite diagnostic, but lacks a consensus implementation. Uniqueness and novelty rely on structure-based matching, using tools such as \texttt{StructureMatcher} \citep{ong2013python} or BAWL \citep{siron2025lemat} which compare atomic arrangements under symmetry operations to determine structure equivalence, whose results depend on the reference set, tolerance thresholds, and whether novelty is computed over all samples or only the stable ones. Most works rely on a single scheme without systematically comparing alternatives or handling disorder (e.g., partial occupancies or mixed species) \citep{siron2025lemat} affecting the relevance of S.U.N. scores.

Additional metrics like diversity and distribution similarity are frequently used. Maximum Mean Discrepancy (MMD) \citep{gretton2012kernel}compares whether samples come from the same distribution by comparing their averaged representations. Earth Mover's Distance (EMD) \citep{rubner2000earth} computes the minimum displacement, under a chosen metric, needed to transform one distribution into another. Distribution similarity, often measured by MMD or EMD, often rewards memorization rather than novelty, misaligned with discovery goals.
On the other hand, diversity lacks a unified definition, implementation, and visualization across studies. Lastly, efficiency metrics like training time, memory usage, and inference cost are under-reported, though increasingly important as generation receives more attention and model sizes grow. 

\textit{matbench-genmetrics} \citep{baird2024matbench} is the closest existing precursor to a standardized evaluation framework for crystal generative models, providing an open-source Python library with four metrics -- validity (via structural fingerprints and space group distributions), uniqueness, novelty, and coverage (a recall/precision-like measure relative to a held-out test set) -- evaluated on time-based splits of the Materials Project (MP). While it represents an important step toward reproducible evaluation, it does not perform thermodynamic stability assessment via energy calculations or convex hull analysis, does not report efficiency or diversity metrics, and relies exclusively on MP as reference, which as we show (\secref{sec:lemtavsmp}), can substantially overestimate novelty and stability relative to a more comprehensive reference. 

A related benchmarking effort by Szymanski et al.\cite{szymanski2025establishing} compares four generative models against non-ML baselines, including random enumeration and data-driven ion exchange \citep{wang2021predicting}, using DFT stability and the Materials Project as reference. While valuable, this study evaluates only four models, relies on a mixed MLIP-DFT hull construction, and does not provide a community leaderboard. We include both the random enumeration baseline and the Wang et al.\citep{wang2021predicting} ion-substitution dataset as baselines evaluated under \lematgen, discussed further in \secref{subsubsec:non-ml}.

To address this fragmentation, we introduce \lematgen: a unified, extensible benchmark that proposes a list of standardized metrics and tasks, enabling reproducible model comparison through public tools and a live leaderboard.

\section{Results}
\subsection{Evaluation Tasks}
Crystal generative modeling spans a range of practical use cases, from learning broad structural priors to targeting application-specific properties. To reflect these different levels of supervision, we conceptually define three evaluation scenarios that form the conceptual landscape in which generative models for materials discovery operate: (i) \emph{unconditional generation}, which assesses a model’s ability to sample realistic crystalline materials without guidance; (ii) \emph{conditional generation}, where models are steered toward property-aware design using inexpensive oracle evaluations; and (iii) \emph{conditional generation under limited budget}, which reflects settings where each oracle call (e.g., DFT or lab experiment) is costly and sample efficiency becomes paramount. Note that an \textit{oracle} is a function that returns a material property score. It can be an ML model such as an MLIP \citep{jacobs2025practical}, a DFT simulation, a lab experiment, or any other evaluation method. Oracles vary in computational cost and accuracy.

In this work, we focus exclusively on the unconditional generation setting. This choice reflects both the maturity of unconditional modeling techniques and the need for standardized metrics to evaluate general-purpose crystal generators before they are deployed or adapted for downstream, property-driven tasks. Establishing rigorous and comparable unconditional metrics is a prerequisite to meaningfully assessing conditional or constrained discovery workflows. We therefore introduce a unified evaluation suite centered on validity, stability, novelty, uniqueness, and diversity, along with standardized reference datasets and MLIP-based energy evaluation protocols. The two conditional settings remain important for real-world discovery, which will be pursued as future extensions of \lematgen.

\subsection{Unconditional Generation Metrics}
\label{subsec:unconditional-gen-metrics}
To evaluate unconditional generation, a representative number of structures (e.g., 2,500) are sampled from the trained model and evaluated using the following metrics, whose formal definitions are provided in \Cref{app:sec:eval-metrics}. 

We define a new \textbf{Validity} metric, building on Xie et al.\cite{xie2021crystal}. It serves as a pre-filtering step to exclude malformed or nonphysical structures, catching common failure modes and reducing computational overhead. These checks include \gloss{CIF} readability, minimum interatomic distances ($> 0.7$ \AA), and density bounds (mass density of $0.01$--$25$ g/cm$^3$ and atomic density of $10^{-5}$--$0.5$ atoms/\AA$^3$ ). We also check for reasonable lattice parameters ($a$, $b$, $c$ within $1$--$100$ \AA~ and $0<\alpha, \beta, \gamma<180$ degrees), a crystal symmetry adhering to a space group determinable by the SpacegroupAnalyzer module in \texttt{pymatgen} \cite{jain2013materials}, and charge neutrality (via oxidation state plausibility analysis, with a tolerance threshold). All results are aggregated into a single validity score: the percentage of structures passing the validity test. Failing any of the above metrics results in an invalid label, and these structures are discarded. While submitters may pre-filter their structures, the ensuing standardized criteria ensure consistency across evaluations. An analysis of our proposed validity metric can be found in \secref{sec:validity-metric-logic}.

Thermodynamic \textbf{Stability} is a key proxy for real-world material feasibility. While formation energy is sometimes used to infer stability, it introduces notable biases, such as favoring strongly bonded, electronegative compositions, and is therefore an unreliable proxy for stability. We thus utilize \emph{energy above the convex hull} and adopt disjoint categories: structures with $E_{\text{hull}} \leq 0$ are \emph{stable}, those with $0 < E_{\text{hull}} \leq 0.1$~eV/atom are \emph{metastable}. We use the term \emph{(meta)stable} to refer to the union of both these sets, aligning with the inclusive threshold common in ML workflows. We note that this range of $E_{\text{hull}}$ values is a thermodynamic heuristic and does not guarantee true metastability. Strict metastability is defined by a structure's occupation of a local minimum on the potential energy surface. Structures within this $E_{\text{hull}}$ range may instead occupy saddle points, exhibiting imaginary phonon modes and, therefore, dynamic instability despite meeting the thermodynamic criteria \citep{togo2015first, sun2016thermodynamic}.

A key methodological detail is how the convex hull is constructed. Mixing MLIP-predicted total energies with a DFT-based convex hull leads to systematic inconsistencies because MLIPs and DFT operate on different energy references depending on the MLIP's training dataset.  As demonstrated empirically (see \secref{sec:self-consistent-hull}), such mixed-reference hulls can degrade stability prediction accuracy when de-referencing energies to make them comparable. For this reason, \lematgen adopts a self-consistent MLIP-based convex hull, where each MLIP both evaluates candidate structures and defines the reference convex hull. This approach cancels potential-specific errors and yields more reliable $E_{\text{hull}}$ estimates, particularly under strict stability thresholds.

The convex hull is constructed from LeMat-Bulk~\citep{siron2025lemat}, which contains approximately 5 million structures and provides the broadest coverage of competing phases relaxed under consistent DFT settings. We choose LeMat-Bulk as the reference because its breadth yields a tighter and a more realistic convex hull for stability assessment. This better reflects the needs of end-users seeking genuinely novel materials, rather than reproducing a narrow training distribution. The rationale for choosing LeMat-Bulk over datasets like MP-20 is expanded in \secref{sec:lemtavsmp}. To enable scalable and open evaluation, we compute $E_{\text{hull}}$ using an ensemble of MLIPs: MACE-MP~\citep{batatia2023foundation}, UMA~\citep{wood2025family}, and Orb-v3~\citep{rhodes2025orb}. For each MLIP, we construct its own convex hull and report the mean and standard deviation of $E_{\text{hull}}$ across models, providing robustness and uncertainty estimates. To avoid expensive on-the-fly computation, we pre-compute MLIP energies for all LeMat-Bulk structures; at evaluation time, we simply filter to the relevant compositional subspace and construct the phase diagram for each generated candidate. The precomputed energies are linked, and the exact model checkpoints used throughout this work are listed in \Cref{app:sec:mlip-checkpoints}.

We additionally assess structural quality via a relaxation check: each structure is relaxed with every MLIP in the ensemble, and we compute the root mean square deviation (RMSD) between the initial and relaxed atomic positions (averaged over each MLIP). Low RMSD indicates that the submitted structure already lies near a local energy minimum. We report both \textit{pre-relaxation} scores (computed on submitted structures) and \textit{post-relaxation} scores (computed after MLIP relaxation); full results for both settings appear in \Cref{sec:app:benchmark_results}. For the public leaderboard, we favor pre-relaxation evaluation: users are encouraged to submit structures that have already been relaxed as part of their generation pipeline, ensuring that stability metrics reflect the model's actual outputs rather than post-hoc refinement, therefore improving comparability across submissions.

\textbf{Novelty} measures the fraction of generated crystals not found in a reference dataset, using structural fingerprints, to identify previously unseen materials. Again, we use LeMat-Bulk~\citep{siron2025lemat} as the reference database for novelty evaluation: the same 5 million structures that define competing phases for the convex hull also define what counts as ``known'' for novelty assessment. To assess if two 3D structures point to the same material, we apply pymatgen's \texttt{StructureMatcher} \citep{ong2013python}. To ensure scalability, we restrict comparisons to structures of matching composition. For workflows requiring comparison across large numbers of model outputs, we also implement the Short-BAWL fingerprint \citep{siron2025lemat} as a computationally efficient alternative within the codebase. Note that novelty should be treated with caution. Structural difference does not always imply functional difference, and the boundary between truly novel and incrementally different materials is inherently fuzzy (see \Cref{fig:frame_breakdown} and \Cref{fig:nu_breakdown}). Still, standardizing fingerprint methods, reference datasets, and thresholds is critical for enabling meaningful comparisons and tracking progress. 

\textbf{Uniqueness} quantifies non-redundancy among generated structures using the same fingerprinting approach as novelty, yielding the percentage of unique structures. Importantly, we apply validity, uniqueness, and novelty sequentially, i.e., we also compute novelty on generated structures that are valid and unique, which offers different information compared to doing it on the whole set of structures. 

Through \textbf{(M.)S.U.N. rate}, we aggregate core evaluation criteria into a single standardized metric following the S.U.N. framework introduced by MatterGen \citep{zeni2025generative}: the S.U.N. rate measures the fraction of generated structures that are \underline{S}table, \underline{U}nique, and \underline{N}ovel, computed sequentially from the submitted set in a funnel-like manner. This serves as our primary benchmark for unconditional generation in \lematgen, with each component precisely defined using fixed thresholds, reference datasets, and fingerprinting methods. To account for synthesizable but metastable materials, we also report the M.S.U.N. rate, relaxing stability to $0 < E_{\text{hull}} \leq 0.1$ eV/atom. 
(M.)S.U.N. combines both and thus sets a practical upper bound on generative performance and provides a consistent, interpretable measure to compare models.

\textbf{Diversity metrics} capture the spread of generated samples across lattice parameters, space groups, elemental compositions, and structural descriptors. We compute the per-feature Shannon entropy and provide options for visualization tools in the \lematgen codebase. 

\textbf{Distribution metrics} compare the distribution of generated structures against reference datasets across key crystallographic and compositional attributes. We elaborate further on the diversity and distribution-based metrics offered as part of \lematgen in \Cref{app:sec:eval-metrics}. 

Beyond output quality, we also report \textbf{Model Efficiency} metrics related to training and inference. Specifically, authors must provide: (i) training compute in FLOPs and time (CPU/GPU days), (ii) inference time to generate 2500 structures on a reference CPU/GPU (e.g., Nvidia A100), and (iii) peak memory usage during inference, together with number of model parameters. Reporting these values will inform model cost and scalability, and support analysis of tradeoffs between performance (e.g., (M.)S.U.N. rate) and deployment feasibility. Please note that while inference cost is typically negligible compared to the cost of downstream validation (e.g., DFT or experiments), standardized reporting of these metrics provides important context for practical usability and future integration into closed-loop discovery pipelines.

\textbf{Herfindahl-Hirschman index (HHI)} quantifies supply risk concentration for materials by measuring the concentration of element production sources and reserves. A high value reflects higher risk. More details on how it is computed are in \Cref{app:sec:eval-metrics}.

\subsection{Benchmarking Results}
We evaluate 12 generative models for crystalline materials on \lematgen. The models considered are: ADiT \citep{joshi2025all}, Crystal-GFN \citep{ai4science2023crystal}, CrystalFormer \citep{cao2024space}, 
DiffCSP \citep{jiao2023crystal}, DiffCSP++ \citep{jiao2024space}, LLaMat2-CIF, LLaMat3-CIF 
\citep{mishra2024foundational}, MatterGen \citep{zeni2025generative}, PLaID++ 
\citep{xu2025plaid++}, SymmCD \citep{levysymmcd}, WyFormer \citep{kazeev2025wyckoff}, 
and WyFormer\textendash DFT \citep{kazeev2025wyckoff}. All models were trained on the MP\textendash20 dataset. 
These span a broad range of architectures: diffusion models 
(DiffCSP, DiffCSP++, MatterGen, SymmCD), autoregressive transformers (CrystalFormer, WyFormer, WyFormer-DFT), LLM-based approaches (LLaMat2-CIF, LLaMat3-CIF), flow matching with VAE (ADiT), RL-guided models (PLaID++) and GFlowNets (Crystal-GFN). For each model, we evaluate 2500 generated structures obtained either directly from the authors or from public repositories 
\citep{kazeev2025wyckoff}. 

The choice of 2,500 structures per model reflects both statistical and computational considerations. Since metrics are computed on progressively smaller subsets after validity filtering and the sequential S.U.N. pipeline, starting from fewer structures would leave downstream metrics estimated on uncomfortably small samples for models with lower validity rates, increasing variance, and reducing the reliability of comparisons. Larger sets (e.g., 10,000) are computationally prohibitive given that each structure undergoes validity checks, structural fingerprinting, and energy evaluation by an ensemble of three MLIPs across all submitted models. CrystalFormer is the only exception, for which only 1000 structures were available and hence, the corresponding metrics should be interpreted with this smaller sample size in mind. 

A detailed listing of all data sources is provided in \Cref{table:source-of-data}. The percentages of novel, unique, and stable structures are calculated over all submitted structures but only count valid structures, effectively setting an upper bound on achievable performance equal to the validity rate. This prevents inflated scores for models with low validity (e.g., novelty appearing high simply because the valid set is small) and ensures the reported values reflect the true frequency of valid-and-novel (or stable, unique, etc.) structures in the model’s full output distribution.

To provide reference points for interpreting generative model performance, we establish baselines by directly sampling structures from existing materials databases. These baselines are not intended as fair comparisons, but rather to establish upper bounds on certain metrics (validity, stability) and lower bounds on others (novelty). They help contextualize what performance levels are achievable when simply reproducing known structures versus genuinely discovering new ones. We sample structures from MP-Exp (MP-20 entries marked as experimental by setting the `Theoretical' indicator to false), AFLOW \citep{curtarolo2012aflow}, Alexandria \citep{schmidt2021crystal, schmidt2024improving}, and OQMD \citep{kirklin2015open, saal2013materials}. For each database, we first randomly sample 10,000 structures without replacement (i.e., each structure is drawn at most once, with no structural modifications or substitutions), then draw 2,500 samples with replacement from this subset (i.e., allowing the same structure to be selected multiple times). This two-step procedure emulates the sampling behavior of generative models, which naturally produce duplicates when sampling from a learned distribution. Sampling with replacement ensures that uniqueness metrics remain meaningful---without it, uniqueness would trivially be 100\%, unlike real generative models.

To ensure evaluations remain consistent, scalable, and physically meaningful, all metrics are computed through a simple sequential pipeline: we first apply a standardized validity check (CIF readability, density bounds, charge neutrality, minimum distances, symmetry consistency), and then compute stability, fingerprint-based, distributional, and diversity metrics on the valid subset of generated structures. This guarantees that all downstream quantities are computed on physically plausible samples and that heavy computations---such as MLIP relaxation or structural matching---remain tractable. Importantly, all evaluations are conducted on the submitted structures \textit{without re-relaxation}. While this simplifies benchmarking and ensures reproducibility, it likely underestimates metrics such as S.U.N. and M.S.U.N. We encourage future submissions to include relaxed structures to better reflect thermodynamic viability. 
More detailed results can be found in \secref{sec:lemtavsmp} and \Cref{sec:app:benchmark_results}, where we report results of using MP-20 as reference set (\Cref{app:subsec:mp20_results}) and of post-relaxation metrics (\Cref{tab:uma_bulk_postrelax}, \Cref{tab:table_10}, \Cref{tab:table_15}, \Cref{tab:table_16}).  

\begin{table}[htbp]
\centering
\caption{Core model evaluation metrics calculated using \textbf{MLIP ensemble} with \textbf{LeMat-Bulk} as the reference set. Submissions are organized into four categories separated by horizontal lines: (1) models that incorporate relaxation as part of their generation pipeline (MatterGen, PLaID++, WyFormer, WyFormer-DFT), (2) other generative models, (3) samples from real-world datasets, (4) pre-relaxed samples from non-ML generative baselines. Best values are shown in \textbf{bold}, second-best values are \underline{underlined}. Arrows indicate whether higher ({\textuparrow}) or lower ({\textdownarrow}) values are better. The energy-based metrics are computed over the full generated sets, which pass the initial validity check, and the relatively high standard deviations reflect the stochasticity in the structural generation process rather than measurement error.}
\label{table:core-benchmark-metrics}
\scriptsize
\begin{adjustbox}{max width=\textwidth}
\begin{tabular}{l c cc ccc cc cc}
\toprule
\textbf{Model} &
\textbf{Valid\% {\textuparrow}} &
\textbf{Unique\% {\textuparrow}} & \textbf{Novel\% {\textuparrow}} &
\multicolumn{3}{c}{\textbf{Energy-based}} &
\multicolumn{2}{c}{\textbf{Stability}} &
\multicolumn{2}{c}{\textbf{Metastability}} \\
\cmidrule(lr){5-7} \cmidrule(lr){8-9} \cmidrule(lr){10-11}
& & & & \textbf{\boldmath $E_{f}$} {\tiny \textbf{(Std) {\textdownarrow}}} & \textbf{\boldmath $E_{\mathrm{hull}}$} {\tiny \textbf{(Std) {\textdownarrow}}} & \textbf{RMSD \tiny{(Std)} {\textdownarrow}}
& \textbf{Stable\% {\textuparrow}} & \textbf{SUN\% {\textuparrow}}
& \textbf{Metastable\% {\textuparrow}} & \textbf{MSUN\% {\textuparrow}} \\
\midrule
MatterGen & \underline{95.7} & \textbf{95.1} & \textbf{70.5} & \meanstd{-0.71}{0.79} & \underline{\meanstd{0.17}{0.17}} & \meanstd{0.39}{0.50} & 0.5 & 0.1 {\scriptsize (2)} & 36.0 & \textbf{15.8} \\
PLaID++ & \textbf{96.0} & 77.8 & 24.2 & \meanstd{-0.50}{0.44} & $\bm{\meanstd{0.08}{0.16}}$ & $\bm{\meanstd{0.13}{0.29}}$ & \textbf{6.4} & \textbf{0.9 {\scriptsize (22)}} & \textbf{67.6} & 7.8 \\
WyFormer & 93.4 & 93.0 & \underline{66.4} & \meanstd{-0.45}{0.95} & \meanstd{0.48}{0.50} & \meanstd{0.81}{0.98} & 0.2 & 0.0 {\scriptsize (1)} & 16.0 & 2.2 \\
WyFormer-DFT & 95.2 & \underline{95.0} & \underline{66.4} & \meanstd{-0.67}{0.90} & \meanstd{0.26}{0.35} & \meanstd{0.42}{0.60} & \underline{1.5} & \underline{0.3} {\scriptsize (7)} & 27.6 & 8.3 \\
\midrule
ADiT & 90.6 & 87.8 & 26.0 & \underline{\meanstd{-0.74}{0.93}} & \meanstd{0.33}{0.44} & \underline{\meanstd{0.38}{0.40}} & 0.1 & 0.0 {\scriptsize (0)} & \underline{37.1} & 1.0 \\
Crystal-GFN & 51.7 & 51.7 & 51.7 & $\bm{\meanstd{-1.34}{2.24}}$ & \meanstd{2.05}{1.94} & \meanstd{1.87}{0.97} & 0.0 & 0.0 {\scriptsize (0)} & 0.0 & 0.0 \\
Crystalformer & 69.9 & 69.4 & 31.8 & \meanstd{-0.18}{1.47} & \meanstd{0.69}{1.31} & \meanstd{0.66}{1.05} & 0.9 & 0.0 {\scriptsize (0)} & 29.3 & 3.2 \\
DiffCSP & \underline{95.7} & 94.8 & 66.2 & \meanstd{-0.65}{0.97} & \meanstd{0.26}{0.56} & \meanstd{0.59}{0.63} & 1.1 & 0.0 {\scriptsize (1)} & 32.2 & \underline{9.3} \\
DiffCSP++ & 95.3 & \textbf{95.1} & 62.0 & \meanstd{-0.53}{0.87} & \meanstd{0.40}{0.43} & \meanstd{0.69}{0.82} & 0.7 & 0.2 {\scriptsize (4)} & 27.6 & 5.3 \\
LLaMat2-CIF & 84.4 & 81.4 & 30.0 & \meanstd{-0.48}{1.00} & \meanstd{0.43}{0.70} & \meanstd{0.54}{0.71} & 0.4 & 0.1 {\scriptsize (3)} & 35.7 & 2.1 \\
LLaMat3-CIF & 15.4 & 15.2 & 10.5 & \meanstd{0.69}{2.61} & \meanstd{1.65}{2.39} & \meanstd{1.01}{1.10} & 0.0 & 0.0 {\scriptsize (0)} & 2.3 & 0.2 \\
SymmCD & 73.4 & 73.0 & 47.0 & \meanstd{-0.04}{1.21} & \meanstd{0.86}{1.05} & \meanstd{0.87}{1.09} & 0.5 & 0.0 {\scriptsize (1)} & 19.4 & 2.5 \\
\midrule
Alexandria & 93.4 & 93.4 & 0.0 & \meanstd{-0.19}{0.94} & \meanstd{0.43}{0.61} & \meanstd{0.14}{0.32} & 0.6 & 0.0 {\scriptsize (0)} & 29.2 & 0.0 \\
OQMD & 96.8 & 96.4 & 0.0 & \meanstd{-0.24}{0.88} & \meanstd{0.39}{0.45} & \meanstd{0.14}{0.28} & 1.1 & 0.0 {\scriptsize (0)} & 33.4 & 0.0 \\
AFLOW & 91.4 & 87.1 & 21.5 & \meanstd{-0.19}{0.86} & \meanstd{0.34}{0.42} & \meanstd{0.12}{0.44} & 2.9 & 0.0 {\scriptsize (0)} & 36.9 & 0.6 \\
MP-Exp & 96.9 & 96.8 & 3.7 & \meanstd{-1.10}{1.11} & \meanstd{0.18}{0.53} & \meanstd{0.20}{0.35} & 2.2 & 0.2 {\scriptsize (4)} & 63.2 & 2.3 \\
\midrule
RandomEnum & 97.0 & 97.0 & 94.5 & \meanstd{-0.65}{0.84} & \meanstd{0.22}{0.13} & \meanstd{0.81}{0.66} & 0.1 & 0.1 {\scriptsize (3)} & 15.8 & 14.1 \\
IonSub & 99.2 & 91.6 & 1.4 & \meanstd{-0.43}{0.40} & \meanstd{0.05}{0.08} & \meanstd{0.07}{0.16} & 7.5 & 0.0 {\scriptsize (1)} & 78.9 & 1.2 \\
\bottomrule
\end{tabular}
\end{adjustbox}
\end{table}

\begin{table}[h!]
\centering
\caption{Model evaluation metrics calculated using \textbf{MLIP ensemble} with \textbf{LeMat-Bulk} as the reference set: distribution, diversity, and HHI. Submissions are organized into four categories separated by horizontal lines: (1) models that incorporate relaxation as part of their generation pipeline (MatterGen, PLaID++, WyFormer, WyFormer-DFT), (2) other generative models, (3) samples from real-world datasets, (4) pre-relaxed samples from non-ML generative baselines. Best values are shown in \textbf{bold}, second-best values are \underline{underlined}. Arrows indicate whether higher ({\textuparrow}) or lower ({\textdownarrow}) values are better. Lower values are better for distribution metrics: JS (Jensen–Shannon divergence), MMD (maximum
mean discrepancy), and FID (Fréchet Inception distance) and for HHI (Herfindahl–Hirschman
index) metrics. Higher values indicate greater diversity across ElemDiv (elemental diversity), SGDiv
(space-group diversity), SizeDiv (crystal size diversity), and SiteDiv (atomic site diversity). “Prod”
and “Res” denote production- and reserve-side HHI components, respectively. These metrics are described in detail in \secref{subsec:unconditional-gen-metrics}.}
\label{table:additional-benchmark-metrics}
\scriptsize
\begin{adjustbox}{max width=\textwidth}
\begin{tabular}{l ccc cccc cc}
\toprule
\textbf{Model} &
\multicolumn{3}{c}{\textbf{Distribution}} &
\multicolumn{4}{c}{\textbf{Diversity}} &
\multicolumn{2}{c}{\textbf{HHI}} \\
\cmidrule(lr){2-4} \cmidrule(lr){5-8} \cmidrule(lr){9-10}
& \textbf{JS {\textdownarrow}} & \textbf{MMD {\textdownarrow}} & \textbf{FID {\textdownarrow}}
& \textbf{ElemDiv {\textuparrow}} & \textbf{SGDiv {\textuparrow}} & \textbf{SizeDiv {\textuparrow}} & \textbf{SiteDiv {\textuparrow}}
& \textbf{Prod {\textdownarrow}} & \textbf{Res {\textdownarrow}} \\
\midrule
MatterGen & 0.44 & 0.006 & 2.13 & 0.64 & 0.14 & 0.24 & 12.78 & 3.52 & 2.72 \\
PLaID++ & 0.43 & 0.029 & 2.69 & 0.68 & 0.26 & 0.21 & 6.01 & 5.23 & 3.36 \\
WyFormer & 0.22 & 0.006 & \underline{1.55} & \underline{0.74} & \underline{0.50} & 0.27 & \underline{23.84} & 3.55 & 2.73 \\
WyFormer-DFT & 0.25 & 0.010 & 1.72 & 0.73 & 0.49 & 0.26 & 23.56 & 3.49 & 2.74 \\
\midrule
ADiT & 0.51 & \textbf{0.002} & 1.72 & 0.71 & 0.03 & 0.23 & 14.18 & 3.52 & 2.78 \\
Crystal-GFN & 0.55 & 0.121 & 43.17 & \textbf{0.75} & 0.29 & \textbf{0.32} & \textbf{32.06} & \underline{3.48} & \textbf{2.34} \\
Crystalformer & 0.26 & \underline{0.003} & 2.31 & 0.71 & 0.35 & \underline{0.31} & 18.59 & 3.78 & 2.82 \\
DiffCSP & 0.45 & 0.008 & 2.25 & 0.73 & 0.14 & 0.24 & 14.42 & \textbf{3.29} & \underline{2.60} \\
DiffCSP++ & \textbf{0.21} & 0.003 & 1.28 & 0.72 & \textbf{0.51} & 0.27 & 20.84 & 3.56 & 2.77 \\
LLaMat2-CIF & 0.32 & \textbf{0.002} & \textbf{1.23} & 0.71 & 0.25 & 0.24 & 9.86 & 3.94 & 2.86 \\
LLaMat3-CIF & 0.34 & 0.005 & 9.48 & 0.70 & 0.10 & 0.29 & 13.15 & 3.83 & 2.81 \\
SymmCD & \underline{0.22} & 0.006 & 1.65 & 0.73 & 0.49 & 0.27 & 19.45 & 3.54 & 2.74 \\
\midrule
Alexandria & 0.09 & 0.000 & 1.31 & 0.71 & 0.50 & 0.25 & 13.50 & 4.02 & 3.33 \\
OQMD & 0.27 & 0.004 & 0.83 & 0.63 & 0.48 & 0.24 & 12.57 & 4.22 & 3.29 \\
AFLOW & 0.28 & 0.004 & 1.78 & 0.69 & 0.46 & 0.25 & 10.07 & 4.12 & 3.22 \\
MP-Exp & 0.33 & 0.021 & 2.90 & 0.72 & 0.71 & 0.28 & 54.79 & 2.78 & 2.27 \\
\midrule
RandomEnum & 0.46 & 0.034 & 2.93 & 0.62 & 0.23 & 0.27 & 15.48 & 3.12 & 3.09 \\
IonSub & 0.32 & 0.004 & 2.04 & 0.67 & 0.36 & 0.26 & 10.82 & 4.15 & 3.86 \\
\bottomrule
\end{tabular}
\end{adjustbox}
\end{table}

\begin{figure}[htbp]
    \centering
    \includegraphics[width=\textwidth]{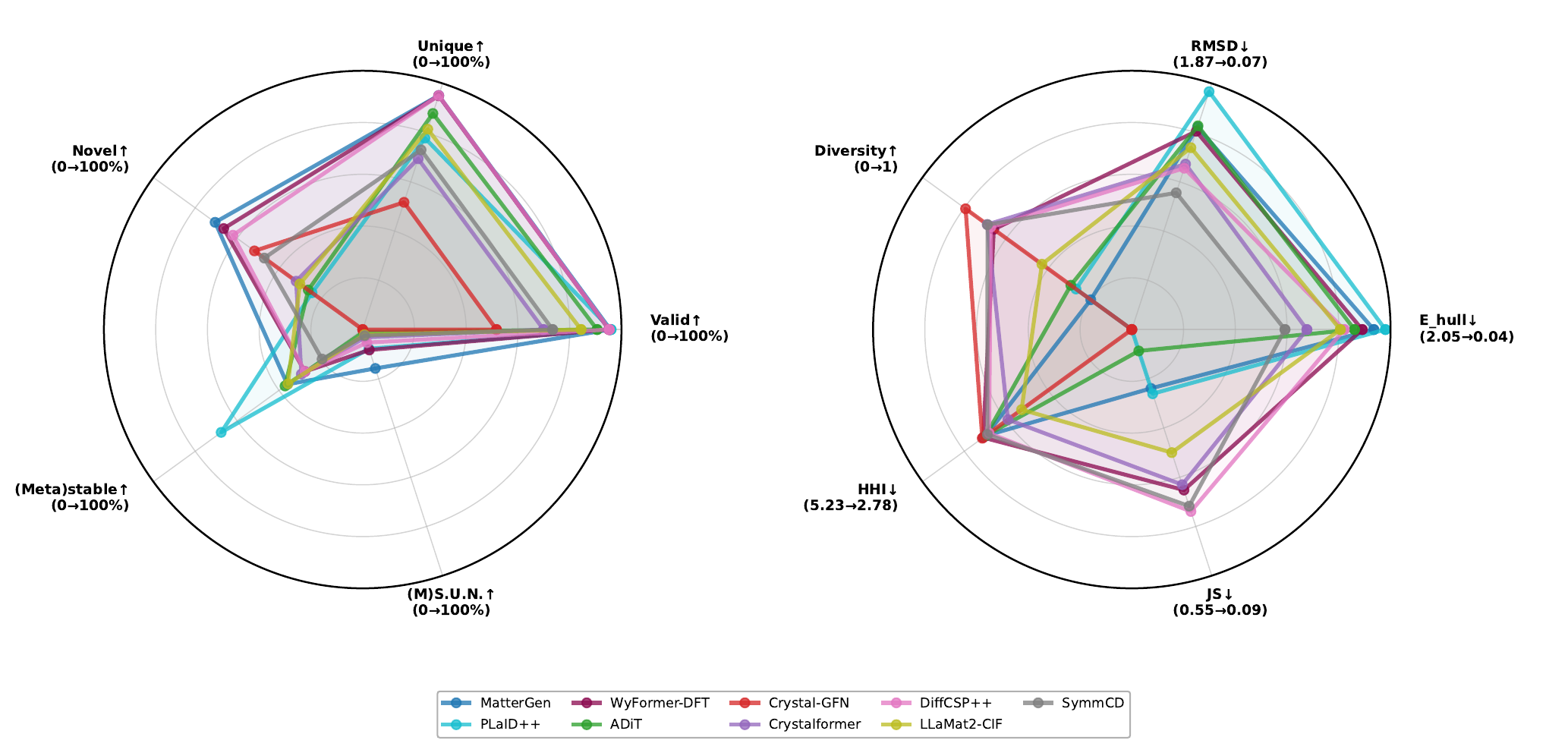}
    \caption{Spider plots comparing generative models using LeMat-Bulk as reference. Left: (M)S.U.N. metrics measuring validity, uniqueness, novelty, (meta)stability, and the combined (M)S.U.N.\ score. Right: Quality metrics including energy above hull ($E_\mathrm{hull}$), structural relaxation (RMSD), diversity (average of elemental, space group, and size diversity normalized), JS distance (Jensen-Shannon distance measuring distributional similarity to reference). All metrics are normalized to the same scale, where outer positions indicate better performance.}
    \label{fig:spider_main_compact}
\end{figure}

\subsection{Results for Generative Models}
\Cref{table:core-benchmark-metrics,table:additional-benchmark-metrics} show that no single
model dominates across all evaluation dimensions. Instead, models occupy different regions of the
trade-off space between thermodynamic stability, novelty, distribution fidelity, and diversity, which
reinforces the need for multi-faceted benchmarking in generative materials design. Validity is often pretty high across models (typically $>90\%$), indicating that generative approaches have mostly solved low-level structural correctness.

Among generative models, MatterGen attains the highest novelty fraction (70.5\%) and the strongest M.S.U.N. rate (15.8\%), indicating that it frequently discovers metastable structures that are also unique and novel relative to LeMat-Bulk. WyFormer-DFT offers a balanced trade-off profile, combining strong novelty and diversity with reasonable energy quality. Diffusion-based approaches demonstrate that continuous denoising processes can effectively explore regions beyond the training distribution. In contrast, Crystal\textendash GFN and PLaID++ are clearly (over)optimised for formation energy and stability, respectively. The latter even shows the best stability metrics and S.U.N, but has substantially lower novelty (24.2\%) and uniqueness (77.8\%) than MatterGen and WyFormer-like models, suggesting a tighter focus on well-explored regions of the MP-20 distribution. Models that incorporate relaxation in their generation pipelines (MatterGen, PLaID++, WyFormer, WyFormer-DFT) achieve systematically lower RMSD values than purely unrelaxed approaches. We examine the interplay between relaxation strategy and benchmark performance in \secref{subsec:prerelax-ablation}. Overall, these results show that generative models have substantial scope for improvement, as their S.U.N. rate never exceeds $1\%$. 
More broadly, the benchmark reveals that no single model excels across all dimensions: strong stability often comes at the cost of novelty, and high exploration capacity does not guarantee thermodynamic feasibility. This underscores the value of multi-metric evaluation in capturing the diverse trade-offs inherent to materials discovery. A visual comparison of the different models can be found in \Cref{fig:spider_main_compact}.

\textbf{Diversity} metrics in \Cref{table:additional-benchmark-metrics} reveal additional aspects of model behavior. WyFormer achieves some of the highest elemental and site diversity (ElemDiv and SiteDiv), demonstrating that Wyckoff-based autoregressive architectures are particularly effective at exploring a wide range of compositions and symmetry environments. SymmCD and DiffCSP++ achieve strong space-group diversity (SGDiv), which is consistent with their explicit treatment of symmetry. Crystal\textendash GFN reaches the highest diversity and the lowest HHI values among generative models, but at the cost of lower stability and validity metrics, highlighting that unconstrained chemical exploration can generate many structurally or thermodynamically implausible candidates. More broadly, we observe that models with very high diversity and novelty often incur larger JS, MMD, and FID values, confirming that aggressive exploration of chemical space naturally
deviates from the reference training distribution. Additionally, higher novelty does not guarantee higher diversity. Mattergen illustrates this: novelty measures how combined structural and compositional attributes deviate from the reference dataset, not each attribute independently.

Taken together, these results show how \textbf{architectural choices} map to distinct performance profiles. LLM-based approaches with reinforcement learning (PLaID++) produce great stability and validity but tend to concentrate in a narrower region of chemical space. Diffusion-based models (DiffCSP, DiffCSP++, MatterGen) strike a balance between stability and novelty, with symmetry-aware variants markedly improving distributional alignment. Autoregressive Wyckoff-based transformers (WyFormer, WyFormer\textendash DFT) and Crystal-GFN systematically enhance diversity and novelty by respectively encoding crystallographic constraints directly in the representation and leveraging the exploration property offered by GFlowNets~\citep{bengio2023gflownetfoundations}. By exposing these trade-offs quantitatively, \lematgen enables users to choose models based on their discovery priorities—whether exploration, thermodynamic plausibility, structural diversity, or supply-chain robustness.

\subsection{Results for Dataset Baselines} 
Because LeMat-Bulk unifies multiple major resources, including Alexandria \citep{schmidt2021crystal, schmidt2024improving} and OQMD \citep{kirklin2015open, saal2013materials}, baseline samples drawn from these databases provide interpretive anchors for generative model performance. As expected, dataset baselines deliver very relaxed structures with near-perfect validity and uniqueness, strong stability metrics, but weak novelty, since these structures already appear in the LeMat-Bulk reference. Their distribution-matching scores (particularly for Alexandria, with the lowest JS and MMD) effectively define the reference behavior that generative models aim to approximate, as shown in \Cref{table:additional-benchmark-metrics}. AFLOW \citep{curtarolo2012aflow} provides a complementary baseline: because it does not contribute directly to LeMat-Bulk, it attains non-zero novelty (21.5\%) alongside strong stability (2.9\% stable, 36.9\% metastable), illustrating the performance of a large, external real dataset under our benchmark. Its relatively high Herfindahl-Hirschman Index (HHI) scores reflect a relatively higher representation of lower abundance/higher cost elements, emphasizing that real-world materials collections can also be composed of elements that are not fully supply chain optimized. The set of real synthesized materials (MP-Exp) outperforms all generative models and computational databases on SGDiv and SiteDiv, suggesting that generative models still underexplore the diversity of crystallographic environments found in nature. Together, these baselines establish that strong performance approaching dataset levels on stability indicates realistic structure generation, while strong novelty performance signals genuine exploration beyond known materials.

\subsection{Results for Non-ML generative baselines}
\label{subsubsec:non-ml}
To directly compare generative models against non-ML discovery strategies, we additionally evaluate two baselines motivated by prior work \citep{szymanski2025establishing, wang2021predicting}. \textbf{Random Enumeration (RandomEnum)} consists of 2,500 structures produced by randomly decorating AFLOW crystal prototypes with charge-balanced compositions, with CHGNet-based metastability pre-filtering ($E_{\text{hull}} \leq 0.1$ eV/atom) applied during generation, following the methodology of Szymanski et al.\cite{szymanski2025establishing}. \textbf{Ion Substitution (IonSub)} consists of 2,500 structures drawn without replacement from the dataset of Wang et al.\cite{wang2021predicting}, who applied the data-mined substitution prediction (DMSP) algorithm \citep{ong2013python} to stable compounds from the Materials Project, replacing one or more ions based on conditional substitution probabilities derived from a probabilistic model trained on the ICSD. After DFT relaxation and filtering for on-hull stability ($E_{\text{hull}} = 0$ eV/atom) and removing duplicates, 6,233 unique stable structures remain, from which we draw our 2,500-structure subsample. The two non-ML baselines use different stability thresholds by necessity: for RandomEnum, structures were generated on-the-fly and a strict on-hull filter ($E_{\text{hull}} = 0$ eV/atom) would have made generation computationally prohibitive, so a looser CHGNet-based metastability threshold ($E_{\text{hull}} \leq 0.1$ eV/atom) was applied during generation. For IonSub, the full DFT-relaxed dataset was available directly from Wang et al.\cite{wang2021predicting}, making it straightforward to filter for strictly on-hull structures at no additional computational cost. Both baselines are processed under the identical \lematgen pipeline as all other submissions. 

The results illustrate the two extremes of the stability-novelty tradeoff in their starkest form. RandomEnum achieves very high novelty (94.5\%) by construction -- randomly decorated prototypes naturally fall outside known databases -- but near-zero strict stability (0.1\%), demonstrating that unconstrained combinatorial exploration of chemical space is thermodynamically highly inefficient. Crucially, this low stability persists even after CHGNet metastability pre-filtering
during generation, because that filter was applied against the MP convex hull, whereas
\lematgen evaluates against the more complete LeMat-Bulk hull, which includes
significantly more competing phases and therefore imposes a more stringent stability
criterion. Its high RMSD (0.81~\AA) further confirms that randomly decorated structures lie far from local energy minima. 

IonSub occupies the opposite extreme: near-zero novelty (1.4\%), since ion-substituted MP compounds are largely already captured within LeMat-Bulk's comprehensive phase coverage, but strong stability (7.5\% stable, 78.9\% metastable) and the lowest relaxation RMSD of all submissions (0.07~\AA). However, these apparent advantages should be interpreted with care: the stability of IonSub is not a property of the generation strategy but a direct artifact of explicit DFT on-hull filtering applied during dataset construction, and its low relaxation RMSD reflects DFT pre-relaxation rather than geometric quality of the generation process itself. 

Taken together, these two baselines confirm that the stability--novelty tradeoff observed across generative models is not an artifact of neural network training, but a fundamental tension in materials discovery: strategies that maximally explore novel chemical space are more likely to produce thermodynamically implausible structures, while strategies anchored to known stable compounds sacrifice novelty. The best generative models in \lematgen begin to navigate this tradeoff more effectively than either non-ML strategy, combining meaningful novelty with reasonable metastability -- though substantial room for improvement remains.

\section{Discussion}
Generative models for crystalline materials have advanced rapidly, yet the absence of shared evaluation standards has made progress difficult to measure. \lematgen addresses this gap by introducing the first unified, open benchmarking framework for unconditional crystal generation. It builds on standardized validity checks, self-consistent MLIP-based stability assessments, and well-defined measures of uniqueness, novelty, and diversity. By grounding stability and novelty evaluations in the large-scale LeMat-Bulk dataset, the benchmark offers a stringent and realistic assessment of a model’s capacity to propose genuinely new materials.

Our evaluation of 12 state-of-the-art generative models highlights the diversity of modeling trade-offs: diffusion and autoregressive architectures tend to explore more compositionally and structurally diverse regions of chemical space, while RL- and LLM-guided approaches produce more stable but less novel candidates. No model dominates all metrics, underscoring the need for multi-dimensional evaluation and improved generative paradigms. The \lematgen leaderboard and open-source implementation provide a common reference point for future model development, reproducibility, and community-driven extensions. We see this as a first step toward closing the loop between computational generation and experimental validation.

This release focuses on unconditional generation evaluation, with conditional generation benchmarks and expanded model implementations planned for future updates. Key challenges remain. Data quality is a persistent bottleneck: widely used datasets often lack compositional diversity, structural metadata, or negative examples necessary for robust generative model training. Most generative models assume idealized, defect-free crystals, overlooking critical phenomena like disorder, doping, and non-stoichiometry that shape real-world functionality. Moreover, stability assessment relies on ensembles of MLIPs, which, despite averaging, can deviate systematically from DFT, especially near stability thresholds. Finally, although we assess thermodynamic plausibility, our framework does not yet capture kinetic barriers, synthesis feasibility, or real-world constraints. Bridging these gaps, especially toward synthesis-aware and experimentally grounded pipelines, will require tighter integration between data, modeling, and validation across disciplines. Environmental and sustainability considerations are discussed in \Cref{app:sec:sustainability}.

\section{Methods}
\label{sec:benchmark-implem}

\subsection{Towards an Open Benchmark Framework}
\label{subsec:leaderboard}

To converge towards a community-wide benchmarking framework, we adopt several concrete steps. Specifically, we provide a fully open-source implementation of the proposed evaluation metrics for unconditional evaluation, which can be used and updated by the community as the field evolves. Our contributions include: (i) use of LeMat-Bulk as the reference dataset for (M.)S.U.N., (ii) an ensemble of MLIPs for robust formation energy prediction, with uncertainty quantification, (iii) energy above hull calculations between an MLIP prediction and its corresponding convex hull, instead of a MLIP vs DFT-hull; (iv) enhanced validity criteria, (v) standardized diversity scores, (vi) resource efficiency reporting, (vii) pre-relaxation and post-relaxation metrics, (viii) RMSD check, 
(ix) a \href{https://huggingface.co/spaces/LeMaterial/LeMat-GenBench}{public leaderboard on Hugging Face} to further facilitate fair model comparisons. 

For the \textbf{leaderboard}, the submission process is as follows: (i) authors submit 2500 generated crystal structures; (ii) authors may also submit their packaged model, granting a compliance badge (optional); (iii) evaluation metrics are computed using our open reference implementation; (iv) results are displayed with multiple views to support comparison across tasks and generation scenarios. The code is available at \url{https://github.com/LeMaterial/lemat-genbench}.

This evaluation framework aims to provide clear, fair, and rigorous standards for evaluating generative models in crystal generation, while remaining flexible to evolving research needs and application contexts. Going further, we aim to extend this benchmark to the conditional generation tasks, to further bridge the gap between computational prediction and experimental realization. This infrastructure becomes particularly important as the field moves toward closed-loop discovery pipelines that integrate computational prediction, experimental validation, and synthesis planning. 

The metrics introduced in \secref{subsec:unconditional-gen-metrics} are not purely intrinsic properties of a model: they depend on concrete methodological choices in the evaluation pipeline. For example, stricter validity filters change how many structures proceed to stability checks, the way we construct the convex hull shifts stability thresholds, and the choice of reference dataset directly affects both novelty and stability via the phase diagram. Therefore, in this section we motivate and empirically justify four key design decisions in \lematgen: the validity criteria, the self-consistent MLIP-based hull construction, the pre- vs.\ post-relaxation protocol, and the choice of reference dataset.

\subsection{Design rationale for the validity metric}
\label{sec:validity-metric-logic}

\begin{figure}[h]
    \centering
    \includegraphics[width=\textwidth]{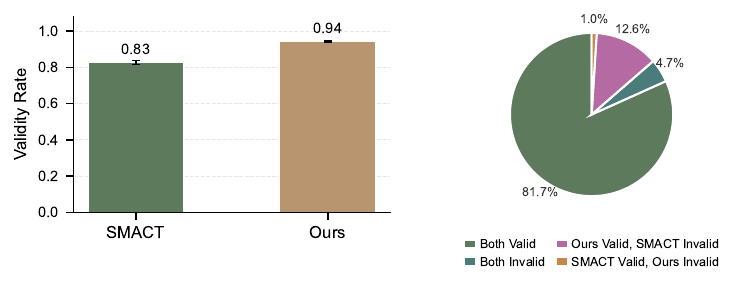}
    \caption{\textbf{Validity comparison between SMACT and our proposed method.} 
    Five random samples of 1,000 structures from LeMat-Bulk were evaluated using both methods.
    \textbf{Left:} Overall validity rates with standard deviation across seeds. 
    \textbf{Right:} Agreement breakdown. Our method recovers 12.6\% of structures 
    that SMACT rejects, primarily f-block and metalloid-containing structures and elements with 
    multiple oxidation states, while disagreeing on only 1.0\% in the opposite direction. 
    Seed-level results in \Cref{fig:seed_variation}.}
    \label{fig:validity_comparison}
\end{figure}

Existing validity checks often reject a large fraction of experimentally confirmed structures, penalizing generative models for producing chemically plausible outputs and distorting benchmark comparisons. This is especially true for SMACT~\citep{davies2019smact}, which only marks $\sim 83\%$ of LeMat-Bulk structures~\citep{siron2025lemat} as valid, with most invalid labels arising from strict charge-neutrality assumptions, fixed oxidation state rules, and metallic structures improperly assessed using oxidation state assignments. Such rigidity causes systematic false negatives for particularly mixed-valence structures, alloys, and structures containing f-block elements.

Our hierarchical validity metric addresses these issues by incorporating empirical oxidation-state frequencies from LeMat-Bulk and ICSD and by introducing a flexible oxidation state assignment protocol for structures failing these strict checks. This enables more flexible treatment of metallic structures and structures containing ambiguous and uncommon oxidation states. As shown in \Cref{fig:validity_comparison} and \Cref{fig:seed_variation}, this approach increases validity from $82.7\%$ (SMACT) to $94.3\%$ while maintaining a low $\sim 1\%$ false-positive rate. It correctly recovers many structures, such as rare-earth oxides and metalloid-containing structures, that SMACT incorrectly rejects.

Because this metric relies on empirical priors, it may slightly favor well-documented elements, but the low false-positive rate suggests this bias is limited in practice. Importantly, many valid structures may still contain scarce, toxic, or potentially radioactive elements; rather than marking them as invalid, we manage these undesirable compositions downstream through the HHI index. This separation ensures that validity reflects structural and chemical soundness, while compositional preferences remain application-specific.

\subsection{Use of a self-consistent MLIP convex hull}
\label{sec:self-consistent-hull}

\begin{figure}
    \centering
    \includegraphics[width=1\linewidth]{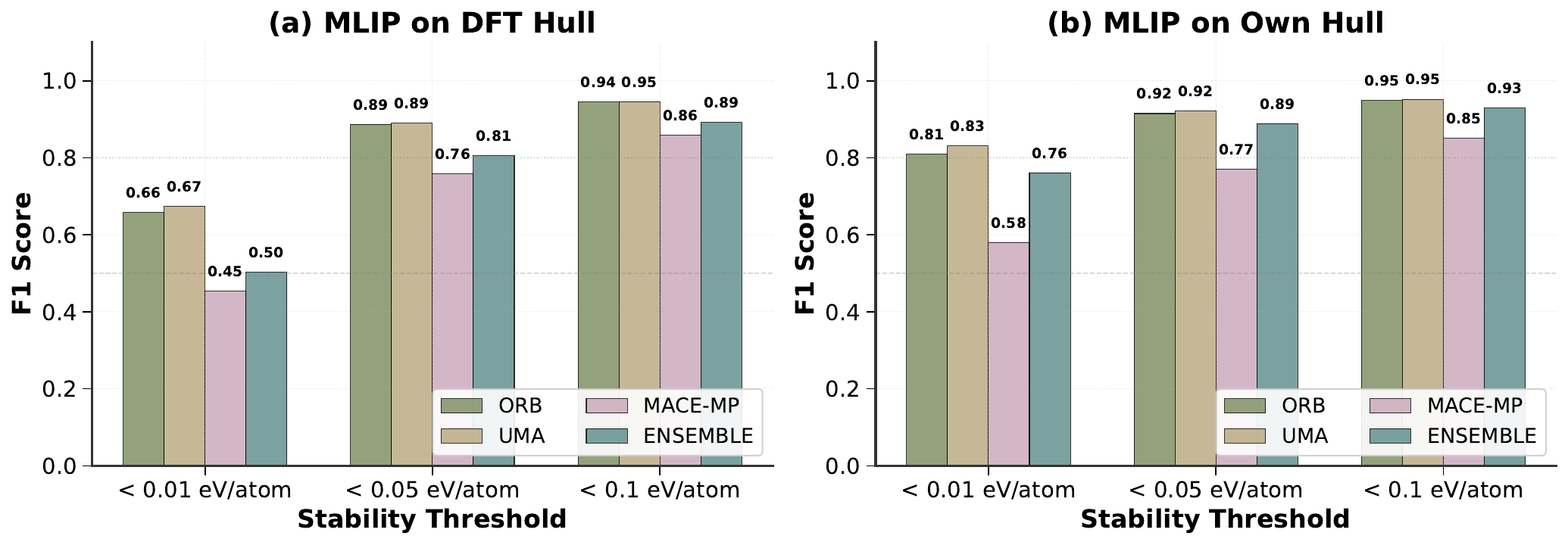}
    \caption{F1-score for stability prediction on LeMat-Bulk. Three MLIPs 
    (\texttt{orb-v3}, \texttt{uma-s1}, \texttt{mace-mp}) are evaluated 
    at different $E_{\text{hull}}$ thresholds. \textbf{(a)} MLIP energies 
    compared against a DFT-constructed hull. \textbf{(b)} Self-consistent 
    approach where each MLIP defines its own hull. The self-consistent 
    method yields consistently higher F1-scores, particularly at strict 
    thresholds.}
    \label{fig:bulk_hull_f1}
\end{figure}

Once structures pass validity screening, stability assessment relies on computing their energy above the convex hull ($E_{\text{hull}}$). Since DFT computations are expensive, a widely used shortcut is to predict candidate energies with an MLIP while comparing them against a DFT-derived hull from a reference dataset. This mixed-reference setup introduces systematic inconsistencies: MLIPs and DFT operate on different energy baselines determined by their respective training sets and reference choices. Small shifts, even a few tens of meV/atom, can flip stability labels. This is problematic because $E_{\text{hull}}$ is often treated as a hard decision boundary when evaluating the quality of structures coming out of generative models.

To eliminate these inconsistencies, \lematgen adopts a self-consistent approach: each MLIP both predicts energies and defines its own convex hull. To compare this approach against the mixed-reference strategy, we treat DFT stability labels from LeMat-Bulk as ground truth. A structure is considered truly stable if its DFT energy lies within a given threshold of the DFT convex hull. We then evaluate how well each MLIP recovers these labels using either (i) the DFT hull or (ii) its self-consistent hull.

In \Cref{fig:bulk_hull_f1}, we calculate $F_1 = 2 \frac{\text{precision} \cdot \text{recall}}{\text{precision} + \text{recall}}$, where precision is the fraction of predicted-stable structures that are stable according to DFT, and recall is the fraction of truly stable structures correctly identified. As shown in \Cref{fig:bulk_hull_f1}, F1-scores improve markedly under the self-consistent setup e.g., Orb-v3: $0.66 \to 0.81$, UMA: $0.67 \to 0.83$ at $E_{\text{hull}} < 0.01$ eV/atom. The gain likely stems from model-specific energy biases canceling out when the same potential defines both candidate energies and its reference hull. Additional analyses are provided in \Cref{app:mlip_hull}. Overall, these results show that the self-consistent MLIP-based hull is systematically more reliable for stability assessment and should be preferred for benchmarking generative models.

\subsection{Pre-Relaxation vs.\ Post-Relaxation metrics}
\label{subsec:prerelax-ablation}

All results in \Cref{table:core-benchmark-metrics} are computed on the structures exactly as generated, without applying any additional relaxation. This is crucial for fair comparison: it avoids introducing the bias of a single external relaxer and directly evaluates the geometric quality of the structures produced by each model. In practice, high-quality generative pipelines should already output relaxed or near-relaxed structures, so evaluating them in their raw form is both more meaningful and more desirable. Nevertheless, since some models explicitly integrate relaxation into their generation pipeline, while others do not, we also compute post-relaxation metrics. In particular, we apply a uniform UMA relaxation to all generated structures (\Cref{tab:uma_bulk_postrelax}), to understand how relaxation affects evaluation. 

Models that internally relax their outputs show minimal change. When comparing with respect to UMA (\Cref{tab:uma_bulk_prerelax}), PLaID++ has the lowest pre-relaxation RMSD among generative models ($0.10$~\AA), comparable to the dataset baselines (AFLOW and Alexandria, $0.11$~\AA~ each) and DFT-relaxed non-ML baselines (IonSub, $0.06$~\AA), confirming that its outputs already reside near MLIP energy minima.

Models without internal relaxation change substantially after refinement.
ADiT shows the strongest shift: stability increases from $0.1\%$ to $17.6\%$ and metastability from $35.4\%$ to $ 62.5\%$. DiffCSP improves similarly (stability: $2.0\%$ to $5.9\%$, M.S.U.N.: $8.2\%$ to $19.4\%$). These gains indicate that several models produce geometrically plausible but not fully relaxed structures, and thus benefit greatly from an additional energy minimization step.

Given these large differences, we report both pre- and post-relaxation results (\Cref{tab:uma_bulk_prerelax}, \Cref{tab:uma_bulk_postrelax}), since the magnitude of the shift carries important diagnostic information about each model’s geometry quality and reliance on downstream refinement. This being said, the \lematgen leaderboard adopts pre-relaxation metrics as the default, to reflect the true generative capability of each model and encourage pipelines that produce well-relaxed structures directly.

\subsection{Reference Dataset: LeMat-Bulk vs MP-20}
\label{sec:lemtavsmp}

Choosing a reference dataset requires balancing two objectives. Model developers benefit from evaluations aligned with the training distribution, where differences isolate architectural and algorithmic contributions. End-users, however, require metrics that reflect real discovery potential, where stability, novelty, and phase competition are assessed against the broadest available set of known materials. To serve both needs, we report MP-20 results in \Cref{app:subsec:mp20_results} but adopt LeMat-Bulk as the primary reference for the public leaderboard.

LeMat-Bulk ($\sim$5M structures) provides the comprehensive phase coverage needed for realistic evaluation. In contrast, MP-20’s limited compositional and structural variety leads to inflated novelty and stability estimates, since many “new” or “stable” structures simply fall outside its constrained reference set. This difference is substantial: novelty drops from 83.1\% to 70.5\% for MatterGen; stability rates are divided by 4--10$\times$ (e.g., PLaID++: 27.2\% to 6.4\%; LLaMat2-CIF: 3.9\% to 0.4\%); and S.U.N.\ rates collapse further (e.g., DiffCSP: 4.1\% to 0.0\%). These shifts illustrate that smaller references can substantially overestimate a model’s practical usefulness.

As generative models scale toward deployment in real discovery settings, evaluation against a comprehensive reference set becomes essential. MP-20 remains valuable for controlled methodological benchmarking, but LeMat-Bulk provides a more reliable basis for assessing discovery-oriented performance. We anticipate expanding reference sets as the field matures and more large-scale, consistently curated materials databases become available.



\section{Acknowledgement}
We acknowledge the fruitful discussions and insights brought by Alfonso Amayuelas, Anuroop Sriram, Benjamin Kurt Miller, Edvin Fako, Hashim Piracha, Kishalay Das, Tristan Deleu, Victor Schmidt, Zekun Danny Ren, Aron Walsh, and Tian Xie. Through A.D., this project was provided with computer and storage resources by GENCI (Grand Équipement National De Calcul Intensif) at CINES and IDRIS under the allocations 2024-AD011015800, 2025-A0181016212, and 2025-AD011016353 on the supercomputers Jean Zay and Adastra’s CPU and A100/MI300 partitions. 


\phantomsection\addcontentsline{toc}{section}{References}
\bibliographystyle{naturemag}
\bibliography{references}

\clearpage
\setcounter{secnumdepth}{3}
\renewcommand{\thefigure}{S\arabic{figure}}
\renewcommand{\thetable}{S\arabic{table}}
\renewcommand{\theequation}{S\arabic{equation}}
\renewcommand{\thesection}{S\arabic{section}}
\renewcommand{\theHfigure}{S\arabic{figure}}
\renewcommand{\theHtable}{S\arabic{table}}
\renewcommand{\theHequation}{S\arabic{equation}}
\renewcommand{\theHsection}{S\arabic{section}}
\setcounter{figure}{0}
\setcounter{table}{0}
\setcounter{equation}{0}
\setcounter{section}{0}
\phantomsection\addcontentsline{toc}{section}{Supplementary Material}

\section*{\centering Supplementary Material}    

\begin{tcolorbox}[title=Box 1: Key Terms in Generative Modeling, colback=gray!5, colframe=gray!40!, fonttitle=\bfseries, coltitle=black]

\textbf{Generative Model:} A machine learning model that learns a data distribution $p(\mathbf{x})$ (or a conditional distribution $p(\mathbf{x}|\mathbf{z})$ or $p(\mathbf{x}|\mathbf{c})$) and can generate new samples $\mathbf{x}' \sim p(\mathbf{x})$ that resemble the training data.

\vspace{0.5em}
\textbf{Latent Space:} A lower-dimensional representation space $\mathbf{z} \in \mathbb{R}^d$ learned by models such as VAEs or GANs, where semantic attributes of the data are often encoded.

\vspace{0.5em}
\textbf{Prior Distribution:} A predefined distribution (e.g., Gaussian) over the latent variables, typically denoted as $p(\mathbf{z})$, from which samples are drawn during generation.

\vspace{0.5em}
\textbf{Decoder / Generator:} A neural network (often denoted $G(\mathbf{z})$) that maps latent codes $\mathbf{z}$ to data samples $\mathbf{x}$.

\vspace{0.5em}
\textbf{Reconstruction Loss:} A metric used in training autoencoders and VAEs that measures how well the generated sample $\hat{\mathbf{x}}$ matches the original input $\mathbf{x}$:
\begin{equation*}
    \mathcal{L}_{\text{recon}} = \| \hat{\mathbf{x}} - \mathbf{x} \|^2 \quad \text{or} \quad -\log p(\mathbf{x}|\mathbf{z}).
\end{equation*}

\vspace{0.5em}
\textbf{KL Divergence:} A measure of how much one probability distribution differs from another. Commonly used in VAEs to regularize the encoder:
\begin{equation*}
    \mathcal{L}_{\text{KL}} = D_{\text{KL}}(q(\mathbf{z}|\mathbf{x}) \| p(\mathbf{z})).
\end{equation*}

\vspace{0.5em}
\textbf{Mode Collapse:} A failure mode in GANs where the generator produces samples with limited diversity, collapsing to a few modes of the data distribution.

\vspace{0.5em}
\textbf{Conditional Generation:} Generation of samples $\mathbf{x}$ based on specified properties or constraints $\mathbf{c}$, e.g., $p(\mathbf{x}|\mathbf{c})$, enabling property-guided design.

\vspace{0.5em}
\textbf{Inverse Design:} The process of searching the input space (e.g., structure, composition) that maps to a desired target property, often using a generative model or an optimization loop in latent space.

\vspace{0.5em}
\textbf{Diffusion Models:} A class of generative models that learn to reverse a stochastic diffusion process. Data $\mathbf{x}_0$ is gradually perturbed into noise via:
\[
q(\mathbf{x}_t | \mathbf{x}_0) = \mathcal{N}(\mathbf{x}_t; \sqrt{\alpha_t} \mathbf{x}_0, (1 - \alpha_t) I).
\]
and a neural network is trained to denoise $\mathbf{x}_t$ to recover $\mathbf{x}_0$ through a learned reverse process $p_\theta(\mathbf{x}_{t-1}|\mathbf{x}_t)$.

\vspace{0.5em}
\textbf{Score-Based Models:} Closely related to diffusion models, they learn the score function $\nabla_{\mathbf{x}} \log p(\mathbf{x})$ and use Langevin dynamics or ODE solvers to sample from the data distribution.

\[
\log p(\mathbf{x}) = \log p(\mathbf{z}) + \sum_{k=1}^{K} \log \left| \det \frac{\partial f_k^{-1}}{\partial \mathbf{x}} \right|.
\]

\vspace{0.5em}
\textbf{Flow Matching:} A recent generative approach that avoids training score functions or simulating diffusion. It directly learns a vector field $\mathbf{v}_\theta(\mathbf{x}, t)$ that maps noise to data through an ODE:
\[
\frac{d\mathbf{x}}{dt} = \mathbf{v}_\theta(\mathbf{x}, t).
\]
This method can be trained via supervised learning on synthetic trajectories or velocity fields between the base and target distributions.

\end{tcolorbox}

\begin{tcolorbox}[title=Box 2:  Key Terms in Crystallography \& Materials Science, colback=gray!5, colframe=gray!40!, fonttitle=\bfseries, coltitle=black]
\hypertarget{box:crystallography}{}

\textbf{Crystal Lattice:}
A crystal structure is periodic in three dimensions.
This periodicity is described by the lattice, which is defined as 
$$
\mathbf{L} = \{\, l_1 \mathbf{a}_1 + l_2 \mathbf{a}_2 + l_3 \mathbf{a}_3 \;\mid\; l_1, l_2, l_3 \in \mathbb{Z} \,\},
$$
where $\mathbf{a}_1$, $\mathbf{a}_2$, $\mathbf{a}_3$ are basis vectors of $\mathbb{R}^3$.

\vspace{0.5em}
\textbf{Unit Cell:}
A unit cell is the smallest unit that can be translated to define the whole lattice. In three dimensions, it is always a parallelepiped.

\vspace{0.5em}
\textbf{Lattice Parameters:}
A lattice is typically defined in two ways: either as a set of three basis vectors, or as a set of lattice parameters $(a, b, c, \alpha, \beta, \gamma)$, where $a, b, c$ are the lengths of edges of the unit cell, and $\alpha, \beta, \gamma$ are the angles between them.

\vspace{0.5em}
\textbf{Symmetry:}
An object's symmetry is given by the set of geometric transformations that map the object onto itself, leaving it invariant.

\vspace{0.5em}
\textbf{Space Group:}
Crystals can be classified by their symmetries.
They possess the translational symmetry of their crystal lattices, and they may also have the point group symmetries of rotations and reflections within a unit cell.
The combination of translational and point group symmetries can yield more transformations that a crystal can be symmetric to, including screw and glide symmetries.
The full set of symmetric transformations that leave a crystal invariant defines the space group of the crystal.
In three dimensions, there are 230 distinct space groups.

\vspace{0.5em}
\textbf{Wyckoff Position:}
Applying symmetry operations to a crystal may leave some atoms unaffected; for example, a rotation about an axis leaves atoms on the axis in the same position.
The set of symmetry operations that do not move a position defines that position's site symmetry.
A Wyckoff position is a set of positions that all have the same site symmetries, or conjugate site symmetries.
For example, all points along a mirror plane may belong to the same Wyckoff position, while a point at the origin of a unit cell may have its own.
Every point in a crystal can be assigned a Wyckoff position.

\vspace{0.5em}
\textbf{Formation Energy (\bm{$E_f$}):}
The formation energy of a crystal is the difference in energy between the crystal and its constituent elements. It can be calculated using: $E_f = E_{\text{total}} - \sum_{i} n_i \mu_i$

\vspace{0.5em}
\textbf{Energy Above Convex Hull (\bm{$E_{\text{hull}}$}):}
The convex hull gives linear combinations of known phases that represent the lowest-energy mixtures of materials; if a material has an energy above the hull ($E_{\text{hull}} > 0$), it is energetically favorable for it to decompose into a combination of stable phases and is therefore thermodynamically unstable. Mathematically, this can be represented as $E_{\text{hull}} = E_{\text{total}} - E_{\text{hull}}^{\min}$
For example, the convex hull of table salt, NaCl, also includes pure stable Na, pure Cl, as well as NaCl$_3$. However, Na$_2$Cl has a higher formation energy than the combination of NaCl and pure Na, so it is unstable. 

\vspace{0.5em}
\textbf{Metastable:}
Even if a crystal is not in its lowest possible energy state, it may still be metastable, meaning that a potential energy barrier prevents it from easily transitioning to a lower-energy state.
A crystal having a low energy above the convex hull while also being at an energy minimum may indicate that it is metastable.
Metastable materials are still important: for example, diamond is metastable, but does not readily convert to a lower energy state under normal conditions.

\vspace{0.5em}
\textbf{CIF:} A Crystallographic Information File is a string-based encoding of a crystal that includes information such as atom positions, unit cell parameters, and chemical elements.

\vspace{0.5em}
\textbf{Relaxation Trajectory:} sequence of intermediate atomic configurations produced during an energy-minimization procedure (e.g., via DFT, MLIPs, or classical force fields). Starting from an initial crystal structure, the optimizer iteratively updates atomic positions and lattice parameters to reduce the system’s total energy.

\end{tcolorbox}

\begin{figure}
\centering
\includegraphics[width=0.8\textwidth]{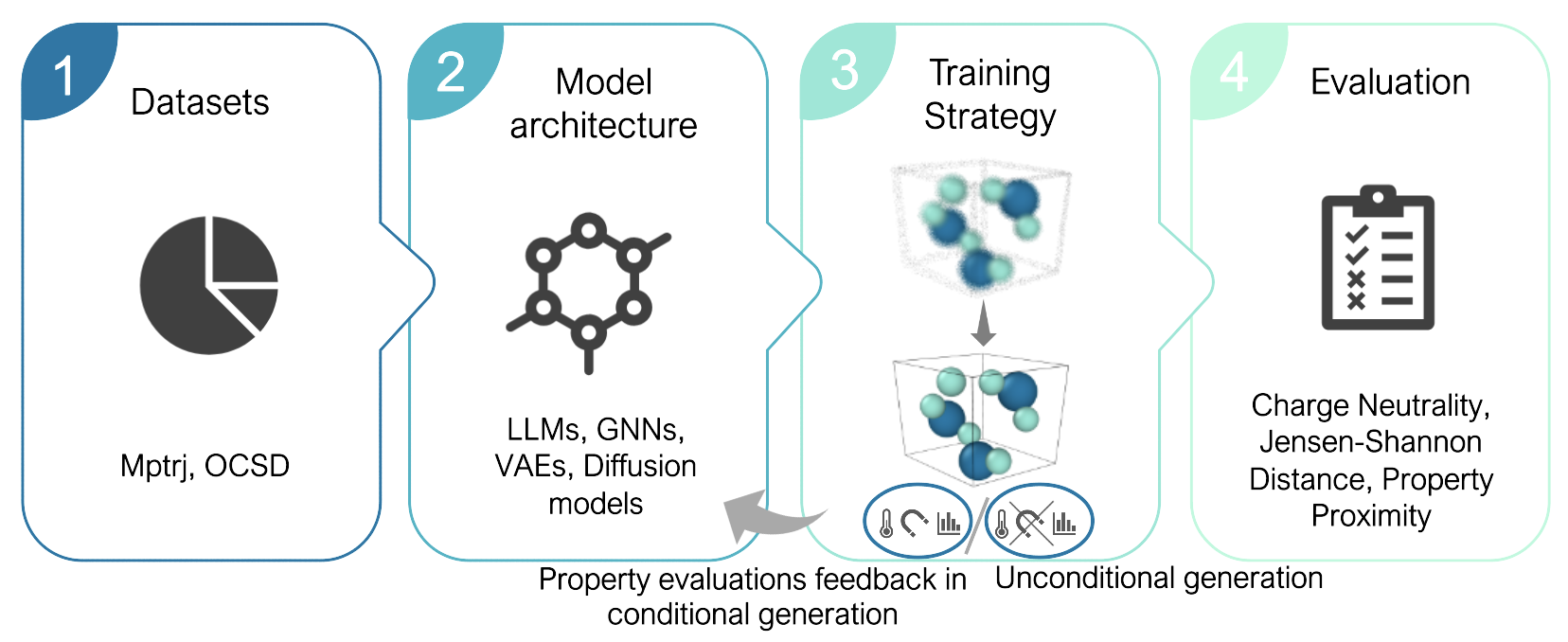}
\caption{An overview of the generative AI paradigm for candidate structure generation and optimization that underpins much of the work reviewed herein.}
\label{fig:pipeline_sup}
\end{figure}

\section{Survey of Generative Models for Crystalline Materials}
\label{app:sec:survey}

\begin{figure}
\centering
\includegraphics[width=\textwidth]{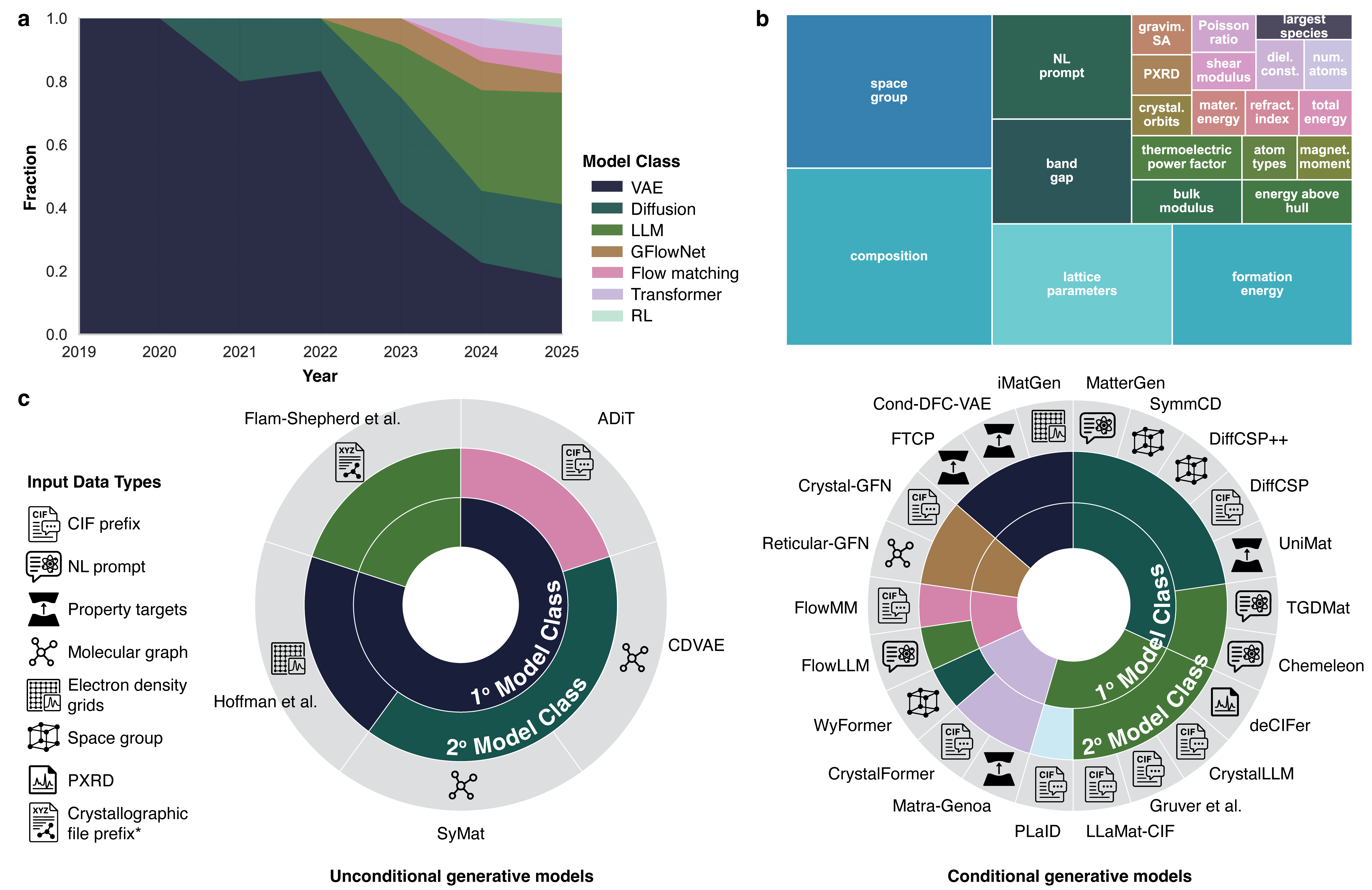}
\caption{Overview of generative models for materials discovery discussed in this work. \textbf{(a)} Evolution of major model architectures over time, showing the early dominance of VAEs and the recent rise of LLMs. \textbf{(b)} Treemap of target properties optimized across models; box size reflects the proportion of papers mentioning each property. Space group, composition, lattice parameters, and formation energy are the most common. \textbf{(c)} Model distribution for unconditional (left) and conditional (right) generation tasks. Most conditional models can also perform unconditional generation, but not vice versa. Methods are grouped by their two dominant architecture families ($1^o$ and $2^o$ model classes) for simplicity, with all colors matching panel (a). As many things have been simplified for this figure, please see \Cref{app:sec:survey} for details on the various methods presented. Each model is annotated with its primary input data type. \textbf{Abbreviations:} LLM = large language model; VAE = variational autoencoder; RL = reinforcement learning; NL prompt = natural language prompt; PXRD = powder X-ray diffraction. ``CIF prefix'' typically includes composition, space group, and lattice parameters; ``Crystallographic file'' refers to any file encoding structure data (e.g., XYZ, PDB, CIF).}
\label{app:fig:taxonomy}
\end{figure}

\subsection{Early Deep Learning Approaches}

Variational autoencoders (VAEs) \citep{kingma2013autoencodingvariationalbayes} were among the first deep generative models applied to crystal structure generation, offering a powerful framework for encoding materials in continuous latent spaces. Early efforts by iMatGen \citep{noh_inverse_2019} employed 3D grid-based image representations of crystal structures to learn a latent space of inorganic materials, further refined by a binary classifier that distinguished stable from unstable structures based on formation energies. This approach successfully demonstrated the ability to generate novel vanadium oxide materials. Subsequent work by~\cite{hoffmann_data-driven_2019} advanced this methodology by combining U-Net segmentation with VAEs in a two-step approach: first generating 3D density maps via a VAE, then segmenting these maps into 3D voxel grids using a U-Net model. Building on this foundation, \cite{court_3-d_2020} introduced conditional deep feature consistent VAEs (Cond-DFC-VAE) to enable property-guided crystal generation. A limitation of these early approaches is the use of voxel grid representations that are difficult to decode into physically meaningful atom coordinates and types. A significant advancement came from \cite{ren_invertible_2022}, who developed a generalized invertible representation encoding both structural information (atom coordinates and types) and lattice parameters through reciprocal space transformations in the FTCP framework. This representation facilitated inverse design of inorganic crystals with targeted properties such as formation energy and bandgap, representing an important step toward property-conditioned generation.

Generative adversarial networks (GANs) \citep{goodfellow2014generative} have also been used for crystal structure prediction and candidate generation. In this framework, a discriminator and a generator neural network are trained, with the first being trained to distinguish between real data and generated samples, and the second trained adversarially to produce samples that fool the discriminator. \cite{kim2020} used GANs to predict crystal structures conditioned on composition. It was notably one of the first methods to introduce a coordinate-based representation for crystal structure generation. \cite{long2021constrained} introduced a method to constrain generation to desired properties by additionally optimizing with respect to a property prediction network. CubicGAN \citep{zhao2021} introduced a learnable space group aware component to the representation, which proved influential, although restricted to cubic systems.

\subsection{Diffusion and Flow-based Models}

Diffusion models \citep{sohl-dickstein15, ho2020, song2021scorebased} have become a staple in image generation and have also been adapted successfully for crystal generation. They have shown promising capabilities in generating stable 3D periodic structures for novel crystalline materials. These approaches operate on the principle of gradually denoising randomly sampled configurations to produce physically plausible structures, incorporating both geometric and chemical constraints.

Early implementations such as CDVAE \citep{xie2021crystal} and SyMat \citep{luo2023towards} combined VAEs with score-based denoising networks operating directly on atomic coordinates. These hybrid models first predict lattice parameters and atomic composition via the VAE component, and then predict atomic coordinates through score-based diffusion. Subsequent architectures like DiffCSP \citep{jiao2023crystal} and MatterGen \citep{zeni2025generative} advanced this approach by adopting joint diffusion frameworks that simultaneously model atomic composition, fractional coordinates, and lattice parameters. These methods leverage SE(3)-equivariant neural networks adapted for periodicity \citep{duval2023hitchhiker} to predict the denoised composition, coordinates, and lattice parameters, ensuring both Euclidean and periodic invariance in the learned distribution. An alternative representation strategy was proposed in UniMat \citep{yang2023scalable}, which introduced a unified crystal representation using a 4D tensor where atomic positions are stored within corresponding atom entries in the periodic table. This approach employs interleaved attention and convolution layers as denoising networks, effectively addressing the challenge of jointly modeling discrete atom types and continuous atomic coordinates.

Recent work has further extended diffusion models to incorporate domain knowledge and constraints. DiffCSP++ \citep{jiao2024space} and SymmCD \citep{levysymmcd} enforce space group symmetry constraints during generation, significantly reducing the effective dimensionality of the generative task. By constraining atomic coordinates to follow specified Wyckoff positions and space group symmetries, these models not only improve computational efficiency but also ensure that generated structures adhere to physically meaningful crystallographic rules. DiffCSP++ is based on using predefined structural templates for each space group, whereas in SymmCD, the templates are also generated, resulting in increased diversity.

Text-guided diffusion models represent another frontier, with approaches like TGDMat \citep{das2025periodic} and Chemeleon \citep{park2025exploration} incorporating contextual representations from pretrained language models to enable generation of stable periodic materials that align with conditions specified in natural language descriptions.

Flow matching (FM) methods \citep{lipman2023flow, albergo2023building, liu2023flow} are closely related to diffusion models~\citep{gao2025diffusion}, and learn a time-dependent velocity field that smoothly transforms an arbitrary base distribution into the target distribution of stable materials. FlowMM \citep{miller2024flowmm} pioneered this approach for crystal generation, defining specialized representations that enforce global rotational and translational symmetries while respecting periodic boundary conditions. Flow matching notably offers faster generation than diffusion models, since it requires fewer evaluations of the neural network. 
Building on this work, FlowLLM \citep{sriram2024flowllm} introduced a hybrid approach that leverages language models to provide a chemically informed base distribution. Rather than starting from simple distributions, FlowLLM generates initial material representations using an LLM trained on known structures \citep{gruver2024fine}, then refines these samples through flow matching that preserves composition while updating atomic positions and lattice parameters. Sampling from a base distribution defined by an LLM resulted in an increase in the rate of stable materials compared to sampling a generic base distribution.

Recent work demonstrates the potential of encoding both periodic (crystalline) and non-periodic (molecular) structures using a single unified flow matching framework. 
ADiT~\citep{joshi2025all} utilizes an autoencoder to learn a joint latent representation of both crystals and molecules, followed by flow matching in the unified latent space to generate new latent representations and decode them to valid crystals or molecules. 
This approach enables learning from and generating diverse materials, including metal-organic frameworks (MOFs), which exhibit both periodic and molecular characteristics. 
Unlike previous approaches that require specialized equivariant architectures, ADiT simplifies the generative process by using a standard Gaussian base distribution and Transformer architecture as the time-dependent velocity field in latent space.
This simplification leads to faster generation times while still demonstrating strong performance on the generation of stable structures.

\subsection{Reinforcement Learning and GFlowNets}

By contrast with generative models that learn from datasets of existing structures, reinforcement learning (RL) \citep{sutton1998reinforcement, arulkumaran2017deep} methods instead train an agent (i.e., a neural network) to optimize a given quantity or reward, without data. The model thus learns a behaviour policy by taking actions that act on a defined environment, like moving pieces (actions) on a chessboard (environment) to play a game, exploring actions that are likely to lead to high rewards (e.g., winning the game). This method has been used by \cite{zamaraeva2023reinforcement} with actions such as swapping atom positions and changing unit cell parameters, and the reward is defined as a prediction of the negative change in energy from that action. Challenges, however, lie in access to accurate, but inexpensive energy calculations, as well as efficient exploration strategies. The reinforcement learning approach was extended to crystal design in work by \cite{govindarajan2024learning}. 
In this work, actions consist of choosing new atoms to sequentially add to sites in a crystal.
An offline RL framework is used, which allows the use of a more expensive DFT-based method for computing rewards.

Generative Flow Networks (GFlowNets) \citep{bengio2021gflownet, bengio2023gflownetfoundations} offer a distinct set of methods, drawing inspiration from reinforcement learning \citep{tiapkin2024gfnmaxentrl, deleu2024gfnmaxentrl}. GFlowNets formulate generation as a sequential decision process, constructing molecules or materials step-by-step using policy network(s) acting on pre-defined environments.
In Crystal-GFN \citep{milaai4science2023crystalgfn}, a space group is first sampled, then the atomic composition, and finally the lattice parameters to form a crystal. It also encodes physico-chemical and symmetry constraints rigorously by masking impossible actions along the generation process. For materials discovery, the reward function typically corresponds to properties of interest such as formation energy, bandgap, or stress. Crystal-GFN demonstrated this capability by generating structures with low formation energy, targeted bandgaps, and high cell density, while other implementations have extended the approach to generate reticular frameworks for CO\textsubscript{2} capture \citep{cipcigan2024reticulargflownet}.

\subsection{Large Language Models and Multimodal ML Systems}

Large language models (LLMs) are increasingly being repurposed across the natural sciences as general-purpose priors for reasoning over sequences, graphs, and spatial data~\citep{zhang2024comprehensive}. This paradigm has recently been extended to materials generation. By representing crystal structures through textual descriptions of unit cells and atomic positions, language models trained to generate discrete tokens have shown competitive and sometimes superior performance compared to specialized geometric models \citep{gruver2024fine}. This unexpected effectiveness may stem from language models' efficiency as information compressors \citep{deletang2023language}, enabling sample-efficient learning in data-constrained materials domains.

Early explorations by \cite{flam2023language} demonstrated that decoder-based transformers trained on text sequences could generate molecules, crystals, and protein binding sites. Building on this foundation, CrystalLLM \citep{antunes2024crystal} fine-tuned language models on Crystallographic Information Files (CIF) for crystal structure generation. The approach was further refined by deCIFer \citep{johansen2025decifer} and \cite{gruver2024fine}; whereas deCIFer conditioned a decoder-only transformer on powder X-ray diffraction (PXRD) signals to generate crystal structures that agree with observed diffraction patterns in CIF format, \cite{gruver2024fine} fine-tuned LLaMA-2 \citep{touvron2023llama} on condensed crystal format descriptions that included unit cell parameters and fractional atomic coordinates. This achieved great generation performance while enabling flexible natural language conditioning, supporting unconditional generation, property-targeted generation, and structure completion.

Matra-Genoa~\citep{de2025generative} is another recent decoder-based autoregressive transformer model designed for inorganic crystal generation. By decomposing crystal structures into an interpretable token-based vocabulary, it efficiently generates structures conditioned on target properties such as composition, space group, and formation energies. Further, Mishra et al.\citep{mishra2024foundational} recently demonstrated that foundational materials domain LLMs (LLaMat) with additional training on semantic and syntactic tasks related to CIF (LLaMat-CIF) enable superior crystal generation with high diversity.

Domain knowledge integration has been explored in models like WyFormer \citep{pmlr-v267-kazeev25a} and Crystalformer \citep{cao2024space}, which incorporate crystallographic priors of space groups and Wyckoff positions \citep{hahn1983international, zhu2024wycryst}. Further improvements have come through reinforcement learning techniques such as direct preference optimization (DPO) \citep{rafailov2023direct}, demonstrated in PLaID \citep{xu2025plaid, xu2025plaid++}, which significantly improved stability and S.U.N. rates by learning from comparative evaluations of generated structures by machine learning interatomic potentials. Their work showcases the potential of leveraging reinforcement learning as a framework for optimizing future long-tail conditional generation tasks.

Multimodal approaches represent another frontier, attempting to integrate structural, spectroscopic, and textual information for more comprehensive materials modeling. Systems like MultiMat \citep{moro2025multimodal}, MoMa \citep{wang2025moma}, and LLM-Fusion \citep{boyar2025llm} demonstrate early efforts to combine representations across modalities, while agentic frameworks like A-Lab \citep{szymanski2023autonomous} point toward autonomous laboratories that can propose, execute, and refine materials synthesis guided by computational predictions.

\section{Evaluation Metrics for Materials Generation}
\label{app:sec:eval-metrics}

Unconditional generation refers to the task of producing valid, stable crystal structures without targeting specific properties or constraints. The following metrics assess the fundamental quality of generated structures:

\paragraph{Fundamental Validity Metrics.} These ensure the outputs are physically meaningful and chemically plausible. In different terms, they serve as a sanity check both for model development and inference time. Note that not all metrics may be relevant for every material system. 

\begin{itemize}
   \item \textbf{Charge Neutrality:} The total valence charge of all atoms must sum to zero:
   \begin{equation}
       \sum_{i=1}^{N} q_i = 0
   \end{equation}
   where $q_i$ is the nominal oxidation state of atom $i$ in the structure. For this to be calculated, the oxidation states of every atom in the structure must first be assigned. Here, we have developed a hierarchical structure for determining oxidation states and charge neutrality: 
   \begin{enumerate}
       \item If all atoms are metals, each atom is assigned a nominal oxidation state of zero, and the structure is labeled as charge balanced. This metallic assignment is determined by using the Pymatgen ``BVAnalyzer" class and the SMACT ``metallicity-score" function with a cutoff of 0.7 \citep{davies2019smact}. If either of these is satisfied, the structure is labeled as a metal and therefore labeled charge-balanced. 
       \item If not all atoms are metals, the Pymatgen ``get-oxi-state-decorated-structure" function \citep{ong2013python} is used to assign oxidation states and determine charge balance. 
       \item However, the function used above can fail to find oxidation states for structures that are not well optimized. It is still necessary to determine whether these structures are charge-balanced, particularly in the case of generative model benchmarks, when many structures may be too far from typical structures for the Pymatgen functions to analyze them. Here, we determine charge neutrality using a data-driven approach from LeMat-Bulk \citep{siron2025lemat}. First, this workflow determines all the possible charge-balanced compositions of oxidation states based on the observed oxidation states in LeMat-Bulk or the ICSD \citep{zagorac2019recent}. The most likely oxidation state assignments for this particular composition are determined by assigning each composition a score based on how probable that particular oxidation state configuration is, as determined by the distribution of oxidation states seen in LeMat-Bulk/ICSD. This score is determined by taking the geometric mean of all probabilities for each individual oxidation state. The geometric mean was chosen to normalize the score to the number of atoms in the structure, so scores between structures with two atoms and structures with ten atoms are comparable. Under the current implementation, if a composition exists from ICSD or LeMat-Bulk oxidation states that charge balances the structure, it passes the validity test. However, it is also possible to introduce a threshold here to filter out structures that require a combination of oxidation states that are highly uncommon, if a stricter validity metric is desired.

       \item If no charge-balanced composition can be made using oxidation states from LeMat-Bulk/ICSD, the structure is passed to the next stage of the validity assessment. In this stage, all historically observed oxidation states for each element, including highly irregular ones, are used to see if the structure can be charge-balanced. This list of oxidation states is drawn from Pymatgen's ``Element" class. Since this list includes some very irregular oxidation states (negative charges on metals, for example), an additional check is used to determine whether this charge-balanced structure is viable. Since we don't have access to probabilities for these very uncommon states, the oxidation state of each element in the system is correlated to the Pauli electronegativity of that element. A negative correlation is desired, as this would show that increasing electronegativity leads to a decreasing oxidation state. Any structures that show a positive correlation are labeled invalid, as this would reflect a very chemically implausible structure overall.   
   \end{enumerate}
   \item \textbf{Minimum Interatomic Distance:} All interatomic distances $d_{ij}$ must exceed a cutoff value $d_{\min}$ to prevent atomic overlap. We use $0.7$ \AA~ in this work.
   \begin{equation}
       d_{ij} > d_{\min} \quad \forall \, i \neq j
   \end{equation}

   \item \textbf{Mass density and atomic number density} should be within reasonable ranges. Mass density is given by $\rho = \frac{M_{\text{total}}}{V_{\text{cell}}}$, in  $(g/cm^3)$. The latter is expressed in atoms/\AA$^3$. We take upper bounds of 25 g/cm³ and 0.5 atoms/\AA$^3$, respectively. 
   
   \item \textbf{Valid crystallographic representation} is a good proxy to determine whether a structure is CIF-readable using \textit{pymatgen}. 

   \item \textbf{Lattice parameters} should be within reasonable ranges. We take upper bounds of 100 \AA~ for a,b,c, and 180 degrees for $\alpha$, $\beta$, $\gamma$, respectively, and lower bounds of 1 \AA~ and zero degrees for a,b,c and $\alpha$, $\beta$, $\gamma$, respectively.
   
\end{itemize}
\begin{figure}
\centering
\includegraphics[width=0.8\textwidth]{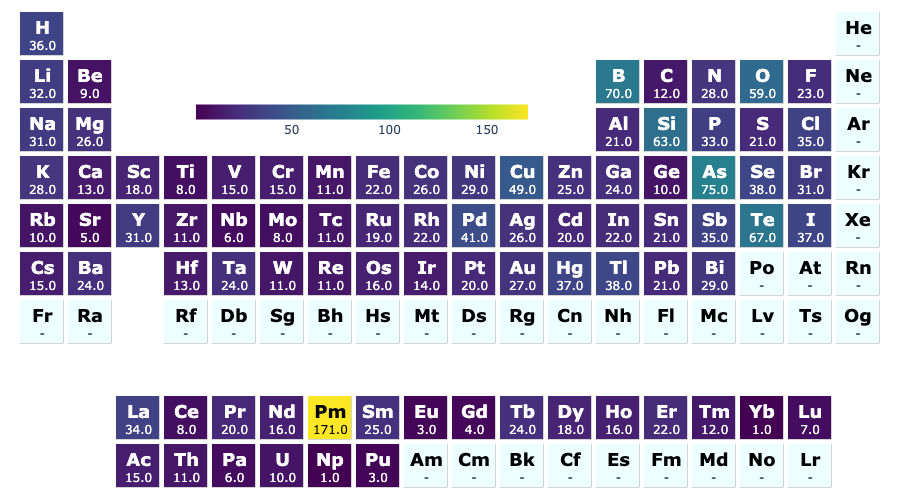}
\caption{The elemental composition of the structures labeled invalid by SMACT yet labeled valid by our model across the 5 random samples of 1000 structures drawn from LeMat-Bulk. Of these 5000 structures, 631 are labeled as valid by our model but invalid by SMACT. The counts indicate the number of times a structure containing that element is found in the overall list of structures. For example, 70 of the 631 structures contain boron.}
\label{fig:smact_invalidity}
\end{figure}
\begin{figure}[htbp]
    \centering
    \includegraphics[width=\textwidth]{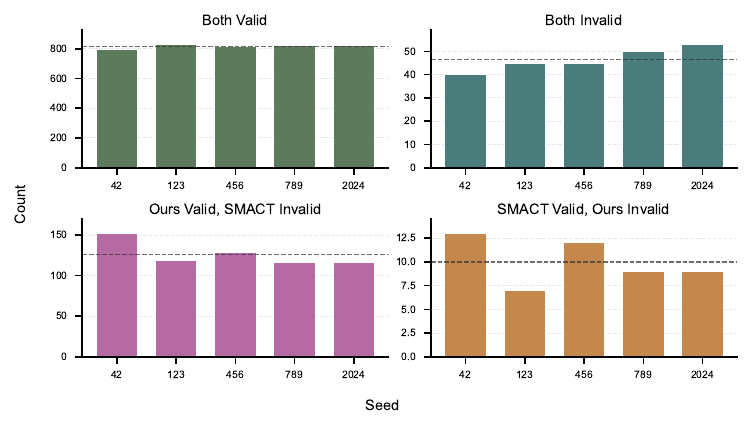}
    \caption{\textbf{Seed variation in validity outcomes.} 
Five independent samples of 1,000 structures each from LeMat-Bulk. 
Categories match the agreement breakdown from \Cref{fig:validity_comparison}. 
Dashed lines indicate mean values. High consistency across seeds demonstrates robustness 
of both validation methods, with our approach consistently validating $\sim$126 additional 
structures per seed while rejecting only $\sim$10 SMACT-approved structures.}
    \label{fig:seed_variation}
\end{figure}
\FloatBarrier
\paragraph{Stability metrics.} These assess the thermodynamic and energetic properties of generated structures:

   \begin{itemize}
   \item \textbf{Formation Energy ($E_f$):}
   \begin{equation}
       E_f = E_{\text{tot}}(\text{compound}) - \sum_{i} n_i \mu_i
   \end{equation}
   where $E_{\text{tot}}$ is the total energy of the crystal, $n_i$ is the number of atoms of element $i$, and $\mu_i$ is the chemical potential of the pure element. The result is normalized per atom: $E_f^{\text{per atom}} = \frac{E_f}{\sum_{i} n_i}$. We want it to be as small (and negative) as possible.

   The chemical potentials $\mu_i$ are derived from the LeMat-Bulk dataset using the same methodology as used in \cite{Ong2008Phase}. 

   \textbf{Multi-MLIP Ensemble Implementation:} The formation energy metric supports ensemble statistics across multiple MLIPs (ORB, MACE, UMA). For each structure, ensemble statistics are computed as:
   \begin{align}
       \langle E_f \rangle &= \frac{1}{N_{\text{MLIP}}} \sum_{k=1}^{N_{\text{MLIP}}} E_f^{(k)} \\
       \sigma_{E_f} &= \sqrt{\frac{1}{N_{\text{MLIP}}-1} \sum_{k=1}^{N_{\text{MLIP}}} \left(E_f^{(k)} - \langle E_f \rangle\right)^2}
   \end{align}
   where $E_f^{(k)}$ is the formation energy predicted by the $k$-th MLIP. The implementation extracts pre-computed ensemble statistics from structure properties (\texttt{formation\_energy\_mean}, \texttt{formation\_energy\_std}) or calculates them from individual MLIP results (\texttt{formation\_energy\_orb}, \texttt{formation\_energy\_mace}, etc.). A minimum of 2 MLIPs is required for ensemble statistics.
   \item \textbf{Energy Above Convex Hull ($E_{\text{hull}}$):}
   \begin{equation}
       E_{\text{hull}} = E_{\text{tot}} - E_{\text{hull}}^{\min}
   \end{equation}
   Structures with $E_{\text{hull}} \leq 0$ are considered stable, while values below approximately 0.1 eV/atom are often deemed metastable. We take LeMat-Bulk \citep{lematerial_2024} as reference point for calculating the convex hull.

   The convex hull is constructed by filtering the LeMat-Bulk dataset to include only compounds containing elements present in the target composition, creating \texttt{PDEntry} objects, and using Pymatgen's \texttt{PhaseDiagram.get\_decomp\_and\_e\_above\_hull()} method. The implementation handles charged species by extracting neutral elements before phase diagram construction. Multi-MLIP ensemble statistics follow the same formulation as formation energy: $\langle E_{\text{hull}} \rangle = \frac{1}{N_{\text{MLIP}}} \sum_{k=1}^{N_{\text{MLIP}}} E_{\text{hull}}^{(k)}$ with corresponding standard deviation calculations.

   \item \textbf{Relaxation Stability:} We use an ensemble of Machine Learning Interatomic Potentials to relax the generated structures (each one is done independently). Then, compute the Root Mean Square Deviation (RMSD) between pre- and post-relaxation atomic positions:
   \begin{equation}
       \text{RMSD} = \sqrt{ \frac{1}{N} \sum_{i=1}^{N} \| \mathbf{r}_i^{\text{init}} - \mathbf{r}_i^{\text{relax}} \|^2 }
   \end{equation}
   Low RMSD indicates minimal distortion and structural robustness under optimization.

   The implementation calculates individual RMSD values for each MLIP relaxation, then computes ensemble statistics: $\langle \text{RMSD} \rangle = \frac{1}{N_{\text{MLIP}}} \sum_{k=1}^{N_{\text{MLIP}}} \text{RMSD}^{(k)}$ where $\text{RMSD}^{(k)}$ is the relaxation RMSD from the $k$-th MLIP. The metric extracts pre-computed values from structure properties (\texttt{relaxation\_rmsd\_mean}, \texttt{relaxation\_rmsd\_std}) or calculates ensemble statistics from individual MLIP results (\texttt{relaxation\_rmsd\_orb}, \texttt{relaxation\_rmsd\_mace}, etc.). We use a force convergence threshold of $0.02$ eV\!/Å and a maximum of $50$ optimization steps for reporting RMSD in this work.
\end{itemize}

\paragraph{Novelty, Uniqueness, and Diversity Metrics.} These evaluate how effectively a model explores the chemical space:

\begin{itemize}
   \item \textbf{Novelty ($\mathcal{N}$):} Evaluates the fraction of generated structures that are not present in a reference dataset of known materials. The novelty score is defined as:
   \begin{equation}
       \mathcal{N} = \frac{|\{x \in G \mid x \notin T\}|}{|G|}
   \end{equation}
   where $G$ is the set of generated structures and $T$ is the reference dataset (LeMat-Bulk).
   
   The implementation supports two comparison methods: \textbf{BAWL fingerprinting} using crystallographic hash strings with the Weisfeiler-Lehman algorithm, and \textbf{structure matching} using Pymatgen's symmetry-aware structural comparison algorithms. For BAWL, novelty is determined by checking if the generated structure's fingerprint exists in the pre-computed reference fingerprint set. For structure matching, each generated structure is compared against reference structures with overlapping elemental compositions using space group analysis and atomic position matching with configurable tolerances. In our paper, we report results using \textit{pymatgen's StructureMatcher} approach for more robust structural comparison against the LeMat-Bulk reference dataset.
    \begin{figure}[htbp]
        \centering
        \includegraphics[width=1.0\linewidth]{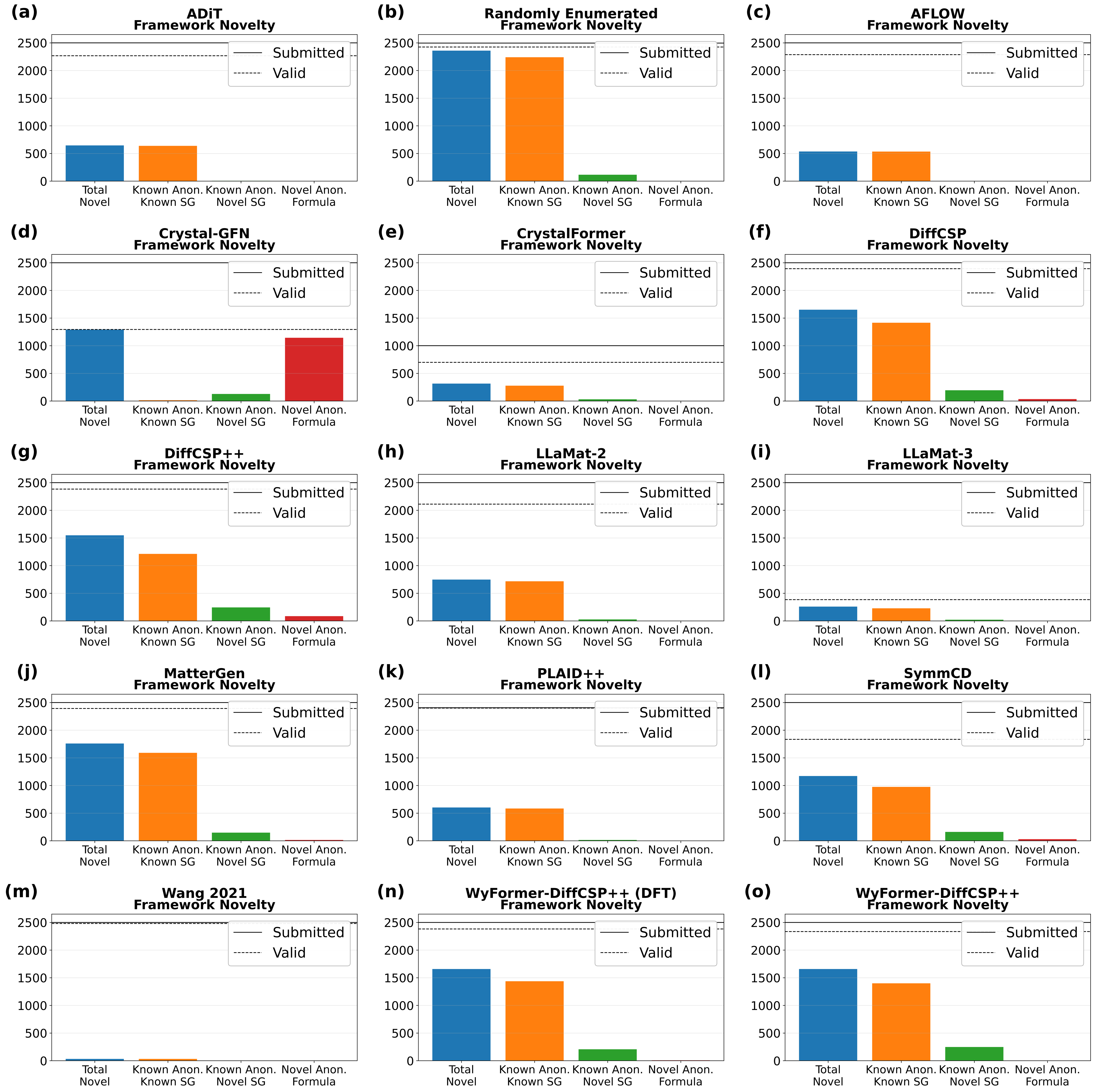}
        \caption{Breakdown of novelty by anonymous composition and space group across generative models, ML, non-ML and dataset baselines, (a) ADiT \citep{joshi2025all}, (b) Random Enumeration \citep{szymanski2025establishing}, (c) AFLOW \citep{curtarolo2012aflow}, (d) Crystal-GFN \citep{milaai4science2023crystalgfn}, (e) CrystalFormer \citep{cao2025crystalformer}, (f) DiffCSP \citep{jiao2023crystal}, (g) DiffCSP++ \citep{jiao2024space}, (h) LLaMat-2 \citep{mishra2024foundational}, (i) LLaMat-3 \citep{mishra2024foundational}, (j) MatterGen \citep{zeni2023mattergen}, (k) PLaID++ \citep{xu2025plaid++}, (l) SymmCD \citep{levysymmcd}, (m) Wang et al. \citep{wang2021predicting}, (n) Wyformer-DiffCSP++ (DFT) \citep{kazeev2025wyckoff}, and (o) Wyformer-DiffCSP++ \citep{kazeev2025wyckoff}. We define three additional classes of novelty, based on the anonymous composition (the stoichiometry with element identities removed, e.g., ABC$_3$) and space group of an input crystal.
\textit{Known anonymous composition and space group}: the generated structure's anonymous composition is present in our LeMat-Bulk reference dataset, and its space group has already been observed for that anonymous composition.
\textit{Known anonymous composition but novel space group}: the generated structure's anonymous composition is present in LeMat-Bulk, but its space group has not previously been observed for that anonymous composition.
\textit{Novel anonymous composition}: the generated structure's anonymous composition is absent from LeMat-Bulk.
These classes are coarse proxies and do not identify structural prototypes: structures belonging to the same prototype can differ in lattice parameters or site occupancies, and distinct prototypes can share the same anonymous composition and space group. A known anonymous composition and space group therefore does not imply a known framework.}
        \label{fig:frame_breakdown}
    \end{figure}

    \begin{figure}[htbp]
        \centering
        \includegraphics[width=1.0\linewidth]{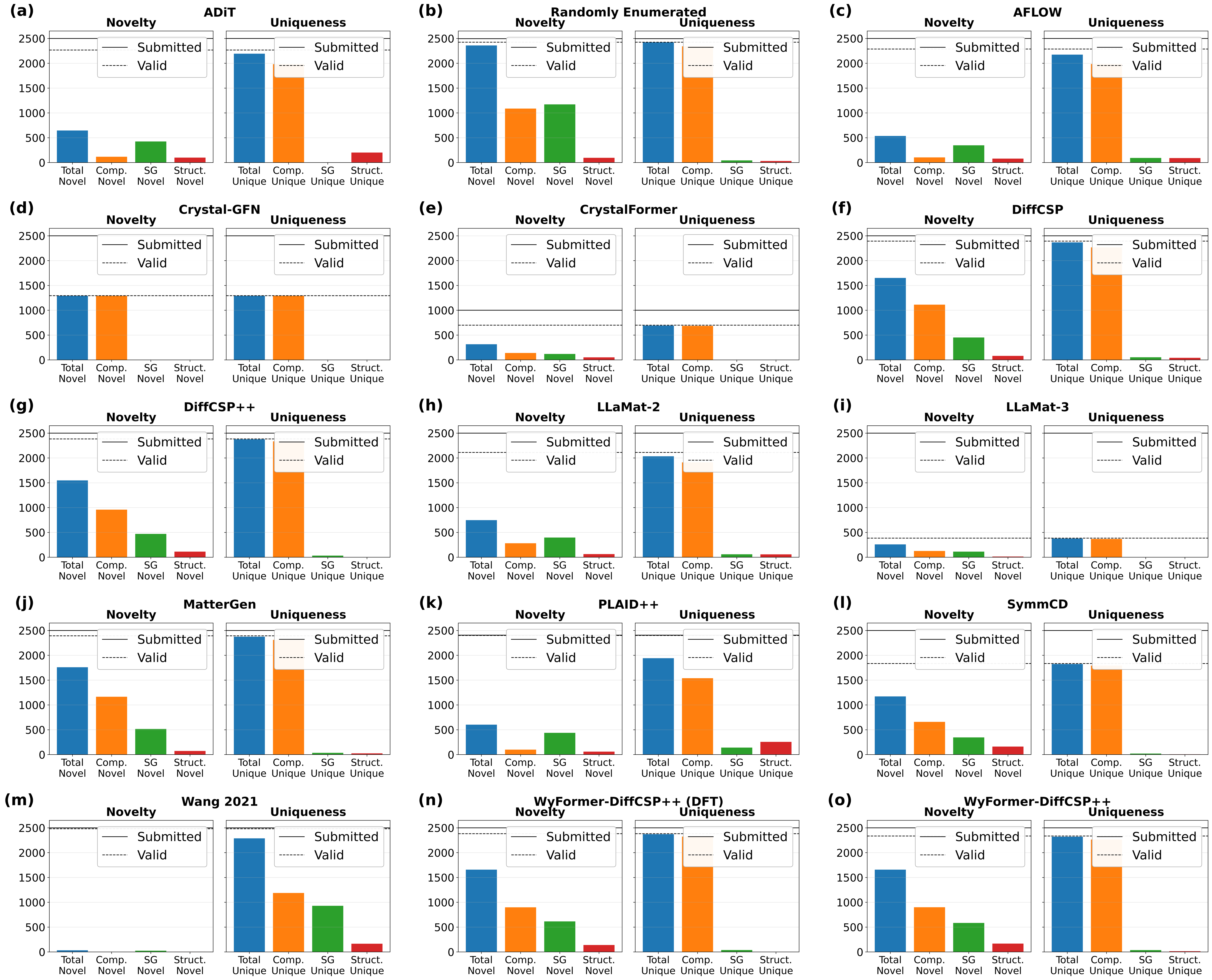}
        \caption{Breakdown of novelty and uniqueness across generative models, ML, non-ML and dataset baselines, (a) ADiT \citep{joshi2025all}, (b) Random Enumeration \citep{szymanski2025establishing}, (c) AFLOW \citep{curtarolo2012aflow}, (d) Crystal-GFN \citep{milaai4science2023crystalgfn}, (e) CrystalFormer \citep{cao2025crystalformer}, (f) DiffCSP \citep{jiao2023crystal}, (g) DiffCSP++ \citep{jiao2024space}, (h) LLaMat-2 \citep{mishra2024foundational}, (i) LLaMat-3 \citep{mishra2024foundational}, (j) MatterGen \citep{zeni2023mattergen}, (k) PLaID++ \citep{xu2025plaid++}, (l) SymmCD \citep{levysymmcd}, (m) Wang et al. \citep{wang2021predicting}, (n) Wyformer-DiffCSP++ (DFT) \citep{kazeev2025wyckoff}, and (o) Wyformer-DiffCSP++ \citep{kazeev2025wyckoff}. This figure decomposes the top-line novelty and uniqueness scores into three mutually exclusive components. Compositional novelty and uniqueness (Comp. Novel / Comp. Unique) denote the number of structures with stoichiometries that are either absent from LeMat-Bulk or appear only once within the generated sample, respectively. Space group metrics (SG Novel / SG Unique) denote the number of structures that share a composition with LeMat-Bulk (or the sample) but have a previously unobserved or non-duplicated space group symmetry for that specific stoichiometry. Geometric metrics (Geom. Novel / Geom. Unique) represent the number of structures that share both composition and space group with LeMat-Bulk (or another structure in the sample), yet are not matched by \texttt{StructureMatcher} under the tolerances used in this work, owing to differences in lattice parameters or atomic coordinates (e.g., a triclinic structure with identical stoichiometry but different lattice values). This category reflects geometric novelty under the chosen matching tolerances and does not imply a new structural framework or prototype. Dashed and dotted lines indicate the total number of submitted and valid structures, respectively.}
        \label{fig:nu_breakdown}
    \end{figure}

   \item \textbf{Uniqueness ($\mathcal{U}$):} Measures the fraction of unique structures within the generated set to assess internal diversity. The uniqueness score is defined as:
   \begin{equation}
       \mathcal{U} = \frac{|\text{unique}(G)|}{|G|}
   \end{equation}
   where $\text{unique}(G)$ returns the set of unique structures based on their fingerprints.
   
   The metric is implemented as a structure-level continuous scoring system rather than a binary classification. For BAWL fingerprinting, individual uniqueness scores are assigned as $u_i = 1/c_i$, where $c_i$ is the count of structures sharing the same fingerprint within the generated set. This assigns a score of 1.0 to truly unique structures while proportionally penalizing duplicated structures. For structure matching, the implementation uses pairwise comparison with an ordered approach: structure $i$ is considered unique if it is not equivalent to any structure $j$ where $j < i$, ensuring deterministic selection of the first occurrence as the unique representative. The overall uniqueness metric is computed as $\mathcal{U} = \frac{1}{|G|}\sum_{i=1}^{|G|} u_i$. Both BAWL fingerprinting and structure matching methods are supported, with structure matching used for reporting results.

   \item \textbf{S.U.N. and M.S.U.N. Rates:} Proportion of generated structures that are simultaneously Stable (or Metastable), Unique, and Novel:
   \begin{equation}
\text{S.U.N. Rate} = \frac{|\text{novel}(\text{unique}({x \in G \mid E_{\text{hull}}(x) \leq 0}))|}{|G|}
\end{equation}
\begin{equation}
\text{M.S.U.N. Rate} = \frac{|\text{novel}(\text{unique}({x \in G \mid 0 < E_{\text{hull}}(x) \leq \tau}))|}{|G|}
\end{equation}
   
   The implementation follows a hierarchical computation order: \textbf{Stability} $\rightarrow$ \textbf{Uniqueness} $\rightarrow$ \textbf{Novelty}. First, structures are classified as stable ($E_{\text{hull}} \leq 0$) or metastable ($0 < E_{\text{hull}} \leq \tau$) using energy above hull values computed by the Multi-MLIP stability preprocessor. Then, uniqueness is evaluated within each stability class separately using the chosen comparison method. Finally, novelty is assessed for unique structures from each stability class. This hierarchical approach provides detailed metrics at each evaluation stage: stability counts, unique-within-stable/metastable counts, and final SUN/MSUN counts. The Multi-MLIP preprocessor assigns ensemble stability properties (e.g., \texttt{e\_above\_hull\_mean}) to structure objects, enabling robust stability classification across multiple MLIPs (ORB, MACE, UMA). We set $\tau$ to \textbf{0.1 eV/atom} for assembling results. 
   
   \item \textbf{Diversity:} We assess diversity by comparing the distribution of generated structures
against reference datasets across key crystallographic and compositional attributes. This includes
space groups, elemental compositions, lattice parameters, and atomic environments. In addition to
distribution plots, we compute feature-level entropy metrics to quantify the spread of generated
samples.

\begin{itemize}
    \item \textbf{Composition, Space Group, Lattice, and Atomic Site Entropy:}
    Given a set of $N$ generated structures, we first compute the frequency $f_i$ of each category
    (e.g., element $i$ for composition, space group index for symmetry, discretized lattice bins,
    or Wyckoff site labels). These frequencies are normalized into a probability distribution
    $p_i = \frac{f_i}{\sum_j f_j}$. We then compute the Shannon entropy
    \[
        H = - \sum_i p_i \log p_i,
    \]
    together with the Vendi score~\citep{friedman2022vendi}, defined as the exponential of the
    Shannon entropy. Although illustrated here for composition entropy, the same procedure
    is applied to the other diversity attributes in our benchmark.
   \end{itemize}
   
\end{itemize}

\begin{figure}[htbp]
    \centering
    \includegraphics[width=0.8\linewidth]{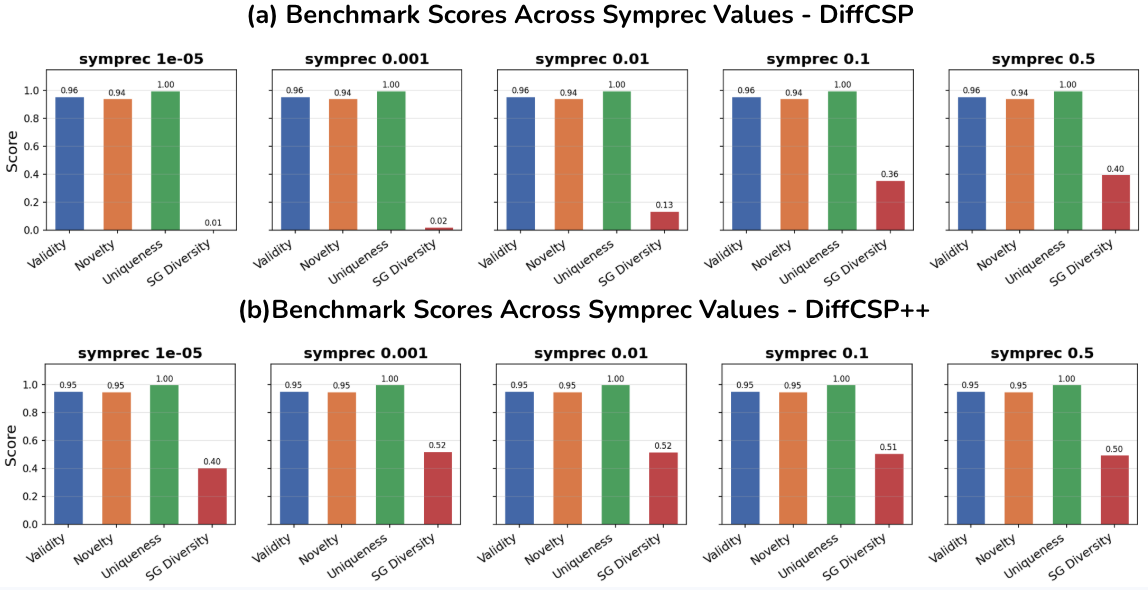}
    \caption{Impact of symmetry precision (symprec) on select benchmarks dependent on space group identification. Benchmark scores for (a) DiffCSP and (b) DiffCSP++ evaluated across symprec values from 1e-5 to 0.5 Å. As shown, varying the symmetry tolerance has no discernible impact on the validity, novelty, or uniqueness values of the generated structures. However, there is a positive correlation between symprec and space group (SG) diversity, which is particularly evident in the DiffCSP model, where the score drops from 0.40 at an extremely loose tolerance down to near-zero (0.01) at the strictest value. This implies that the generated samples contain approximate high-symmetry features, but their precise atomic positions deviate enough from ideal sites that the fixed value of 0.01 used in this work, the established default from Pymatgen \cite{jain2013materials}, drops them into lower symmetry classifications.}
    \label{fig:symprec_sweep}
\end{figure}

\paragraph{Distribution-Level Metrics.} To evaluate how well the distribution of generated structures
matches that of real materials, we use the following metrics:

\begin{itemize}

  \item \textbf{Jensen--Shannon Distance} \citep{fuglede2004jensen}:
   \begin{equation}
       \text{JSD}(P, Q) =
       \sqrt{\frac{1}{2} D_{KL}(P \,\|\, M) +
             \frac{1}{2} D_{KL}(Q \,\|\, M)}
   \end{equation}
   where $P$ and $Q$ are the distributions of generated and real samples, respectively,
   $M = \tfrac{1}{2}(P + Q)$ is their mixture distribution, and $D_{KL}$ denotes the
   Kullback--Leibler divergence.

  \item \textbf{Maximum Mean Discrepancy (MMD)} \citep{tolstikhin2016minimax}:
   \begin{equation}
       \text{MMD}^2(P, Q) =
       \mathbb{E}_{x, x'}[k(x, x')] +
       \mathbb{E}_{y, y'}[k(y, y')] -
       2\,\mathbb{E}_{x, y}[k(x, y)]
   \end{equation}
   where $P$ and $Q$ are distributions of generated and real samples, and $k$ is a kernel
   function.

  \item \textbf{Fréchet Distance Metrics} \citep{heusel2017gans, preuer2018frechet}:
   Variants such as the Fréchet ChemNet Distance (FCD) compare the distributions of generated
   and reference structures:
   \begin{equation}
       \text{FD}(G, T) =
       \| \mu_G - \mu_T \|^2 +
       \text{Tr}\!\left(
           \Sigma_G + \Sigma_T -
           2\,(\Sigma_G \Sigma_T)^{1/2}
       \right)
   \end{equation}
   where $\mu$ and $\Sigma$ denote the mean and covariance in embedding space.

\end{itemize}

\paragraph{Model Efficiency.} This measures how effectively a model learns from limited training data \citep{gao2022sample}:
\begin{itemize}
   \item \textbf{Generic Metrics}: training dataset size, number of model parameters, number of epochs required for training, training time and associated computational infrastructure, inference time on 2500 structures. 
   \item \textbf{Learning Curve Analysis:} Performance (e.g., S.U.N. rate, property prediction accuracy) as a function of the number of expensive function evaluations (e.g., DFT calculations) required for training, i.e., the number of labeled data points.
\end{itemize}

\paragraph{Herfindahl-Hirschman Index (HHI) Metrics.} 
\label{par:hhi}
The Herfindahl-Hirschman index quantifies supply risk concentration for materials by measuring the concentration of element production sources and reserves. The elemental data for calculating HHI metrics are derived from the \href{https://sparks.mse.utah.edu/about.html}{calculator} made available by  Mansouri et al.\cite{mansouri2017balancing}. For a given crystal structure with composition, we compute:

\begin{itemize}
  \item \textbf{Compound HHI Value}: For a compound with chemical formula represented by composition $C$:
   \begin{equation}
       \text{HHI}_{\text{compound}} = \sum_{i} x_i \cdot \text{HHI}_i
   \end{equation}
   where $x_i$ is the fractional composition of element $i$ in the compound, and $\text{HHI}_i$ is the element-specific HHI value.
   
   \item \textbf{Production HHI}: Measures supply risk based on the concentration of element production sources (market concentration):
   \begin{equation}
       \text{HHI}_{\text{production}} = \sum_{j} s_{j}^2 \times 10000
   \end{equation}
   where $s_j$ is the market share of producer $j$ for a given element.
   
   \item \textbf{Reserve HHI}: Measures long-term supply risk based on the concentration of element reserves (geographic distribution):
   \begin{equation}
       \text{HHI}_{\text{reserve}} = \sum_{k} r_{k}^2 \times 10000
   \end{equation}
   where $r_k$ is the fraction of global reserves held by country/region $k$.
   
   \item \textbf{Scaling Convention}: HHI values are typically scaled from the classical range [0, 10000] to a convenience range [0, 10]:
   \begin{equation}
       \text{HHI}_{\text{scaled}} = \frac{\text{HHI}_{\text{classical}}}{1000}
   \end{equation}
   
   \item \textbf{Combined HHI Score}: The final benchmark score combines both production and reserve metrics using weighted averaging:
   \begin{equation}
       \text{HHI}_{\text{combined}} = w_{\text{prod}} \cdot \text{HHI}_{\text{production}} + w_{\text{res}} \cdot \text{HHI}_{\text{reserve}}
   \end{equation}
   where $w_{\text{prod}} = 0.25$ and $w_{\text{res}} = 0.75$ by default, prioritizing long-term supply security over short-term market dynamics.
   
   \item \textbf{Missing Element Handling}: Elements not found in the HHI lookup tables are assigned the maximum risk value (10000 unscaled / 10 scaled) to represent maximum supply uncertainty for rare or untracked elements.
   
   \item \textbf{Risk Categories}: For the scaled [0, 10] range:
   \begin{align}
       \text{Low Risk} &: \text{HHI}_{\text{scaled}} \leq 2.0 \\
       \text{Moderate Risk} &: 2.0 < \text{HHI}_{\text{scaled}} \leq 5.0 \\
       \text{High Risk} &: \text{HHI}_{\text{scaled}} > 5.0
   \end{align}
\end{itemize}

\subsection{MLIP Checkpoints and Reference Data}
\label{app:sec:mlip-checkpoints}

Energy-, stability-, and embedding-based metrics depend on the specific MLIP checkpoints used, since checkpoints within the same model family are trained on different datasets and therefore yield different energy scales and representations. For reproducibility, the checkpoints used for every result in this work are listed below, together with the corresponding package versions. All checkpoints are pinned explicitly in the \lematgen codebase (\url{https://github.com/LeMaterial/lemat-genbench}).

\begin{center}
\small
\begin{tabular}{p{0.30\textwidth} l p{0.46\textwidth}}
\toprule
\textbf{Use} & \textbf{MLIP} & \textbf{Checkpoint} \\
\midrule
\multirow{3}{0.30\textwidth}{Energies, relaxation, $E_f$, $E_{\text{hull}}$, and convex-hull reference energies}
 & MACE-MP & MACE-MP-0b3-medium, \texttt{medium-0b3} \\
 & UMA & \texttt{uma-s-1}, \texttt{omat} task head \\
 & Orb-v3 & \texttt{orb\_v3\_conservative\_inf\_omat} \\
\midrule
\multirow{3}{0.30\textwidth}{Embeddings (Fr\'echet distance, t-SNE)}
 & MACE-MP & MACE-MPA-0-medium, \texttt{medium-mpa-0} (640-dim.) \\
 & UMA & \texttt{uma-s-1}, \texttt{omat} task head (128-dim.) \\
 & Orb-v3 & \texttt{orb\_v3\_conservative\_inf\_omat} (256-dim.) \\
\midrule
Post-relaxation results (\Cref{sec:app:benchmark_results}) & UMA & \texttt{uma-s-1}, \texttt{omat} task head \\
\bottomrule
\end{tabular}
\end{center}

\paragraph{Self-consistent convex hull.} For each MLIP, the total energies of all LeMat-Bulk structures were pre-computed with the same energy checkpoint that is used to evaluate generated structures, so that candidate energies and the reference hull share the same energy scale. These energies, together with the resulting hull reference sets, are publicly available at \url{https://huggingface.co/datasets/LeMaterial/LeMat-Bulk-MLIP-Hull}. The energies predicted by MACE-MP-0b3-medium, \texttt{uma-s-1}, and \texttt{orb\_v3\_conservative\_inf\_omat} are stored in the columns \texttt{mace\_mp\_energy}, \texttt{uma\_energy}, and \texttt{orb\_conserv\_inf\_energy}, respectively, and the corresponding hull reference sets (structures within 0.001~eV/atom of each MLIP's hull) are provided as the \texttt{mace\_mp}, \texttt{uma}, and \texttt{orb\_conserv\_inf} subsets.

\paragraph{Embedding-based metrics.} Reference statistics (mean and covariance) for the Fr\'echet distance are computed from graph embeddings of LeMat-Bulk structures, available at \url{https://huggingface.co/datasets/LeMaterial/LeMat-GenBench-embeddings}. For MACE, the reference embeddings were computed with MACE-MPA-0-medium, which therefore differs from the checkpoint used for energies; embeddings of generated structures are computed with the same checkpoint as the reference. 

\section{Benchmark Results}
\label{sec:app:benchmark_results}
\begin{table}[htbp]
\centering
\scriptsize
\caption{Training datasets and data sources used for the reported generative crystal structure models. Figshare of Kazeev et al. \citep{kazeev2025wyckoff} is here: \url{https://figshare.com/articles/dataset/Generated_crystals_for_WyFormer_DiffCSP_DiffCSP_WyCryst_SymmCD_CrystalFormer_MiAD/29145101}}
\label{table:source-of-data}
\begin{tabular}{lcc}
\toprule
\textbf{Model} & \textbf{Training Dataset} & \textbf{Source of Submitted Structures} \\
\midrule
ADiT \citep{joshi2025all} & MP-20 & Authors \\ %
Crystal-GFN \citep{ai4science2023crystal} & MP-20 & Authors \\
Crystalformer\citep{cao2024space} & MP-20 & Figshare of Kazeev et al. \\ %
DiffCSP \citep{jiao2023crystal} & MP-20 & Figshare of Kazeev et al. \\ %
DiffCSP++ \citep{jiao2024space} & MP-20 & Figshare of Kazeev et al. \\ %
LLaMat2-CIF \citep{mishra2024foundational} & MP-20 & Authors \\ %
LLaMat3-CIF \citep{mishra2024foundational} & MP-20 & Authors \\ 
MatterGen \citep{zeni2025generative} & MP-20 & Figshare of Kazeev et al. \\
PLaID++ \citep{xu2025plaid} & MP-20 & Authors \\ %
SymmCD \citep{levysymmcd} & MP-20 & Figshare of Kazeev et al. \\
WyFormer \citep{kazeev2025wyckoff} & MP-20 & Authors \\
WyFormer-DFT \citep{kazeev2025wyckoff} & MP-20 & Authors\\
Alexandria \citep{schmidt2021crystal, schmidt2024improving} & --- & Sampled from source database \\
OQMD \citep{saal2013materials, kirklin2015open} & --- & Sampled from source database \\
AFLOW \citep{curtarolo2012aflow} & --- & Sampled from source database \\
MP-Exp \citep{jain2013materials} & --- & Sampled from source database \\
IonSub \citep{wang2021predicting} & --- & \url{https://doi.org/10.24435/materialscloud:96-09} \\
RandomEnum \citep{szymanski2025establishing} & --- & Generated on-the-fly \\
\bottomrule
\end{tabular}
\end{table}

As summarized in \Cref{table:source-of-data}, since most models included in this study were trained on MP-20, we report 
results using the multi-MLIP ensemble benchmark on both the LeMat-Bulk and MP\textendash20 datasets as
references. Evaluating on both reference sets ensures consistency between the training 
distribution and the evaluation database, allowing authors to better interpret model behavior 
without being penalized for discrepancies arising solely from dataset choice. 

However, LeMat-Bulk is currently the most comprehensive and diverse collection of inorganic 
crystal structures. Performance on this dataset, therefore, provides the most informative assessment 
of generative modeling capability and serves as the primary reference for the public leaderboard. 
Metrics that depend on a reference dataset are expected to vary between the two settings, including 
novelty scores, energy-based metrics, stability and metastability rates, and embedding-based 
distributional metrics such as the Fr\'{e}chet distance.

Several generative models considered here perform structure relaxation as part of their generation 
pipeline (MatterGen, PLaID++, WyFormer, and WyFormer\textendash DFT). To provide a fair and 
transparent comparison, we report two sets of results for all models and baselines: (i) metrics 
computed directly on the submitted structures, and (ii) metrics computed after relaxing the structures 
using the UMA \texttt{uma-s-1} potential (\texttt{omat} task head; see \Cref{app:sec:mlip-checkpoints}). In this setting, only \texttt{uma-s-1} is used for evaluating
formation energies and constructing the convex hull. Using the full multi\textendash MLIP ensemble 
would not be meaningful, since one of the potentials is already used for relaxation and the goal is 
to measure geometric and energetic quality in a self\textendash consistent manner.

For all relaxation procedures, we use the default configuration of our relaxation pipeline:
a force convergence threshold of $0.02$ eV\!/Å and a maximum of $500$ optimization steps (the RMSD check in the pre-relaxation results uses at most
$50$ steps; see \Cref{app:sec:eval-metrics}). These 
settings correspond to the default parameters of the relaxation module:
\begin{verbatim}
--fmax 0.02
--steps 500
\end{verbatim}
This ensures that all post\textendash relaxation metrics are evaluated under a uniform and reproducible 
protocol across models and datasets.

\subsection{Model Training Details}

\begin{table}[htbp]
\centering
\caption{Training and inference times for 2{,}500 samples for the reported generative models, with respective compute information.} \scriptsize
\begin{adjustbox}{max width=\textwidth}
\begin{tabular}{l c c c c}
\toprule
\textbf{Model} & \textbf{Training Time (hrs)} & \textbf{Train GPU} & \textbf{Inference Time (s)} & \textbf{Inference GPU} \\
\midrule
SymmCD        & 26    & V100                 & 1{,}400 & RTX8000 \\
Crystalformer & 13    & A100                 & 100     & V100 \\
ADiT          & 576   & V100                 & 300     & V100 \\
PLaID++       & 10    & H100                 & 345     & H100 \\
MatterGen     & 80    & A100                 & 18{,}000 & V100 \\
DiffCSP       & \textendash & \textendash    & 2{,}335 & A40 \\
DiffCSP++     & 19.5  & RTX 6000             & 6{,}150 & A40 \\
WyFormer      & 11.5  & RTX 6000             & 0.625   & RTX6000 \\
Crystal-GFN   & 12    & CPU                  & 3000 & CPU \\
LLaMat2-CIF   & 3264  & A100                 & 3897 & A100 \\
LLaMat3-CIF   & 144   & CS2 Cerebras  & 4235 & A100 \\
\bottomrule
\end{tabular}
\end{adjustbox}
\end{table}
\FloatBarrier

\subsection{Benchmark Results with LeMat-Bulk as the reference database}
\label{app:subsec:lemat_bulk_results}

\begin{figure}[htbp]
\centering
\begin{subfigure}[b]{0.48\textwidth}
    \centering
    \includegraphics[width=\textwidth, height=0.7\textwidth, keepaspectratio=true]{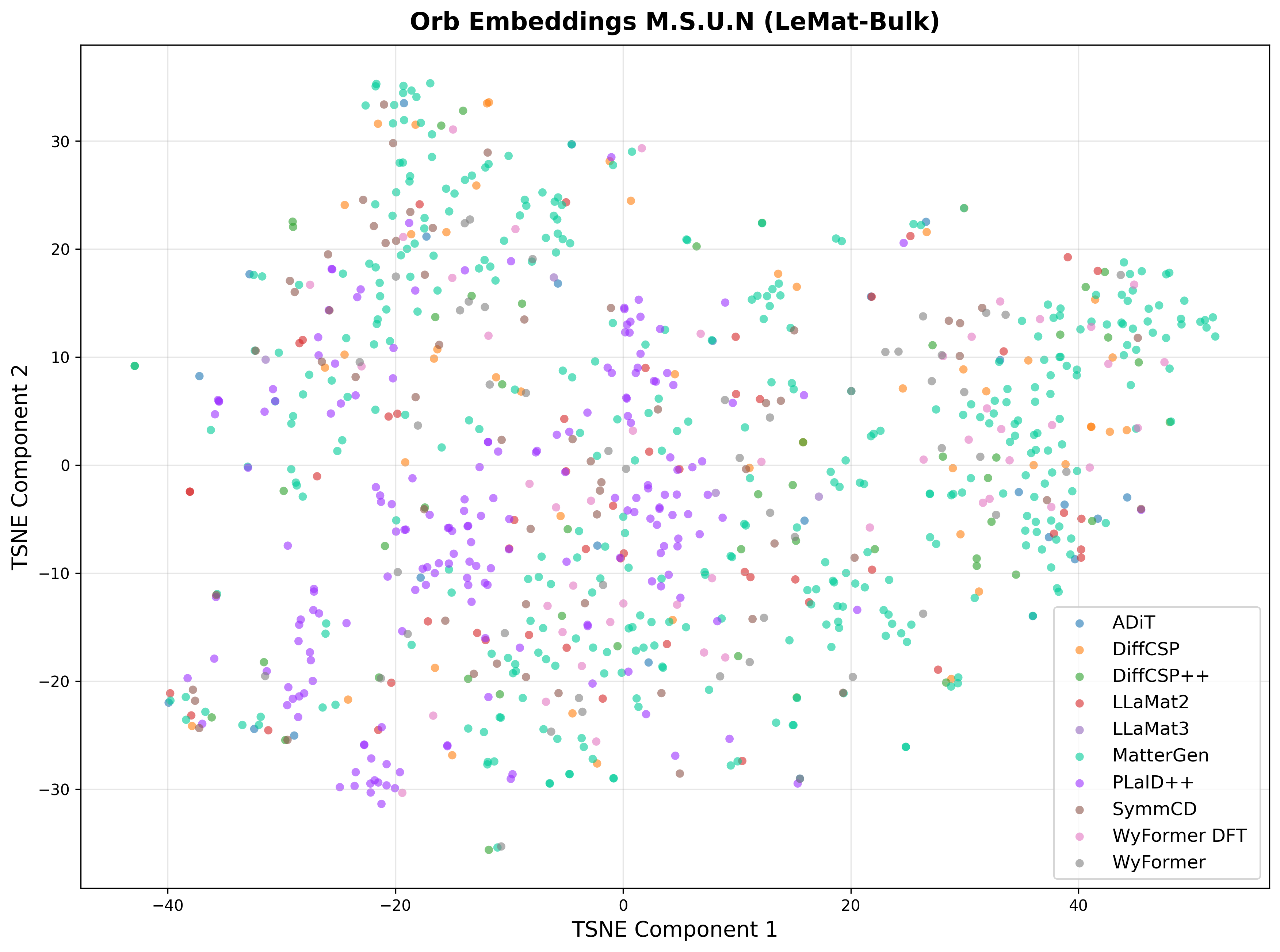}
    \caption{t-SNE visualization of ORB graph embeddings for M.S.U.N. materials with LeMat-Bulk as the reference set.}
    \label{fig:lemat_tsne_msun}
\end{subfigure}
\hfill
\begin{subfigure}[b]{0.48\textwidth}
    \centering
    \includegraphics[width=\textwidth, height=0.7\textwidth, keepaspectratio=true]{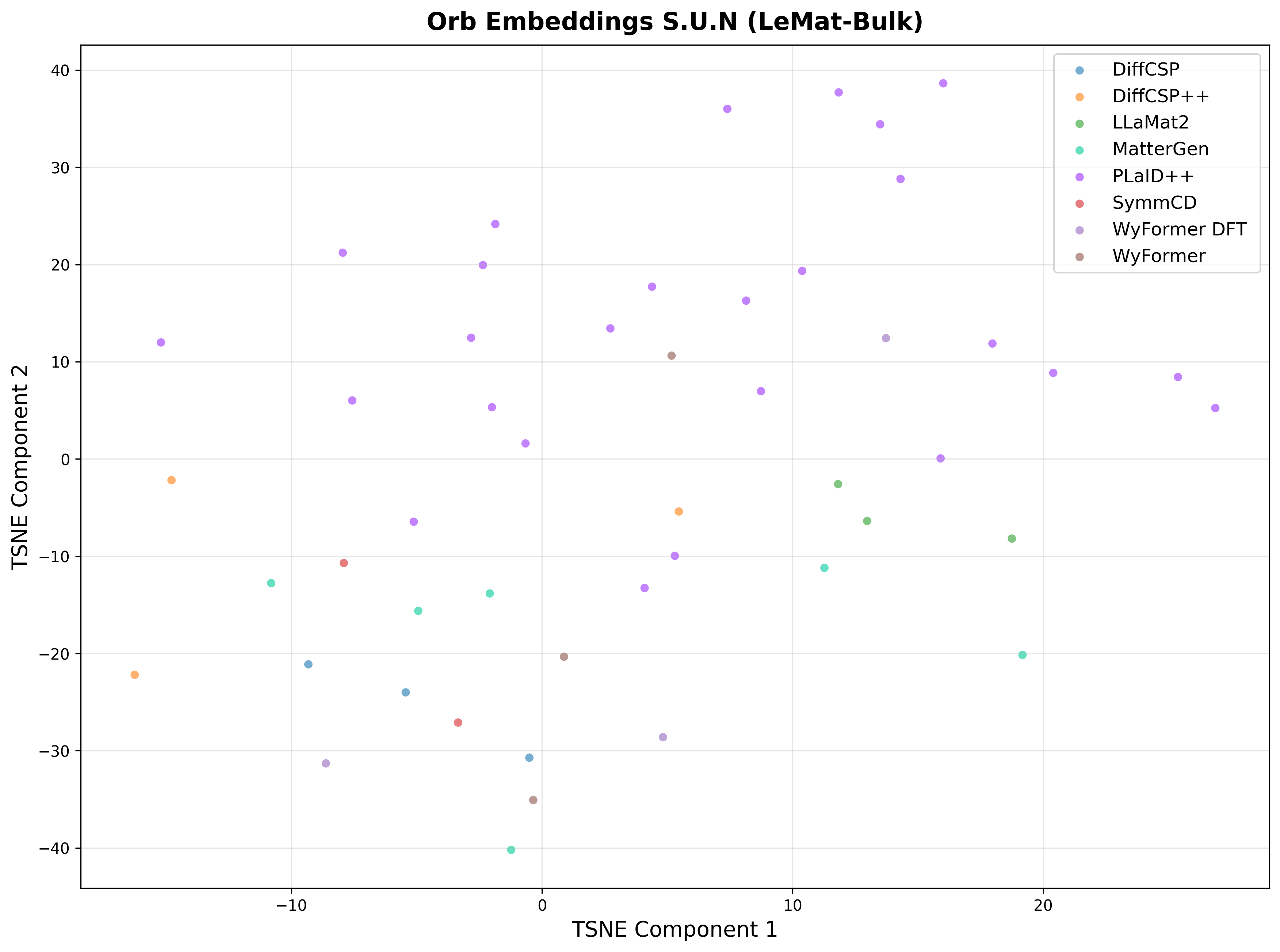}
    \caption{t-SNE visualization of ORB graph embeddings for S.U.N. materials with LeMat-Bulk as the reference set.}
    \label{fig:lemat_tsne_sun}
\end{subfigure}
\caption{Two-dimensional t-SNE visualizations (perplexity = 30) of ORB graph embeddings of generative model structures computed with LeMat-Bulk as the reference set. The visualizations demonstrate how MatterGen, DiffCSP, and PLaID++ generate a wide diversity of novel crystal structures that cover a breadth of chemical space.}
\label{fig:lemat_tsne}
\end{figure}

We visualize S.U.N. and M.S.U.N. structures with respect to LeMat-Bulk from various generative models using t-SNE, a dimensionality reduction method in \Cref{fig:lemat_tsne}. The resulting plots show that most models produce samples that are broadly dispersed throughout crystallographic space, rather than collapsing into small or isolated clusters. This indicates that many contemporary crystal generators are capable of proposing chemically diverse structures. One standout model in these plots is MatterGen, whose samples appear to have the highest coverage of ORB's embedding space.

\begin{figure}[htbp]
\centering
\begin{subfigure}[b]{0.48\textwidth}
    \centering
    \includegraphics[width=\textwidth, height=0.7\textwidth, keepaspectratio=true]{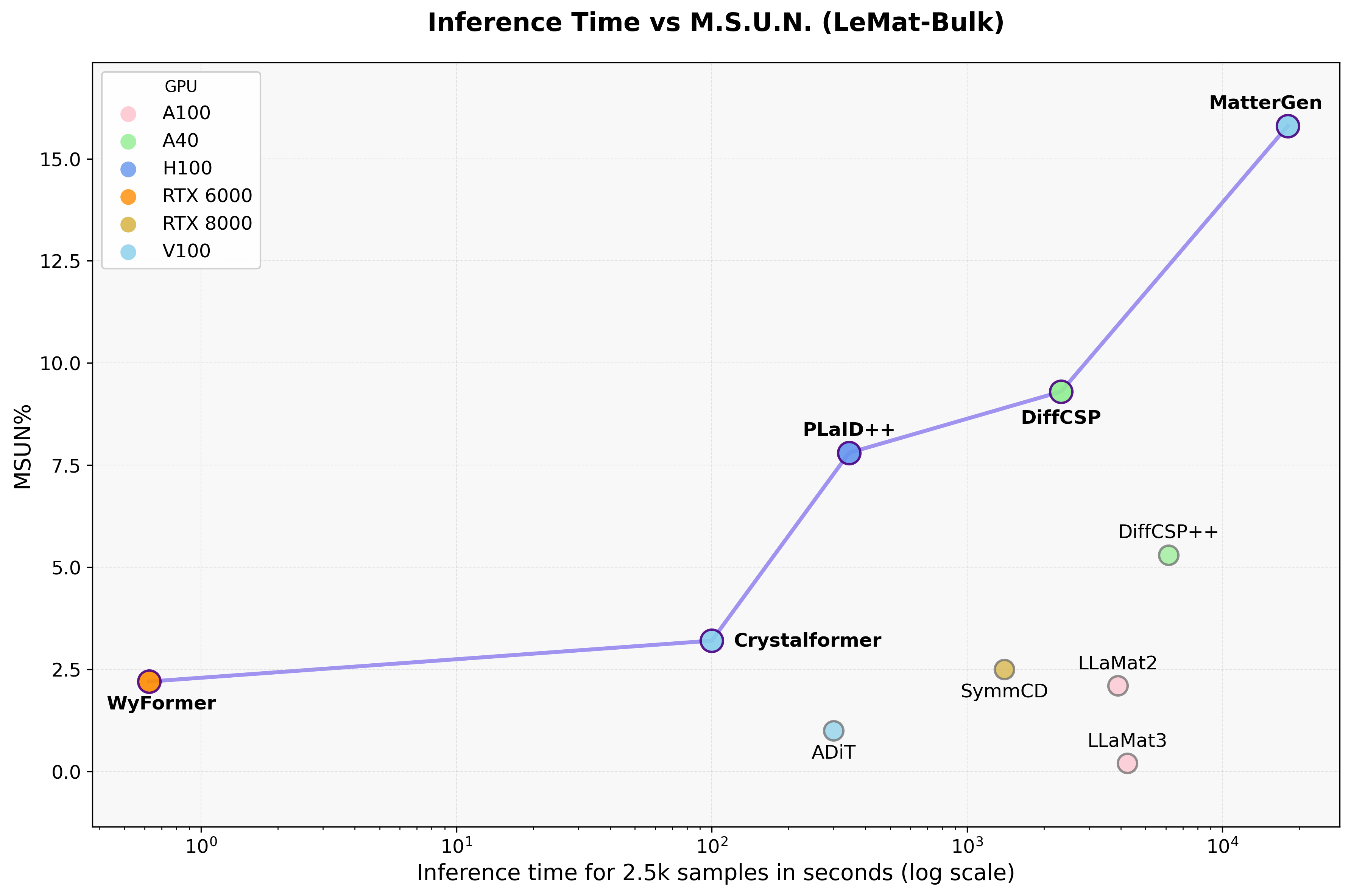}
    \caption{Inference time vs MSUN\%}
    \label{fig:inference_time_msun}
\end{subfigure}
\hfill
\begin{subfigure}[b]{0.48\textwidth}
    \centering
    \includegraphics[width=\textwidth, height=0.7\textwidth, keepaspectratio=true]{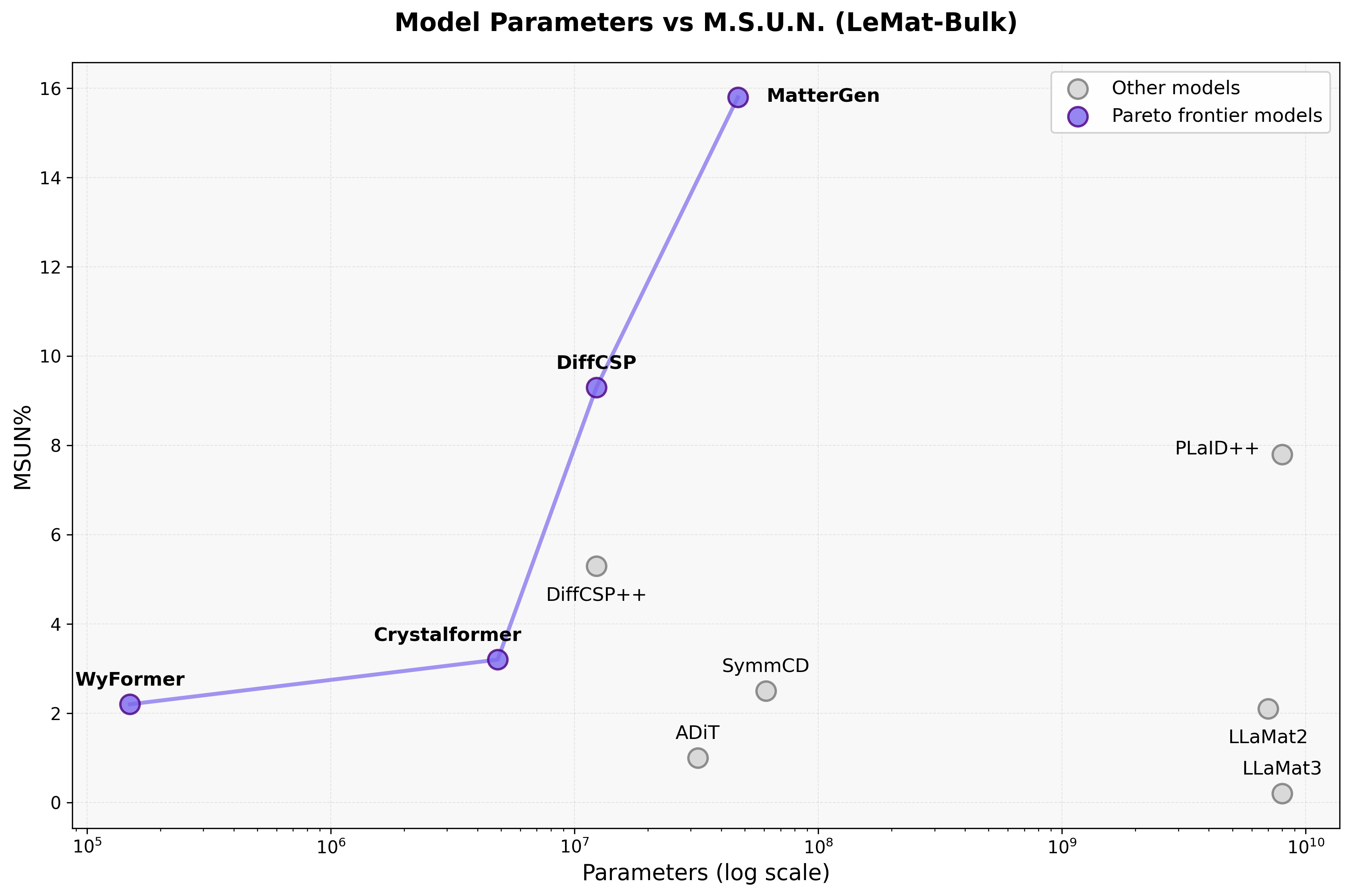}
    \caption{Model parameters vs MSUN\%}
    \label{fig:params_msun}
\end{subfigure}
\caption{Performance-efficiency trade-offs for crystal structure generation models. (a) Inference time (log scale) versus MSUN\% shows the computational cost required to achieve MSUN structures. Models on the Pareto frontier (connected by the blue line) represent optimal time-performance trade-offs. (b) Model parameter count (log scale) versus MSUN\% illustrates the relationship between model size and generation quality. WyFormer, Crystalformer, DiffCSP, and MatterGen form the Pareto frontier, demonstrating that larger models can achieve better performance when properly designed.}
\label{fig:efficiency_tradeoffs}
\end{figure}

We report the model efficiency metrics in~\Cref{fig:efficiency_tradeoffs}. Our Pareto frontier demonstrates that there does exist a correlation between model parameters and general M.S.U.N. performance; however, careful architectural design is still critical. On the inference-time axis, WyFormer occupies the extreme low-latency regime while MatterGen sits at the opposite end, delivering the highest M.S.U.N.\% at the expense of the longest runtimes. While the reported runtimes are not perfectly comparable---since models were evaluated on heterogeneous hardware---they nonetheless highlight the importance of prioritizing architectures that deliver strong performance under realistic computational budgets, which we view as critical for scalable downstream discovery.

A similar picture emerges when comparing M.S.U.N.\% against the number of model parameters. WyFormer and Crystalformer achieve competitive M.S.U.N.\% with comparatively small parameter counts, and DiffCSP and MatterGen extend this frontier to larger models. In contrast, SymmCD, ADiT, and especially LLaMat2/3 use one to two orders of magnitude more parameters than the frontier models yet deliver comparable or worse M.S.U.N.\%. Together, these results suggest that architectures explicitly tailored to crystal generation (WyFormer, Crystalformer, DiffCSP, and MatterGen) provide better M.S.U.N.\%-per-parameter and M.S.U.N.\%-per-second than more generic, heavily overparameterized generators.

\begin{table}[htbp]
\centering
\caption{Model evaluation metrics calculated using \textbf{MLIP ensemble}  with \textbf{LeMat-Bulk} as the reference set: Validity, Energy, Stability, and Metastability. All models provide 2500 structures except Crystalformer (1000 structures). Submissions are organized into five categories separated by horizontal lines: (1) models that incorporate relaxation as part of their generation pipeline (MatterGen, PLaID++, WyFormer, WyFormer-DFT), (2) other generative models, (3) samples from real-world datasets that contribute to LeMat-Bulk (Alexandria, OQMD) as baselines, (4) samples from the AFLOW dataset and MP-Exp structures, (5) pre-relaxed samples from non-ML generative baselines. Best values are shown in \textbf{bold}, second-best values are \underline{underlined}. Arrows indicate whether higher ({\textuparrow}) or lower ({\textdownarrow}) values are better.}
\scriptsize
\begin{adjustbox}{max width=\textwidth}
\begin{tabular}{l cccc c c ccc ccc ccc}
\toprule
\textbf{Model} &
\multicolumn{4}{c}{\textbf{Validity}} &
\textbf{Unique\% {\textuparrow}} & \textbf{Novel\% {\textuparrow}} &
\multicolumn{3}{c}{\textbf{Energy-based}} &
\multicolumn{3}{c}{\textbf{Stability}} &
\multicolumn{3}{c}{\textbf{Metastability}} \\
\cmidrule(lr){2-5} \cmidrule(lr){8-10} \cmidrule(lr){11-13} \cmidrule(lr){14-16}
& \textbf{Valid\% {\textuparrow}} & \textbf{CN\% {\textuparrow}} & \textbf{MinDist\% {\textuparrow}} & \textbf{PhysPlau\% {\textuparrow}}
&  &  & \textbf{FormE (Std) {\textdownarrow}} & \textbf{$E_{\mathrm{hull}}$ (Std) {\textdownarrow}} & \textbf{RMSD (Std) {\textdownarrow}}
& \textbf{Stable\% {\textuparrow}} & \textbf{U-Stable\% {\textuparrow}} & \textbf{SUN\% {\textuparrow}}
& \textbf{Metastable\% {\textuparrow}} & \textbf{U-Meta\% {\textuparrow}} & \textbf{MSUN\% {\textuparrow}} \\
\midrule
MatterGen & \underline{95.7} & 95.9 & \underline{99.8} & \textbf{100.0} & \textbf{95.1} & \textbf{70.5} & \meanstd{-0.71}{0.79} & \underline{\meanstd{0.17}{0.17}} & \meanstd{0.39}{0.50} & 0.5 {\scriptsize (12)} & 0.5 {\scriptsize (12)} & 0.1 {\scriptsize (2)} & 36.0 {\scriptsize (900)} & 35.5 {\scriptsize (887)} & \textbf{15.8} {\scriptsize (395)} \\
PLaID++ & \textbf{96.0} & 96.0 & 96.2 & 96.2 & 77.8 & 24.2 & \meanstd{-0.50}{0.44} & $\bm{\meanstd{0.08}{0.16}}$ & $\bm{\meanstd{0.13}{0.29}}$ & \textbf{6.4} {\scriptsize (159)} & \textbf{4.4} {\scriptsize (109)} & \textbf{0.9} {\scriptsize (22)} & \textbf{67.6} {\scriptsize (1689)} & \textbf{52.3} {\scriptsize (1307)} & 7.8 {\scriptsize (194)} \\
WyFormer & 93.4 & 95.2 & 98.2 & \textbf{100.0} & 93.0 & \underline{66.4} & \meanstd{-0.45}{0.95} & \meanstd{0.48}{0.50} & \meanstd{0.81}{0.98} & 0.2 {\scriptsize (6)} & 0.2 {\scriptsize (6)} & 0.0 {\scriptsize (1)} & 16.0 {\scriptsize (400)} & 15.8 {\scriptsize (394)} & 2.2 {\scriptsize (54)} \\
WyFormer-DFT & 95.2 & 95.2 & \textbf{100.0} & \textbf{100.0} & \underline{95.0} & \underline{66.4} & \meanstd{-0.67}{0.90} & \meanstd{0.26}{0.35} & \meanstd{0.42}{0.60} & \underline{1.5} {\scriptsize (38)} & \underline{1.5} {\scriptsize (37)} & \underline{0.3} {\scriptsize (7)} & 27.6 {\scriptsize (689)} & 27.4 {\scriptsize (684)} & 8.3 {\scriptsize (208)} \\
\midrule
ADiT & 90.6 & \textbf{99.0} & 91.5 & \textbf{100.0} & 87.8 & 26.0 & \underline{\meanstd{-0.74}{0.93}} & \meanstd{0.33}{0.44} & \underline{\meanstd{0.38}{0.40}} & 0.1 {\scriptsize (2)} & 0.1 {\scriptsize (2)} & 0.0 {\scriptsize (0)} & \underline{37.1} {\scriptsize (928)} & \underline{35.6} {\scriptsize (889)} & 1.0 {\scriptsize (26)} \\
Crystal-GFN& 51.7 & 58.3 & 91.1 & 96.3 & 51.7 & 51.7 & $\bm{\meanstd{-1.34}{2.24}}$ & \meanstd{2.05}{1.94} & \meanstd{1.87}{0.97} & 0.0 {\scriptsize (0)} & 0.0 {\scriptsize (0)} & 0.0 {\scriptsize (0)} & 0.0 {\scriptsize (0)} & 0.0 {\scriptsize (0)} & 0.0 {\scriptsize (0)} \\
Crystalformer & 69.9 & 96.3 & 72.9 & \underline{99.9} & 69.4 & 31.8 & \meanstd{-0.18}{1.47} & \meanstd{0.69}{1.31} & \meanstd{0.66}{1.05} & 0.9 {\scriptsize (9)} & 0.9 {\scriptsize (9)} & 0.0 {\scriptsize (0)} & 29.3 {\scriptsize (293)} & 29.0 {\scriptsize (290)} & 3.2 {\scriptsize (32)} \\
DiffCSP & \underline{95.7} & 96.2 & 99.5 & \textbf{100.0} & 94.8 & 66.2 & \meanstd{-0.65}{0.97} & \meanstd{0.26}{0.56} & \meanstd{0.59}{0.63} & 1.1 {\scriptsize (27)} & 1.0 {\scriptsize (25)} & 0.0 {\scriptsize (1)} & 32.2 {\scriptsize (804)} & 31.4 {\scriptsize (785)} & \underline{9.3} {\scriptsize (232)} \\
DiffCSP++ & 95.3 & 95.5 & 99.7 & \textbf{100.0} & \textbf{95.1} & 62.0 & \meanstd{-0.53}{0.87} & \meanstd{0.40}{0.43} & \meanstd{0.69}{0.82} & 0.7 {\scriptsize (17)} & 0.7 {\scriptsize (17)} & 0.2 {\scriptsize (4)} & 27.6 {\scriptsize (689)} & 27.4 {\scriptsize (684)} & 5.3 {\scriptsize (133)} \\
LLaMat2 & 84.4 & \underline{97.0} & 87.1 & 99.6 & 81.4 & 30.0 & \meanstd{-0.48}{1.00} & \meanstd{0.43}{0.70} & \meanstd{0.54}{0.71} & 0.4 {\scriptsize (10)} & 0.4 {\scriptsize (10)} & 0.1 {\scriptsize (3)} & 35.7 {\scriptsize (892)} & 33.4 {\scriptsize (836)} & 2.1 {\scriptsize (53)} \\
LLaMat3 & 15.4 & 82.5 & 16.8 & 84.2 & 15.2 & 10.5 & \meanstd{0.69}{2.61} & \meanstd{1.65}{2.39} & \meanstd{1.01}{1.10} & 0.0 {\scriptsize (0)} & 0.0 {\scriptsize (0)} & 0.0 {\scriptsize (0)} & 2.3 {\scriptsize (58)} & 2.2 {\scriptsize (56)} & 0.2 {\scriptsize (5)} \\
SymmCD & 73.4 & 95.9 & 76.4 & \textbf{100.0} & 73.0 & 47.0 & \meanstd{-0.04}{1.21} & \meanstd{0.86}{1.05} & \meanstd{0.87}{1.09} & 0.5 {\scriptsize (12)} & 0.5 {\scriptsize (12)} & 0.0 {\scriptsize (1)} & 19.4 {\scriptsize (485)} & 19.1 {\scriptsize (478)} & 2.5 {\scriptsize (62)} \\
\midrule
Alexandria & 93.4 & 93.4 & 99.0 & 99.0 & 93.4 & 0.0 & \meanstd{-0.19}{0.94} & \meanstd{0.43}{0.61} & \meanstd{0.14}{0.32} & 0.6 {\scriptsize (14)} & 0.6 {\scriptsize (14)} & 0.0 {\scriptsize (0)} & 29.2 {\scriptsize (730)} & 29.2 {\scriptsize (730)} & 0.0 {\scriptsize (0)} \\
OQMD & 96.8 & 96.8 & 99.4 & 99.5 & 96.4 & 0.0 & \meanstd{-0.24}{0.88} & \meanstd{0.39}{0.45} & \meanstd{0.14}{0.28} & 1.1 {\scriptsize (27)} & 1.1 {\scriptsize (27)} & 0.0 {\scriptsize (0)} & 33.4 {\scriptsize (835)} & 33.3 {\scriptsize (832)} & 0.0 {\scriptsize (0)} \\
\midrule
AFLOW & 91.4 & 91.4 & 100.0 & 100.0 & 87.1 & 21.5 & \meanstd{-0.19}{0.86} & \meanstd{0.34}{0.42} & \meanstd{0.12}{0.44} & 2.9 {\scriptsize (72)} & 2.8 {\scriptsize (69)} & 0.0 {\scriptsize (0)} & 36.9 {\scriptsize (922)} & 33.4 {\scriptsize (834)} & 0.6 {\scriptsize (15)} \\
MP-Exp & 96.9 & 97.1 & 98.8 & 98.9 & 96.8 & 3.7 & \meanstd{-1.10}{1.11} & \meanstd{0.18}{0.53} & \meanstd{0.20}{0.35} & 2.2 {\scriptsize (54)} & 2.2 {\scriptsize (54)} & 0.2 {\scriptsize (4)} & 63.2 {\scriptsize (1581)} & 63.2 {\scriptsize (1580)} & 2.3 {\scriptsize (57)} \\
\midrule
RandomEnum & 97.0 & 99.0 & 97.8 & 99.8 & 97.0 & 94.5 &  \meanstd{-0.65}{0.84} & \meanstd{0.22}{0.13} & \meanstd{0.81}{0.66} & 0.1 {\scriptsize (3)} & 0.1 {\scriptsize (3)} & 0.1 {\scriptsize (3)} & 15.8 {\scriptsize (396)} & 15.8 {\scriptsize (396)} & 14.1 {\scriptsize (352)} \\
IonSub & 99.2 & 99.3 & 100.0 & 100.0 & 91.6 & 1.4 &  \meanstd{-0.43}{0.40} & \meanstd{0.05}{0.08} & \meanstd{0.07}{0.16} & 7.5 {\scriptsize (188)} & 7.1 {\scriptsize (177)} & 0.0 {\scriptsize (1)} & 78.9 {\scriptsize (1973)} & 73.0 {\scriptsize (1826)} & 1.2 {\scriptsize (30)} \\
\bottomrule
\end{tabular}
\end{adjustbox}
\end{table}

\begin{table}[htbp]
\centering
\caption{Model evaluation metrics calculated using \textbf{MLIP ensemble} with \textbf{LeMat-Bulk} as the reference set: Distribution, Diversity, and HHI. All models provide 2500 structures except Crystalformer (1000 structures). Submissions are organized into five categories separated by horizontal lines: (1) models that incorporate relaxation as part of their generation pipeline (MatterGen, PLaID++, WyFormer, WyFormer-DFT), (2) other generative models, (3) samples from real-world datasets that contribute to LeMat-Bulk (Alexandria, OQMD) as baselines, (4) samples from the AFLOW dataset and MP-Exp structures, (5) pre-relaxed samples from non-ML generative baselines. Best values are shown in \textbf{bold}, second-best values are \underline{underlined}. Arrows indicate whether higher ({\textuparrow}) or lower ({\textdownarrow}) values are better.}
\scriptsize
\begin{adjustbox}{max width=\textwidth}
\begin{tabular}{l ccc cccc cc}
\toprule
\textbf{Model} &
\multicolumn{3}{c}{\textbf{Distribution}} &
\multicolumn{4}{c}{\textbf{Diversity}} &
\multicolumn{2}{c}{\textbf{HHI}} \\
\cmidrule(lr){2-4} \cmidrule(lr){5-8} \cmidrule(lr){9-10}
& \textbf{JS {\textdownarrow}} & \textbf{MMD {\textdownarrow}} & \textbf{FID {\textdownarrow}}
& \textbf{ElemDiv {\textuparrow}} & \textbf{SGDiv {\textuparrow}} & \textbf{SizeDiv {\textuparrow}} & \textbf{SiteDiv {\textuparrow}}
& \textbf{Prod {\textdownarrow}} & \textbf{Res {\textdownarrow}} \\
\midrule
MatterGen & 0.44 & 0.006 & 2.13 & 0.64 & 0.14 & 0.24 & 12.78 & 3.52 & 2.72 \\
PLaID++ & 0.43 & 0.029 & 2.69 & 0.68 & 0.26 & 0.21 & 6.01 & 5.23 & 3.36 \\
WyFormer & \underline{0.22} & 0.006 & 1.55 & \underline{0.74} & \underline{0.50} & 0.27 & \underline{23.84} & 3.55 & 2.73 \\
WyFormer-DFT & 0.25 & 0.010 & 1.72 & 0.73 & 0.49 & 0.26 & 23.56 & 3.49 & 2.74 \\
\midrule
ADiT & 0.51 & \textbf{0.002} & 1.72 & 0.71 & 0.03 & 0.23 & 14.18 & 3.52 & 2.78 \\
Crystal-GFN & 0.55 & 0.121 & 43.17 & \textbf{0.75} & 0.29 & \textbf{0.32} & \textbf{32.06} & \underline{3.48} & \textbf{2.34} \\
Crystalformer & 0.26 & \underline{0.003} & 2.31 & 0.71 & 0.35 & \underline{0.31} & 18.59 & 3.78 & 2.82 \\
DiffCSP & 0.45 & 0.008 & 2.25 & 0.73 & 0.14 & 0.24 & 14.42 & \textbf{3.29} & \underline{2.60} \\
DiffCSP++ & \textbf{0.21} & \underline{0.003} & \underline{1.28} & 0.72 & \textbf{0.51} & 0.27 & 20.84 & 3.56 & 2.77 \\
LLaMat2 & 0.32 & \textbf{0.002} & \textbf{1.23} & 0.71 & 0.25 & 0.24 & 9.86 & 3.94 & 2.86 \\
LLaMat3 & 0.34 & 0.005 & 9.48 & 0.70 & 0.10 & 0.29 & 13.15 & 3.83 & 2.81 \\
SymmCD & \underline{0.22} & 0.006 & 1.65 & 0.73 & 0.49 & 0.27 & 19.45 & 3.54 & 2.74 \\
\midrule
Alexandria & 0.09 & 0.000 & 1.31 & 0.71 & 0.50 & 0.25 & 13.50 & 4.02 & 3.33 \\
OQMD & 0.27 & 0.004 & 0.83 & 0.63 & 0.48 & 0.24 & 12.57 & 4.22 & 3.29 \\
\midrule
AFLOW & 0.28 & 0.004 & 1.78 & 0.69 & 0.46 & 0.25 & 10.07 & 4.12 & 3.22 \\
MP-Exp & 0.33 & 0.021 & 2.90 & 0.72 & 0.71 & 0.28 & 54.79 & 2.78 & 2.27 \\
\midrule
RandomEnum & 0.46 & 0.034 & 2.93 & 0.62 & 0.23 & 0.27 & 15.48 & 3.12 & 3.09 \\
IonSub & 0.32 & 0.004 & 2.04 & 0.67 & 0.36 & 0.26 & 10.82 & 4.15 & 3.86 \\
\bottomrule
\end{tabular}
\end{adjustbox}
\end{table}

\begin{table}[htbp]
\centering
\caption{Model evaluation metrics calculated using \textbf{uma} with \textbf{LeMat-Bulk} as the reference set: Validity, Energy, Stability, and Metastability. All models provide 2500 structures except Crystalformer (1000 structures). Submissions are organized into five categories separated by horizontal lines: (1) models that incorporate relaxation as part of their generation pipeline (MatterGen, PLaID++, WyFormer, WyFormer-DFT), (2) other generative models, (3) samples from real-world datasets that contribute to LeMat-Bulk (Alexandria, OQMD) as baselines, (4) samples from the AFLOW dataset and MP-Exp structures, (5) pre-relaxed samples from non-ML generative baselines. Best values are shown in \textbf{bold}, second-best values are \underline{underlined}. Arrows indicate whether higher ({\textuparrow}) or lower ({\textdownarrow}) values are better.}
\scriptsize
\label{tab:uma_bulk_prerelax}
\begin{adjustbox}{max width=\textwidth}
\begin{tabular}{l cccc c c ccc ccc ccc}
\toprule
\textbf{Model} &
\multicolumn{4}{c}{\textbf{Validity}} &
\textbf{Unique\% {\textuparrow}} & \textbf{Novel\% {\textuparrow}} &
\multicolumn{3}{c}{\textbf{Energy-based}} &
\multicolumn{3}{c}{\textbf{Stability}} &
\multicolumn{3}{c}{\textbf{Metastability}} \\
\cmidrule(lr){2-5} \cmidrule(lr){8-10} \cmidrule(lr){11-13} \cmidrule(lr){14-16}
& \textbf{Valid\% {\textuparrow}} & \textbf{CN\% {\textuparrow}} & \textbf{MinDist\% {\textuparrow}} & \textbf{PhysPlau\% {\textuparrow}}
&  &  & \textbf{FormE (Std) {\textdownarrow}} & \textbf{$E_{\mathrm{hull}}$ (Std) {\textdownarrow}} & \textbf{RMSD (Std) {\textdownarrow}}
& \textbf{Stable\% {\textuparrow}} & \textbf{U-Stable\% {\textuparrow}} & \textbf{SUN\% {\textuparrow}}
& \textbf{Metastable\% {\textuparrow}} & \textbf{U-Meta\% {\textuparrow}} & \textbf{MSUN\% {\textuparrow}} \\
\midrule
MatterGen & \underline{95.7} & 95.9 & \underline{99.8} & \textbf{100.0} & \textbf{95.1} & \textbf{70.5} & \underline{\meanstd{-0.69}{0.80}} & \underline{\meanstd{0.19}{0.19}} & \meanstd{0.26}{0.30} & 1.3 {\scriptsize (32)} & 1.2 {\scriptsize (31)} & 0.1 {\scriptsize (3)} & 32.7 {\scriptsize (818)} & 32.2 {\scriptsize (806)} & \textbf{14.2 {\scriptsize (354)}} \\
PLaID++ & \textbf{96.0} & 96.0 & 96.2 & 96.2 & 77.8 & 24.2 & \meanstd{-0.48}{0.48} & $\bm{\meanstd{0.09}{0.17}}$ & $\bm{\meanstd{0.10}{0.20}}$ & \textbf{10.0} {\scriptsize (250)} & \textbf{6.9} {\scriptsize (173)} & \textbf{1.0} {\scriptsize (24)} & \textbf{62.1} {\scriptsize (1552)} & \textbf{47.8} {\scriptsize (1194)} & 7.4 {\scriptsize (185)} \\
WyFormer & 93.4 & 95.2 & 98.2 & \textbf{100.0} & 93.0 & \underline{66.4} & \meanstd{-0.41}{0.97} & \meanstd{0.51}{0.52} & \meanstd{0.53}{0.54} & 0.5 {\scriptsize (12)} & 0.5 {\scriptsize (12)} & 0.0 {\scriptsize (1)} & 15.2 {\scriptsize (381)} & 15.0 {\scriptsize (375)} & 2.1 {\scriptsize (53)} \\
WyFormer-DFT & 95.2 & 95.2 & \textbf{100.0} & \textbf{100.0} & \underline{95.0} & \underline{66.4} & \meanstd{-0.65}{1.07} & \meanstd{0.28}{0.65} & \underline{\meanstd{0.19}{0.28}} & \underline{2.2 {\scriptsize (56)}} & \underline{2.2 {\scriptsize (55)}} & \underline{0.2 {\scriptsize (5)}} & 25.9 {\scriptsize (648)} & 25.7 {\scriptsize (643)} & 7.8 {\scriptsize (195)} \\
\midrule
ADiT & 90.6 & \textbf{99.0} & 91.5 & \textbf{100.0} & 87.8 & 26.0 & $\bm{\meanstd{-0.72}{0.95}}$ & \meanstd{0.34}{0.45} & \meanstd{0.31}{0.29} & 0.1 {\scriptsize (3)} & 0.1 {\scriptsize (3)} & 0.0 {\scriptsize (0)} & \underline{35.4 {\scriptsize (886)}} & \underline{33.9 {\scriptsize (848)}} & 0.9 {\scriptsize (23)} \\
Crystal-GFN & 51.7 & 58.3 & 91.1 & 96.3 & 51.7 & 51.7 & \meanstd{-0.56}{4.46} & \meanstd{2.83}{4.23} & \meanstd{0.83}{0.34} & 0.0 {\scriptsize (0)} & 0.0 {\scriptsize (0)} & 0.0 {\scriptsize (0)} & 0.0 {\scriptsize (0)} & 0.0 {\scriptsize (0)} & 0.0 {\scriptsize (0)} \\
Crystalformer & 69.9 & 96.3 & 72.9 & \underline{99.9} & 69.4 & 31.8 & \meanstd{-0.16}{1.48} & \meanstd{0.71}{1.31} & \meanstd{0.46}{0.61} & 1.4 {\scriptsize (14)} & 1.4 {\scriptsize (14)} & 0.0 {\scriptsize (0)} & 28.4 {\scriptsize (284)} & 28.1 {\scriptsize (281)} & 3.2 {\scriptsize (32)} \\
DiffCSP & \underline{95.7} & 96.2 & 99.5 & \textbf{100.0} & 94.8 & 66.2 & \meanstd{-0.62}{0.98} & \meanstd{0.28}{0.58} & \meanstd{0.37}{0.38} & 2.0 {\scriptsize (49)} & 1.8 {\scriptsize (46)} & \underline{0.2} {\scriptsize (4)} & 29.7 {\scriptsize (743)} & 29.1 {\scriptsize (727)} & \underline{8.2} {\scriptsize (205)} \\
DiffCSP++ & 95.3 & 95.5 & 99.7 & \textbf{100.0} & \textbf{95.1} & 62.0 & \meanstd{-0.50}{0.88} & \meanstd{0.42}{0.44} & \meanstd{0.47}{0.50} & 1.2 {\scriptsize (30)} & 1.2 {\scriptsize (30)} & \underline{0.2} {\scriptsize (4)} & 25.4 {\scriptsize (634)} & 25.2 {\scriptsize (629)} & 4.8 {\scriptsize (120)} \\
LLaMat2 & 84.4 & \underline{97.0} & 87.1 & 99.6 & 81.4 & 30.0 & \meanstd{-0.45}{1.03} & \meanstd{0.45}{0.72} & \meanstd{0.43}{0.49} & 1.0 {\scriptsize (26)} & 1.0 {\scriptsize (25)} & \underline{0.2} {\scriptsize (4)} & 33.4 {\scriptsize (834)} & 31.2 {\scriptsize (781)} & 1.9 {\scriptsize (48)} \\
LLaMat3 & 15.4 & 82.5 & 16.8 & 84.2 & 15.2 & 10.5 & \meanstd{0.79}{2.73} & \meanstd{1.74}{2.52} & \meanstd{0.75}{0.84} & 0.0 {\scriptsize (0)} & 0.0 {\scriptsize (0)} & 0.0 {\scriptsize (0)} & 2.2 {\scriptsize (54)} & 2.1 {\scriptsize (52)} & 0.2 {\scriptsize (5)} \\
SymmCD & 73.4 & 95.9 & 76.4 & \textbf{100.0} & 73.0 & 47.0 & \meanstd{0.00}{1.23} & \meanstd{0.89}{1.08} & \meanstd{0.58}{0.61} & 1.2 {\scriptsize (29)} & 1.2 {\scriptsize (29)} & 0.1 {\scriptsize (2)} & 18.6 {\scriptsize (466)} & 18.4 {\scriptsize (460)} & 2.4 {\scriptsize (60)} \\
\midrule
Alexandria & 93.4 & 93.4 & 99.0 & 99.0 & 93.4 & 0.0 & \meanstd{-0.16}{0.97} & \meanstd{0.45}{0.64} & \meanstd{0.11}{0.39} & 1.2 {\scriptsize (29)} & 1.2 {\scriptsize (29)} & 0.0 {\scriptsize (0)} & 27.8 {\scriptsize (695)} & 27.8 {\scriptsize (695)} & 0.0 {\scriptsize (0)} \\
OQMD & 96.8 & 96.8 & 99.4 & 99.5 & 96.4 & 0.0 & \meanstd{-0.21}{0.89} & \meanstd{0.40}{0.46} & \meanstd{0.13}{0.30} & 3.6 {\scriptsize (90)} & 3.6 {\scriptsize (90)} & 0.0 {\scriptsize (0)} & 30.5 {\scriptsize (762)} & 30.4 {\scriptsize (759)} & 0.0 {\scriptsize (0)} \\
\midrule
AFLOW & 91.4 & 91.4 & 100.0 & 100.0 & 87.1 & 21.5 & \meanstd{-0.18}{0.86} & \meanstd{0.35}{0.42} & \meanstd{0.11}{0.40} & 5.8 {\scriptsize (146)} & 5.4 {\scriptsize (136)} & 0.0 {\scriptsize (0)} & 33.5 {\scriptsize (838)} & 30.4 {\scriptsize (760)} & 0.7 {\scriptsize (17)} \\
MP-Exp & 96.9 & 97.1 & 98.8 & 98.9 & 96.8 & 3.7 & \meanstd{-1.07}{1.32} & \meanstd{0.20}{0.87} & \meanstd{0.22}{0.42} & 4.0 {\scriptsize (99)} & 4.0 {\scriptsize (99)} & 0.2 {\scriptsize (4)} & 59.6 {\scriptsize (1490)} & 59.6 {\scriptsize (1489)} & 2.2 {\scriptsize (54)} \\
\midrule
RandomEnum & 97.0 & 99.0 & 97.8 & 99.8 & 97.0 & 94.5 &  \meanstd{-0.62}{0.86} & \meanstd{0.25}{0.15} & \meanstd{0.80}{0.77} & 0.1 {\scriptsize (3)} & 0.1 {\scriptsize (3)} & 0.1 {\scriptsize (2)} & 14.8 {\scriptsize (369)} & 14.8 {\scriptsize (369)} & 13.1 {\scriptsize (327)} \\
IonSub & 99.2 & 99.3 & 100.0 & 100.0 & 91.6 & 1.4 &  \meanstd{-0.42}{0.43} & \meanstd{0.05}{0.09} & \meanstd{0.06}{0.17} & 16.5 {\scriptsize (412)} & 15.7 {\scriptsize (392)} & 0.1 {\scriptsize (2)} & 68.7 {\scriptsize (1718)} & 63.3 {\scriptsize (1582)} & 1.0 {\scriptsize (26)} \\
\bottomrule
\end{tabular}
\end{adjustbox}
\end{table}

\begin{table}[htbp]
\centering
\caption{Model evaluation metrics calculated using \textbf{uma} with \textbf{LeMat-Bulk} as the reference set: Distribution, Diversity, and HHI. All models provide 2500 structures except Crystalformer (1000 structures). Submissions are organized into five categories separated by horizontal lines: (1) models that incorporate relaxation as part of their generation pipeline (MatterGen, PLaID++, WyFormer, WyFormer-DFT), (2) other generative models, (3) samples from real-world datasets that contribute to LeMat-Bulk (Alexandria, OQMD) as baselines, (4) samples from the AFLOW dataset and MP-Exp structures, (5) pre-relaxed samples from non-ML generative baselines. Best values are shown in \textbf{bold}, second-best values are \underline{underlined}. Arrows indicate whether higher ({\textuparrow}) or lower ({\textdownarrow}) values are better.}
\scriptsize
\begin{adjustbox}{max width=\textwidth}
\begin{tabular}{l ccc cccc cc}
\toprule
\textbf{Model} &
\multicolumn{3}{c}{\textbf{Distribution}} &
\multicolumn{4}{c}{\textbf{Diversity}} &
\multicolumn{2}{c}{\textbf{HHI}} \\
\cmidrule(lr){2-4} \cmidrule(lr){5-8} \cmidrule(lr){9-10}
& \textbf{JS {\textdownarrow}} & \textbf{MMD {\textdownarrow}} & \textbf{FID {\textdownarrow}}
& \textbf{ElemDiv {\textuparrow}} & \textbf{SGDiv {\textuparrow}} & \textbf{SizeDiv {\textuparrow}} & \textbf{SiteDiv {\textuparrow}}
& \textbf{Prod {\textdownarrow}} & \textbf{Res {\textdownarrow}} \\
\midrule
MatterGen & 0.44 & 0.006 & 0.38 & 0.64 & 0.14 & 0.24 & 12.78 & 3.52 & 2.72 \\
PLaID++ & 0.43 & 0.029 & 0.96 & 0.68 & 0.26 & 0.21 & 6.01 & 5.23 & 3.36 \\
WyFormer & \underline{0.22} & 0.006 & \underline{0.10} & \underline{0.74} & \underline{0.50} & 0.27 & \underline{23.84} & 3.55 & 2.73 \\
WyFormer-DFT & 0.25 & 0.010 & 0.26 & 0.73 & 0.49 & 0.26 & 23.56 & 3.49 & 2.74 \\
\midrule
ADiT & 0.51 & \textbf{0.002} & 0.27 & 0.71 & 0.03 & 0.24 & 14.18 & 3.52 & 2.78 \\
Crystal-GFN & 0.55 & 0.121 & 1.70 & \textbf{0.75} & 0.29 & \textbf{0.32} & \textbf{32.06} & \underline{3.48} & \textbf{2.34} \\
Crystalformer & 0.26 & \underline{0.003} & 0.74 & 0.71 & 0.35 & \underline{0.31} & 18.59 & 3.78 & 2.82 \\
DiffCSP & 0.45 & 0.008 & 0.11 & 0.73 & 0.13 & 0.24 & 14.42 & \textbf{3.29} & \underline{2.60} \\
DiffCSP++ & \textbf{0.21} & \underline{0.003} & \textbf{0.09} & 0.72 & \textbf{0.51} & 0.27 & 20.84 & 3.56 & 2.77 \\
LLaMat2 & 0.32 & \textbf{0.002} & 0.16 & 0.71 & 0.25 & 0.24 & 9.86 & 3.94 & 2.86 \\
LLaMat3 & 0.34 & 0.005 & 5.14 & 0.70 & 0.10 & 0.30 & 13.15 & 3.83 & 2.81 \\
SymmCD & \underline{0.22} & 0.006 & 0.96 & 0.73 & 0.49 & 0.27 & 19.45 & 3.54 & 2.74 \\
\midrule
Alexandria & 0.09 & 0.000 & 0.01 & 0.71 & 0.50 & 0.25 & 13.50 & 4.02 & 3.33 \\
OQMD & 0.27 & 0.004 & 0.14 & 0.63 & 0.48 & 0.24 & 12.57 & 4.22 & 3.29 \\
\midrule
AFLOW & 0.28 & 0.004 & 0.12 & 0.69 & 0.46 & 0.25 & 10.07 & 4.12 & 3.22 \\
MP-Exp & 0.33 & 0.021 & 0.69 & 0.72 & 0.71 & 0.28 & 54.79 & 2.78 & 2.27 \\
\midrule
RandomEnum & 0.46 & 0.034 & 0.18 & 0.62 & 0.23 & 0.27 & 15.48 & 3.12 & 3.09 \\
IonSub & 0.32 & 0.004 & 0.74 & 0.67 & 0.36 & 0.26 & 10.82 & 4.15 & 3.86 \\
\bottomrule
\end{tabular}
\end{adjustbox}
\end{table}

\begin{table}[htbp]
\centering
\caption{Model evaluation metrics calculated using \textbf{uma} with \textbf{LeMat-Bulk} as the reference set \textbf{after relaxation}: Validity, Energy, Stability, and Metastability. All models provide 2500 structures except Crystalformer (1000 structures). Submissions are organized into five categories separated by horizontal lines: (1) models that incorporate relaxation as part of their generation pipeline (MatterGen, PLaID++, WyFormer, WyFormer-DFT), (2) other generative models, (3) samples from real-world datasets that contribute to LeMat-Bulk (Alexandria, OQMD) as baselines, (4) samples from the AFLOW dataset and MP-Exp structures, (5) pre-relaxed samples from non-ML generative baselines. Best values are shown in \textbf{bold}, second-best values are \underline{underlined}. Arrows indicate whether higher ({\textuparrow}) or lower ({\textdownarrow}) values are better.}
\scriptsize
\label{tab:uma_bulk_postrelax}
\begin{adjustbox}{max width=\textwidth}
\begin{tabular}{l cccc c c ccc ccc ccc}
\toprule
\textbf{Model} &
\multicolumn{4}{c}{\textbf{Validity}} &
\textbf{Unique\% {\textuparrow}} & \textbf{Novel\% {\textuparrow}} &
\multicolumn{3}{c}{\textbf{Energy-based}} &
\multicolumn{3}{c}{\textbf{Stability}} &
\multicolumn{3}{c}{\textbf{Metastability}} \\
\cmidrule(lr){2-5} \cmidrule(lr){8-10} \cmidrule(lr){11-13} \cmidrule(lr){14-16}
& \textbf{Valid\% {\textuparrow}} & \textbf{CN\% {\textuparrow}} & \textbf{MinDist\% {\textuparrow}} & \textbf{PhysPlau\% {\textuparrow}}
&  &  & \textbf{FormE (Std) {\textdownarrow}} & \textbf{$E_{\mathrm{hull}}$ (Std) {\textdownarrow}} & \textbf{RMSD (Std) {\textdownarrow}}
& \textbf{Stable\% {\textuparrow}} & \textbf{U-Stable\% {\textuparrow}} & \textbf{SUN\% {\textuparrow}}
& \textbf{Metastable\% {\textuparrow}} & \textbf{U-Meta\% {\textuparrow}} & \textbf{MSUN\% {\textuparrow}} \\
\midrule
MatterGen & 95.9 & 95.9 & \textbf{100.0} & \textbf{100.0} & \textbf{95.3} & \textbf{70.4} & \meanstd{-0.74}{0.80} & \meanstd{0.14}{0.14} & $\bm{\meanstd{0.00}{0.02}}$ & 4.5 {\scriptsize (113)} & 4.4 {\scriptsize (111)} & 0.8 {\scriptsize (20)} & 39.5 {\scriptsize (987)} & 39.0 {\scriptsize (976)} & \textbf{21.7} {\scriptsize (543)} \\
PLaID++ & 96.0 & 96.0 & 96.2 & 96.2 & 77.8 & 23.8 & \meanstd{-0.49}{0.47} & \underline{\meanstd{0.08}{0.15}} & $\bm{\meanstd{0.00}{0.01}}$ & \underline{16.2} {\scriptsize (404)} & \underline{10.9} {\scriptsize (272)} & \textbf{1.3} {\scriptsize (32)} & \underline{57.2} {\scriptsize (1430)} & 44.9 {\scriptsize (1123)} & 7.9 {\scriptsize (198)} \\
WyFormer & 95.0 & 95.0 & 99.8 & 99.8 & 94.5 & 66.0 & \meanstd{-0.68}{0.89} & \meanstd{0.25}{0.28} & \underline{\meanstd{0.01}{0.05}} & 4.0 {\scriptsize (101)} & 3.9 {\scriptsize (97)} & 0.5 {\scriptsize (12)} & 27.5 {\scriptsize (687)} & 27.2 {\scriptsize (679)} & 10.4 {\scriptsize (260)} \\
WyFormer-DFT & 95.2 & 95.2 & \textbf{100.0} & \textbf{100.0} & 94.9 & \underline{66.3} & \meanstd{-0.71}{0.89} & \meanstd{0.22}{0.25} & $\bm{\meanstd{0.00}{0.02}}$ & 3.8 {\scriptsize (94)} & 3.6 {\scriptsize (90)} & 0.4 {\scriptsize (10)} & 27.6 {\scriptsize (691)} & 27.6 {\scriptsize (689)} & 10.3 {\scriptsize (257)} \\
\midrule
ADiT & \textbf{98.8} & \textbf{98.9} & 99.8 & \underline{99.9} & 95.1 & 24.3 & \underline{\meanstd{-1.04}{0.90}} & $\bm{\meanstd{0.06}{0.09}}$ & $\bm{\meanstd{0.00}{0.01}}$ & \textbf{17.6} {\scriptsize (441)} & \textbf{16.5} {\scriptsize (413)} & 0.6 {\scriptsize (16)} & \textbf{62.5} {\scriptsize (1562)} & \textbf{60.2} {\scriptsize (1506)} & 10.0 {\scriptsize (250)} \\
Crystal-GFN & 55.3 & 57.4 & 98.0 & 95.7 & 55.3 & 55.3 & $\bm{\meanstd{-2.54}{2.08}}$ & \meanstd{0.86}{1.74} & \meanstd{0.12}{0.14} & 0.0 {\scriptsize (0)} & 0.0 {\scriptsize (0)} & 0.0 {\scriptsize (0)} & 1.0 {\scriptsize (24)} & 1.0 {\scriptsize (24)} & 1.0 {\scriptsize (24)} \\
Crystalformer & 91.0 & 92.5 & 94.6 & 96.1 & 90.5 & 47.5 & \meanstd{-0.73}{0.86} & \meanstd{0.17}{0.24} & \underline{\meanstd{0.01}{0.08}} & 9.6 {\scriptsize (96)} & 9.5 {\scriptsize (95)} & 0.8 {\scriptsize (8)} & 39.6 {\scriptsize (396)} & 39.3 {\scriptsize (393)} & 9.8 {\scriptsize (98)} \\
DiffCSP & 96.2 & 96.2 & \textbf{100.0} & \textbf{100.0} & 94.8 & 65.7 & \meanstd{-0.76}{0.82} & \meanstd{0.14}{0.15} & \underline{\meanstd{0.01}{0.05}} & 5.9 {\scriptsize (148)} & 5.3 {\scriptsize (133)} & 0.8 {\scriptsize (20)} & 40.0 {\scriptsize (1000)} & 39.5 {\scriptsize (988)} & \underline{19.4} {\scriptsize (484)} \\
DiffCSP++ & 95.5 & 95.5 & \textbf{100.0} & \textbf{100.0} & \underline{95.2} & 59.8 & \meanstd{-0.72}{0.86} & \meanstd{0.20}{0.21} & \underline{\meanstd{0.01}{0.04}} & 5.3 {\scriptsize (132)} & 5.2 {\scriptsize (131)} & 0.5 {\scriptsize (13)} & 31.6 {\scriptsize (789)} & 31.4 {\scriptsize (785)} & 10.1 {\scriptsize (253)} \\
LLaMat2 & \underline{97.0} & \underline{97.0} & \underline{99.9} & \textbf{100.0} & 93.8 & 38.2 & \meanstd{-0.77}{0.85} & \meanstd{0.13}{0.19} & $\bm{\meanstd{0.00}{0.04}}$ & 11.3 {\scriptsize (282)} & 10.6 {\scriptsize (266)} & \underline{1.0} {\scriptsize (24)} & 48.1 {\scriptsize (1202)} & \underline{45.8} {\scriptsize (1145)} & 9.1 {\scriptsize (227)} \\
LLaMat3 & 69.8 & 70.5 & 82.2 & 83.3 & 69.6 & 64.5 & \meanstd{-0.38}{0.71} & \meanstd{0.27}{0.26} & \meanstd{0.02}{0.06} & 0.7 {\scriptsize (17)} & 0.7 {\scriptsize (17)} & 0.2 {\scriptsize (4)} & 14.6 {\scriptsize (366)} & 14.6 {\scriptsize (364)} & 12.1 {\scriptsize (302)} \\
SymmCD & 95.6 & 95.9 & 99.7 & \textbf{100.0} & \underline{95.2} & 60.7 & \meanstd{-0.68}{0.84} & \meanstd{0.23}{0.25} & \underline{\meanstd{0.01}{0.05}} & 5.6 {\scriptsize (140)} & 5.5 {\scriptsize (138)} & 0.6 {\scriptsize (15)} & 30.6 {\scriptsize (764)} & 30.3 {\scriptsize (758)} & 9.4 {\scriptsize (236)} \\
\midrule
Alexandria & 93.4 & 93.4 & 99.0 & 99.0 & 93.4 & 4.6 & \meanstd{-0.18}{0.95} & \meanstd{0.43}{0.60} & \meanstd{0.00}{0.02} & 1.7 {\scriptsize (43)} & 1.7 {\scriptsize (43)} & 0.0 {\scriptsize (1)} & 28.1 {\scriptsize (702)} & 28.1 {\scriptsize (702)} & 0.8 {\scriptsize (20)} \\
OQMD & 96.8 & 96.8 & 99.5 & 99.5 & 96.4 & 4.2 & \meanstd{-0.22}{0.90} & \meanstd{0.38}{0.45} & \meanstd{0.00}{0.00} & 6.3 {\scriptsize (158)} & 6.2 {\scriptsize (156)} & 0.2 {\scriptsize (4)} & 29.2 {\scriptsize (730)} & 29.2 {\scriptsize (729)} & 0.9 {\scriptsize (22)} \\
\midrule
AFLOW & 91.4 & 91.4 & 100.0 & 100.0 & 87.1 & 21.6 & \meanstd{-0.18}{0.87} & \meanstd{0.34}{0.42} & \meanstd{0.00}{0.00} & 12.3 {\scriptsize (307)} & 11.3 {\scriptsize (282)} & 0.0 {\scriptsize (0)} & 28.3 {\scriptsize (708)} & 25.6 {\scriptsize (639)} & 0.8 {\scriptsize (21)} \\
MP-Exp & 97.0 & 97.1 & 98.9 & 98.9 & 97.0 & 13.6 & \meanstd{-1.12}{1.09} & \meanstd{0.16}{0.49} & \meanstd{0.00}{0.01} & 10.7 {\scriptsize (268)} & 10.7 {\scriptsize (268)} & 0.4 {\scriptsize (10)} & 56.2 {\scriptsize (1404)} & 56.1 {\scriptsize (1403)} & 6.2 {\scriptsize (154)} \\
\midrule
RandomEnum & 98.5 & 98.8 & 99.4 & 99.6 & 98.5 & 95.5 &  \meanstd{-0.69}{0.83} & \meanstd{0.17}{0.11} & \meanstd{0.00}{0.04} & 0.8 {\scriptsize (21)} & 0.8 {\scriptsize (21)} & 0.5 {\scriptsize (13)} & 28.3 {\scriptsize (708)} & 28.3 {\scriptsize (708)} & 26.5 {\scriptsize (662)} \\
IonSub & 99.3 & 99.3 & 100.0 & 100.0 & 91.8 & 2.0 &  \meanstd{-0.42}{0.43} & \meanstd{0.04}{0.08} & \meanstd{0.00}{0.00} & 21.1 {\scriptsize (528)} & 19.7 {\scriptsize (492)} & 0.2 {\scriptsize (4)} & 65.2 {\scriptsize (1631)} & 60.0 {\scriptsize (1499)} & 1.3 {\scriptsize (33)} \\
\bottomrule
\label{tab:uma_post_relax}
\end{tabular}
\end{adjustbox}
\end{table}

\begin{table}[htbp]
\centering
\caption{Model evaluation metrics calculated using \textbf{uma} with \textbf{LeMat-Bulk} as the reference set \textbf{after relaxation}: Distribution, Diversity, and HHI. All models provide 2500 structures except Crystalformer (1000 structures). Submissions are organized into five categories separated by horizontal lines: (1) models that incorporate relaxation as part of their generation pipeline (MatterGen, PLaID++, WyFormer, WyFormer-DFT), (2) other generative models, (3) samples from real-world datasets that contribute to LeMat-Bulk (Alexandria, OQMD) as baselines, (4) samples from the AFLOW dataset and MP-Exp structures, (5) pre-relaxed samples from non-ML generative baselines. Best values are shown in \textbf{bold}, second-best values are \underline{underlined}. Arrows indicate whether higher ({\textuparrow}) or lower ({\textdownarrow}) values are better.}
\scriptsize
\label{tab:table_10}
\begin{adjustbox}{max width=\textwidth}
\begin{tabular}{l ccc cccc cc}
\toprule
\textbf{Model} &
\multicolumn{3}{c}{\textbf{Distribution}} &
\multicolumn{4}{c}{\textbf{Diversity}} &
\multicolumn{2}{c}{\textbf{HHI}} \\
\cmidrule(lr){2-4} \cmidrule(lr){5-8} \cmidrule(lr){9-10}
& \textbf{JS {\textdownarrow}} & \textbf{MMD {\textdownarrow}} & \textbf{FID {\textdownarrow}}
& \textbf{ElemDiv {\textuparrow}} & \textbf{SGDiv {\textuparrow}} & \textbf{SizeDiv {\textuparrow}} & \textbf{SiteDiv {\textuparrow}}
& \textbf{Prod {\textdownarrow}} & \textbf{Res {\textdownarrow}} \\
\midrule
MatterGen & 0.35 & 0.006 & 0.51 & 0.64 & 0.32 & 0.24 & 12.81 & 3.51 & 2.72 \\
PLaID++ & 0.43 & 0.029 & 1.02 & 0.68 & 0.30 & 0.21 & 6.01 & 5.23 & 3.36 \\
WyFormer & \underline{0.28} & 0.008 & \underline{0.22} & \textbf{0.74} & \textbf{0.51} & 0.27 & 24.27 & 3.52 & 2.71 \\
WyFormer-DFT & 0.29 & 0.010 & 0.26 & \underline{0.73} & 0.49 & 0.27 & 23.57 & 3.49 & 2.74 \\
\midrule
ADiT & 0.36 & \underline{0.003} & 1.03 & 0.71 & 0.40 & 0.24 & 14.50 & \underline{3.41} & 2.70 \\
Crystal-GFN & 0.64 & 0.092 & 2.94 & \textbf{0.74} & 0.12 & \textbf{0.32} & \textbf{31.50} & 3.45 & \textbf{2.33} \\
Crystalformer & 0.29 & 0.006 & 0.43 & 0.71 & 0.39 & \underline{0.31} & 21.57 & 3.56 & 2.69 \\
DiffCSP & 0.35 & 0.009 & 0.30 & \underline{0.73} & 0.34 & 0.25 & 14.42 & \textbf{3.27} & \underline{2.59} \\
DiffCSP++ & \textbf{0.27} & 0.004 & 0.43 & 0.72 & \underline{0.50} & 0.27 & 20.90 & 3.55 & 2.77 \\
LLaMat2 & \underline{0.28} & \textbf{0.002} & 0.77 & 0.71 & 0.38 & 0.24 & 11.91 & 3.84 & 2.79 \\
LLaMat3 & 0.43 & \underline{0.003} & \textbf{0.15} & 0.72 & 0.17 & 0.27 & \underline{28.26} & 3.48 & 2.93 \\
SymmCD & \underline{0.28} & 0.008 & 0.28 & \underline{0.73} & \underline{0.50} & 0.27 & 20.87 & 3.51 & 2.68 \\
\midrule
Alexandria & 0.14 & 0.000 & 0.01 & 0.71 & 0.50 & 0.26 & 13.50 & 4.02 & 3.33 \\
OQMD & 0.28 & 0.003 & 0.16 & 0.63 & 0.46 & 0.24 & 12.57 & 4.22 & 3.29 \\
\midrule
AFLOW & 0.29 & 0.004 & 0.13 & 0.69 & 0.50 & 0.25 & 10.07 & 4.12 & 3.22 \\
MP-Exp & 0.35 & 0.022 & 0.74 & 0.72 & 0.68 & 0.28 & 54.79 & 2.78 & 2.27 \\
\midrule
RandomEnum & 0.47 & 0.035 & 0.26 & 0.62 & 0.34 & 0.27 & 15.48 & 3.13 & 3.10 \\
IonSub & 0.33 & 0.004 & 0.77 & 0.67 & 0.42 & 0.26 & 10.81 & 4.15 & 3.86 \\
\bottomrule
\end{tabular}
\end{adjustbox}
\end{table}
\FloatBarrier

\subsection{Using a Self-Consistent MLIP-Based Convex Hull}
\label{app:mlip_hull}

A key challenge in evaluating generated structures with Machine-Learned 
Interatomic Potentials (MLIPs) is the accurate calculation of the energy 
above the convex hull ($E_{\text{hull}}$). Mixing total energies from an 
MLIP with a convex hull built from reference DFT calculations introduces 
inconsistencies: the MLIP and DFT methods operate on different energy 
references and possess distinct systematic errors. Since stability thresholds 
are tight (often a few tens of meV/atom), even small systematic shifts 
can change a structure's classification from stable to unstable.

To address this, our benchmark constructs the convex hull using the same 
MLIP that evaluates the candidate structures. This self-consistent 
approach allows systematic errors to cancel, yielding more reliable 
$E_{\text{hull}}$ estimates.

\begin{figure}[h]
    \centering
    \includegraphics[width=1\linewidth]{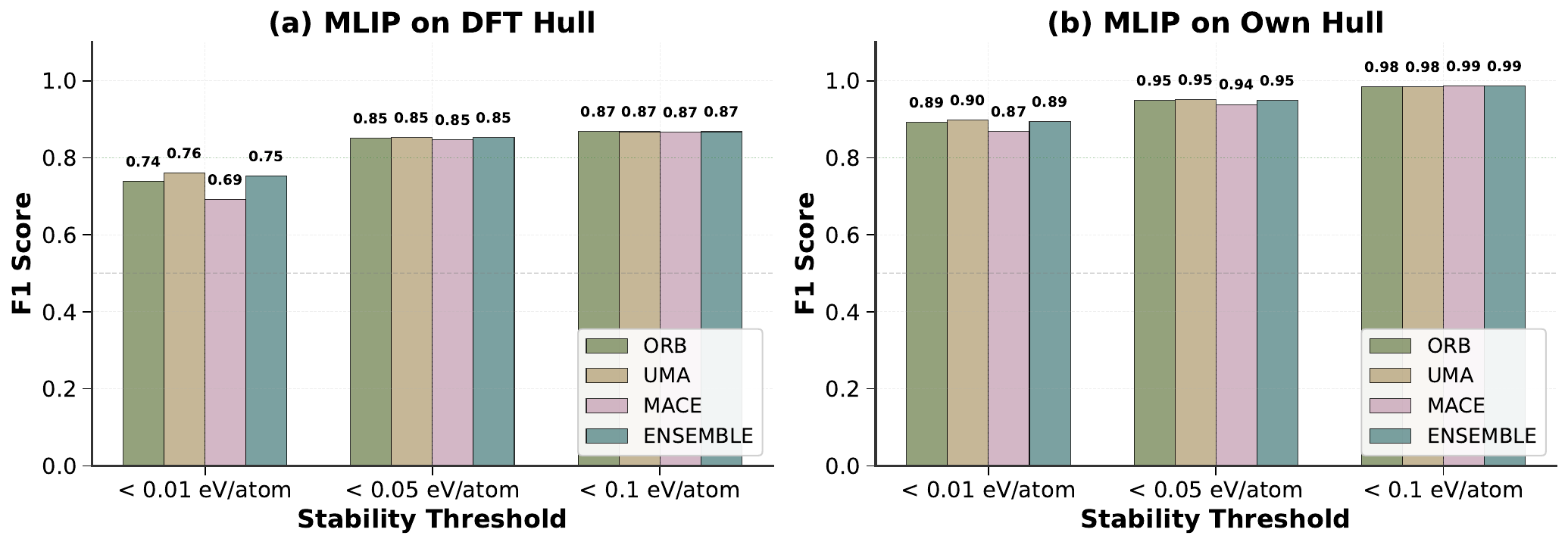}
    \caption{F1-score for stability prediction on MP-20. Format matches \Cref{fig:bulk_hull_f1}.}
    \label{fig:mp20_hull_f1}
\end{figure}

\paragraph{Empirical validation.}
\Cref{fig:bulk_hull_f1} and \Cref{fig:mp20_hull_f1} compare F1-scores for stability 
prediction under the two approaches. On both LeMat-Bulk and MP-20, the 
self-consistent hull (panel~b) consistently outperforms the mixed-reference 
DFT hull (panel~a), with the largest improvements at the strictest 
threshold ($<0.01$~eV/atom). For example, on LeMat-Bulk, the ORB F1-score 
improves from 0.66 to 0.81.

\begin{figure}[hbpt]
    \centering
    \includegraphics[width=1\linewidth]{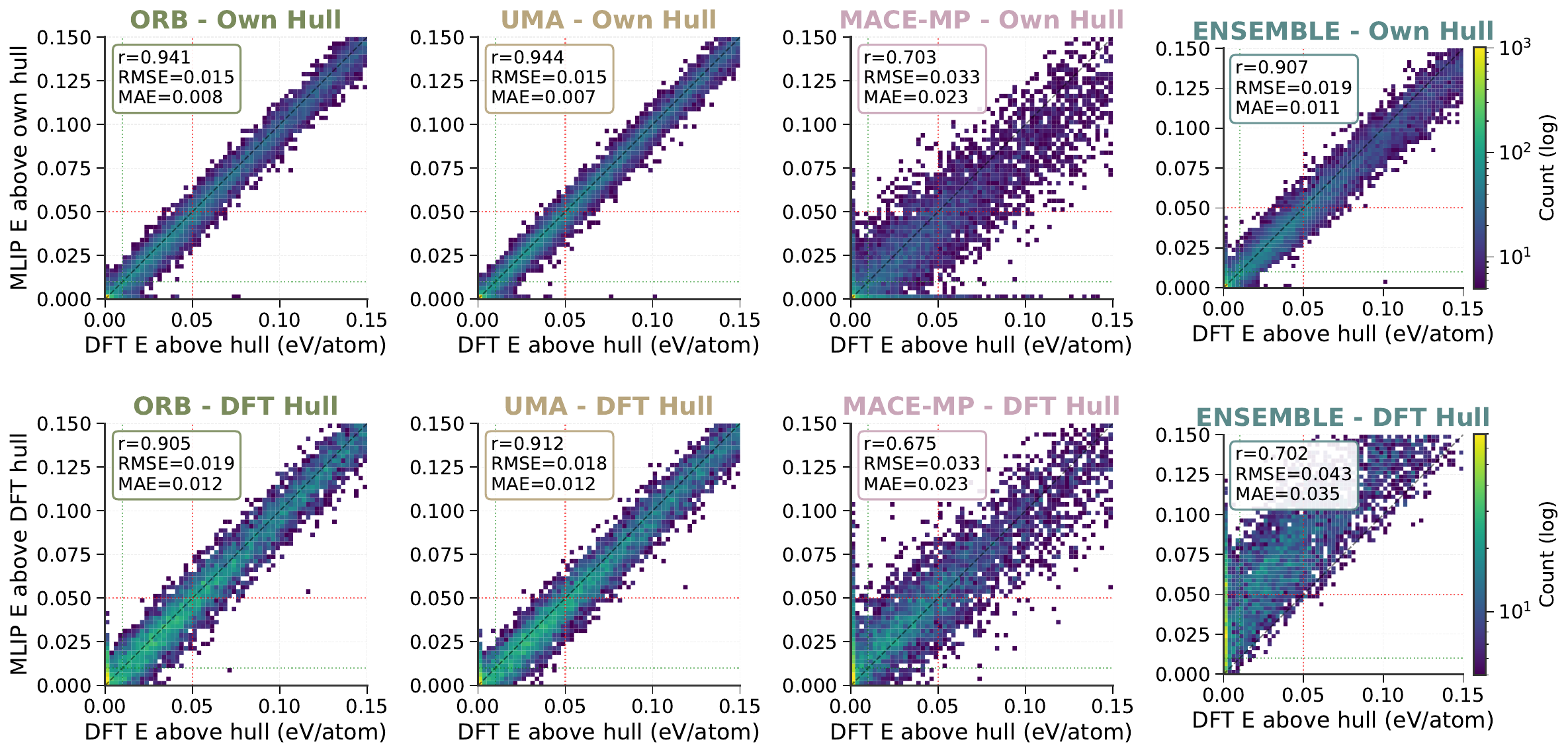}
    \caption{Correlation between MLIP-predicted and DFT-calculated 
    $E_{\text{hull}}$ on LeMat-Bulk. \textit{Top row (Own Hull):} 
    Self-consistent approach yields high correlation ($r > 0.94$ for 
    ORB and UMA), low error (MAE~$\leq 0.008$~eV/atom), and tight 
    clustering around the diagonal. The ensemble (rightmost) achieves 
    intermediate performance (MAE~$=$~0.011~eV/atom). \textit{Bottom 
    row (DFT Hull):} Mixed-reference approach shows degraded correlation 
    and higher MAE. Notably, the ensemble (MAE~$=$~0.035~eV/atom) 
    performs worse than all individual models, illustrating how 
    averaging heterogeneous biases can compound error.}
    \label{fig:bulk_hull_scatter}
\end{figure}

\Cref{fig:bulk_hull_scatter} reveals why. The self-consistent approach 
(top row) yields systematically lower MAE and higher correlation with 
DFT ground truth than the mixed-reference approach (bottom row). For 
ORB, MAE decreases from 0.012 to 0.008~eV/atom; for UMA, from 0.012 to 
0.007~eV/atom.

\paragraph{Ensemble composition.}
Although MACE-MP exhibits higher variance and lower correlation with 
DFT ($r \approx 0.7$) than ORB and UMA ($r > 0.9$), we retain it in 
the ensemble to ensure model diversity. ORB and UMA were both trained 
on OMat24 in addition to Materials Project, making their predictions 
partially correlated; relying on them alone risks collapsing the 
ensemble toward a single effective model. MACE-MP, trained exclusively 
on Materials Project, provides an independent perspective that guards 
against this collapse. On the self-consistent hull, systematic biases 
cancel within each model, and the remaining variance-type errors are 
effectively reduced by averaging across architectures with diverse 
training data. This design choice reflects a deliberate trade-off: 
accepting slightly higher individual-model error in exchange for 
more robust and diverse ensemble estimates.

\FloatBarrier

\subsection{Benchmark Results with MP-20 as the Reference Database}
\label{app:subsec:mp20_results}

\begin{figure}[htbp]
\centering
\begin{subfigure}[b]{0.48\textwidth}
    \centering
    \includegraphics[width=\textwidth, height=0.7\textwidth, keepaspectratio=true]{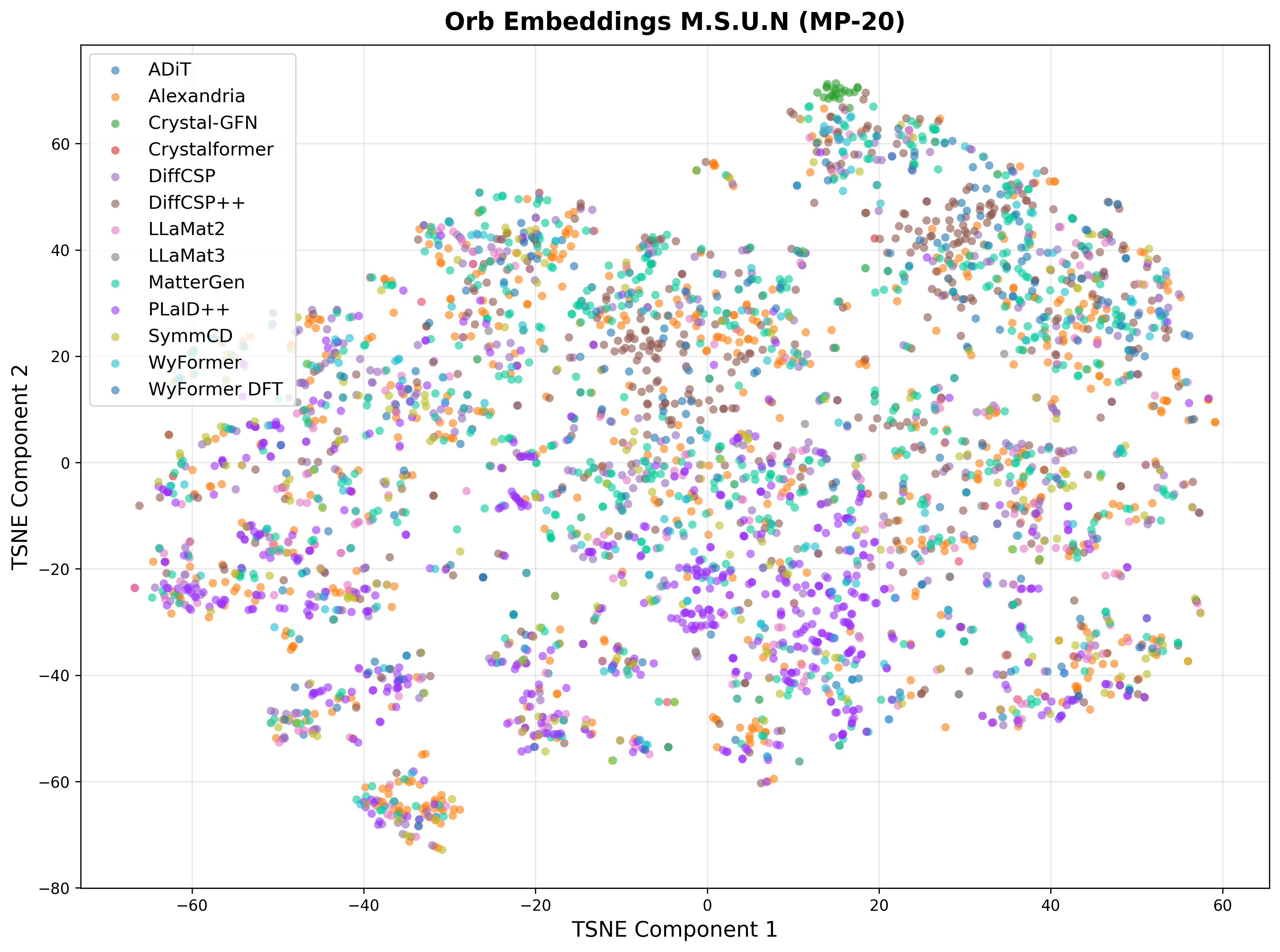}
    \caption{t-SNE visualization of ORB graph embeddings for M.S.U.N. materials.}
    \label{fig:mp20_tsne_msun}
\end{subfigure}
\hfill
\begin{subfigure}[b]{0.48\textwidth}
    \centering
    \includegraphics[width=\textwidth, height=0.7\textwidth, keepaspectratio=true]{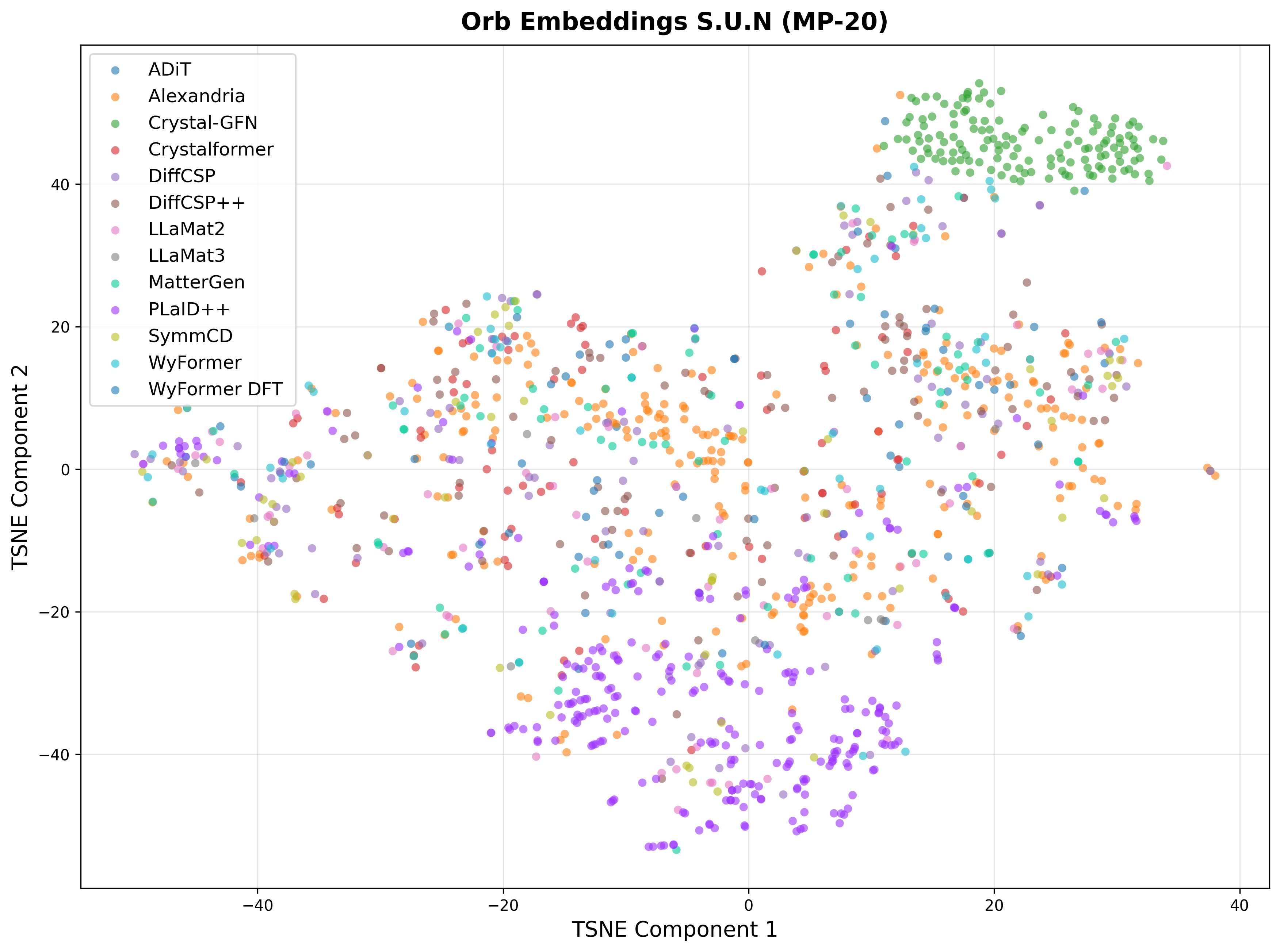}
    \caption{t-SNE visualization of ORB graph embeddings for S.U.N. materials.}
    \label{fig:mp20_tsne_sun}
\end{subfigure}
\caption{Two-dimensional t-SNE visualizations (perplexity = 30) of ORB graph embeddings of generative model structures computed with MP-20 as the reference set. The visualizations demonstrate how most models generate a wide diversity of novel crystal structures that cover a breadth of chemical space.}
\label{fig:mp20_tsne}
\end{figure}

We visualize S.U.N. and M.S.U.N. structures from various generative models using t-SNE, a dimensionality reduction method in~\Cref{fig:mp20_tsne}. Across both plots, most models produce structures that are spread out broadly throughout the embedding space rather than collapsing into small, isolated regions. This suggests that current crystal generators are relatively capable of proposing chemically diverse candidates and are not restricted to narrow pockets of the MP-20 distribution. Although there is some overlap, different models still preferentially populate distinct regions of the embedding manifold. This complementary coverage supports the idea that combining multiple generative models within a discovery pipeline could increase the overall diversity of viable candidates.

One notable pattern appears in \Cref{fig:mp20_tsne_sun}, where Crystal-GFN produces a comparatively tight, well-defined cluster. Despite having the second-highest S.U.N. rate, this indicates the model may be biased toward a narrower slice of chemically stable structures relative to other methods. In contrast, models like PLaID++, WyFormer, and MatterGen reflect broader exploration of the structural landscape.

\begin{figure}[htbp]
\centering
\begin{subfigure}[b]{0.48\textwidth}
    \centering
    \includegraphics[width=\textwidth, height=0.7\textwidth, keepaspectratio=true]{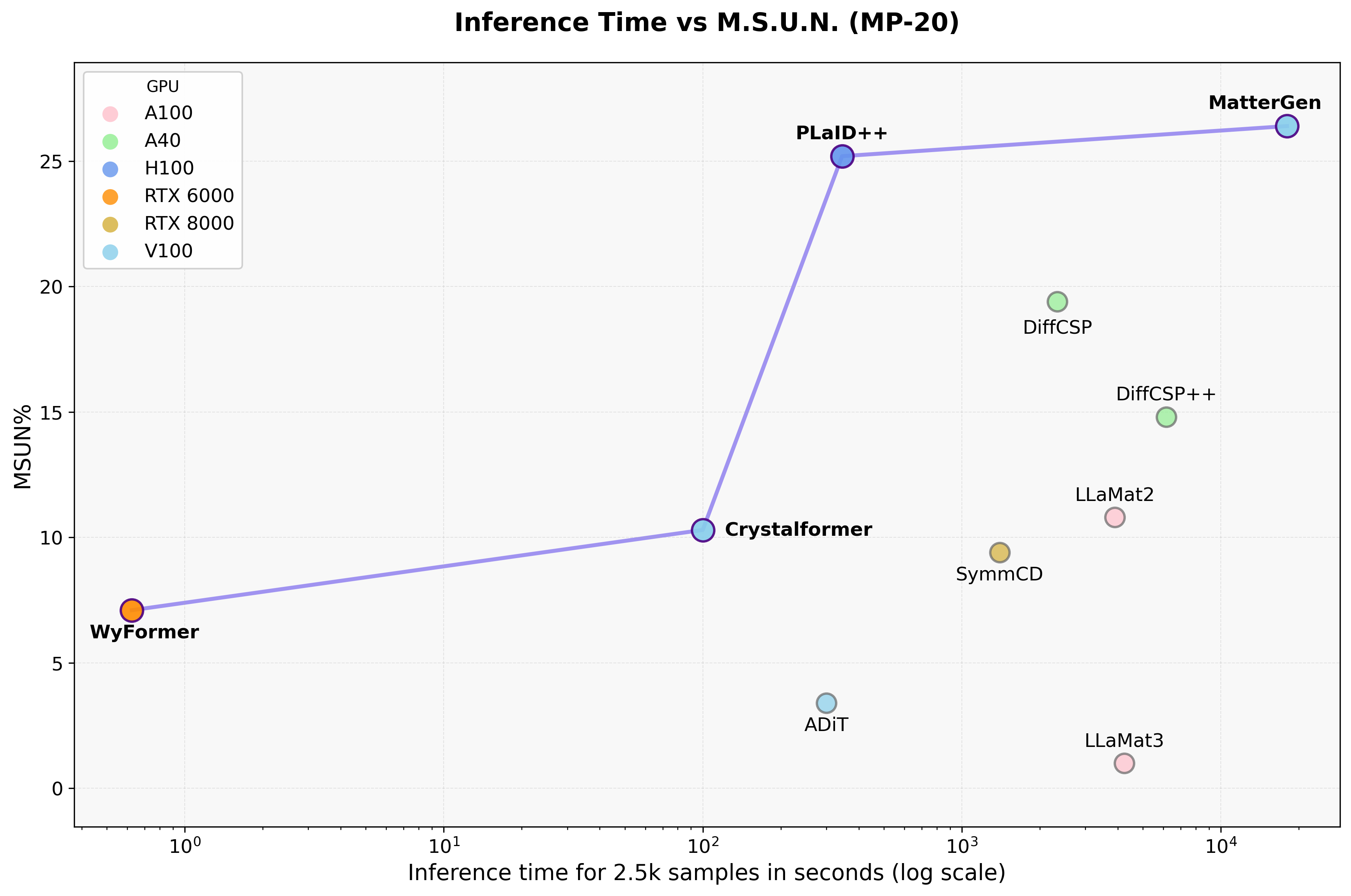}
    \caption{Inference time vs MSUN\%}
    \label{fig:inference_time_msun_mp20}
\end{subfigure}
\hfill
\begin{subfigure}[b]{0.48\textwidth}
    \centering
    \includegraphics[width=\textwidth, height=0.7\textwidth, keepaspectratio=true]{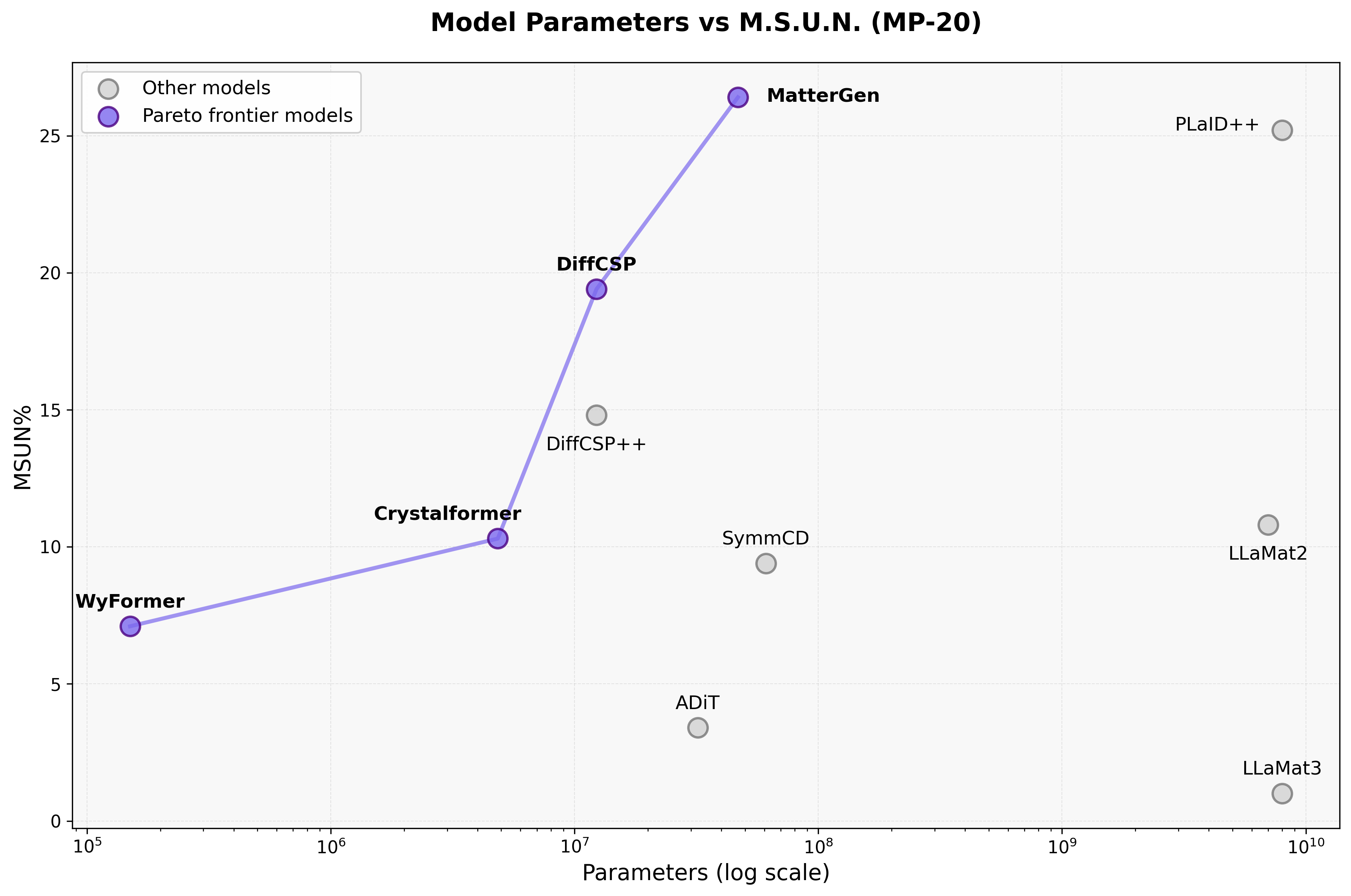}
    \caption{Model parameters vs MSUN\%}
    \label{fig:params_msun_mp20}
\end{subfigure}
\caption{Performance-efficiency trade-offs for crystal structure generation models evaluated on MP-20 measured using MSUN. (a) Inference time (log scale) versus MSUN\% shows the computational cost required to achieve MSUN structures. Models on the Pareto frontier (connected by the blue line) represent optimal time-performance trade-offs. (b) Model parameter count (log scale) versus MSUN\% illustrates the relationship between model size and generation quality. WyFormer, Crystalformer, DiffCSP, and MatterGen form the Pareto frontier.}
\label{fig:efficiency_tradeoffs_mp20}
\end{figure}

\begin{figure}[htbp]
\centering
\begin{subfigure}[b]{0.48\textwidth}
    \centering
    \includegraphics[width=\textwidth, height=0.7\textwidth, keepaspectratio=true]{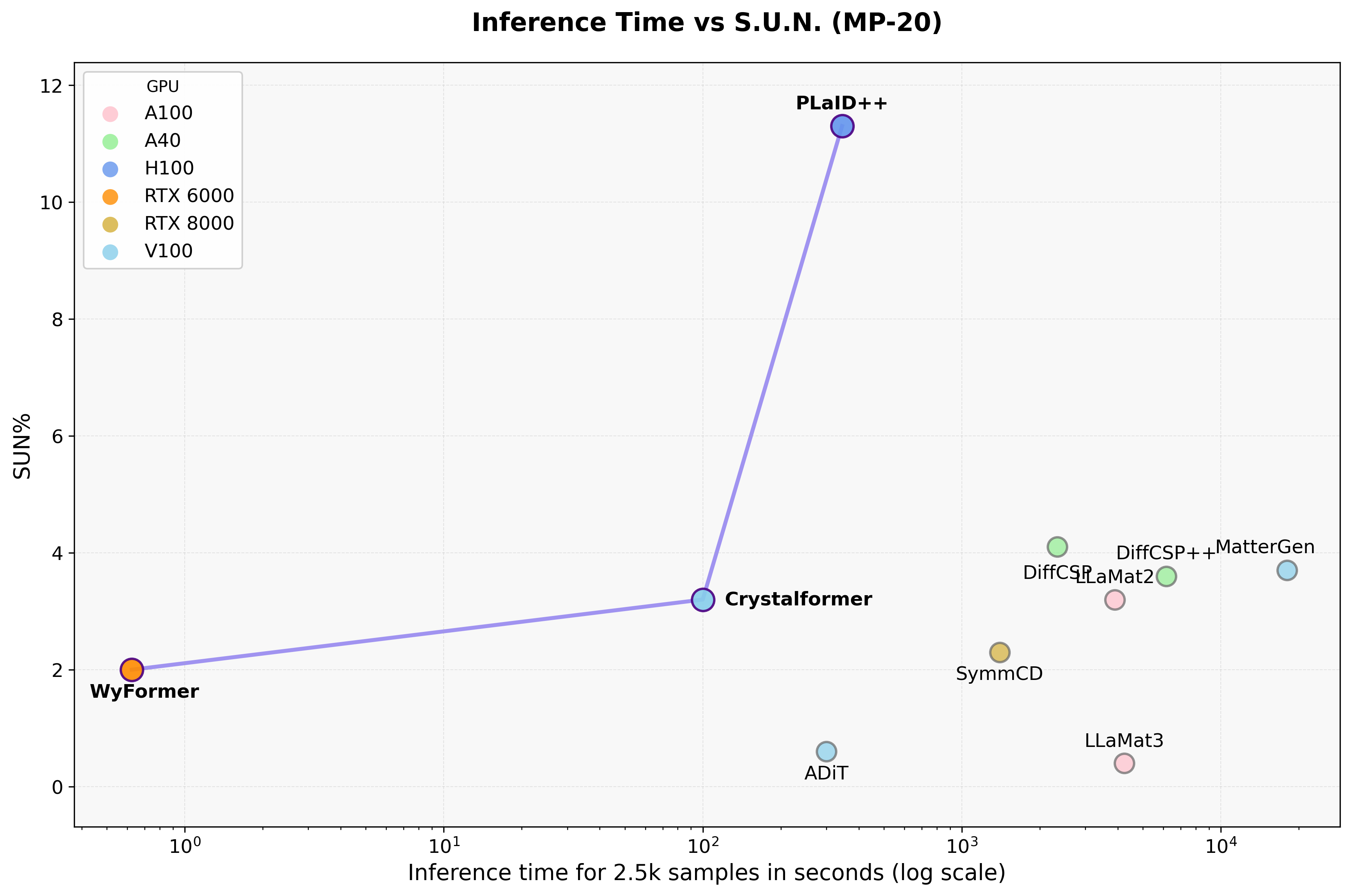}
    \caption{Inference time vs SUN\%}
    \label{fig:inference_time_sun_mp20}
\end{subfigure}
\hfill
\begin{subfigure}[b]{0.48\textwidth}
    \centering
    \includegraphics[width=\textwidth, height=0.7\textwidth, keepaspectratio=true]{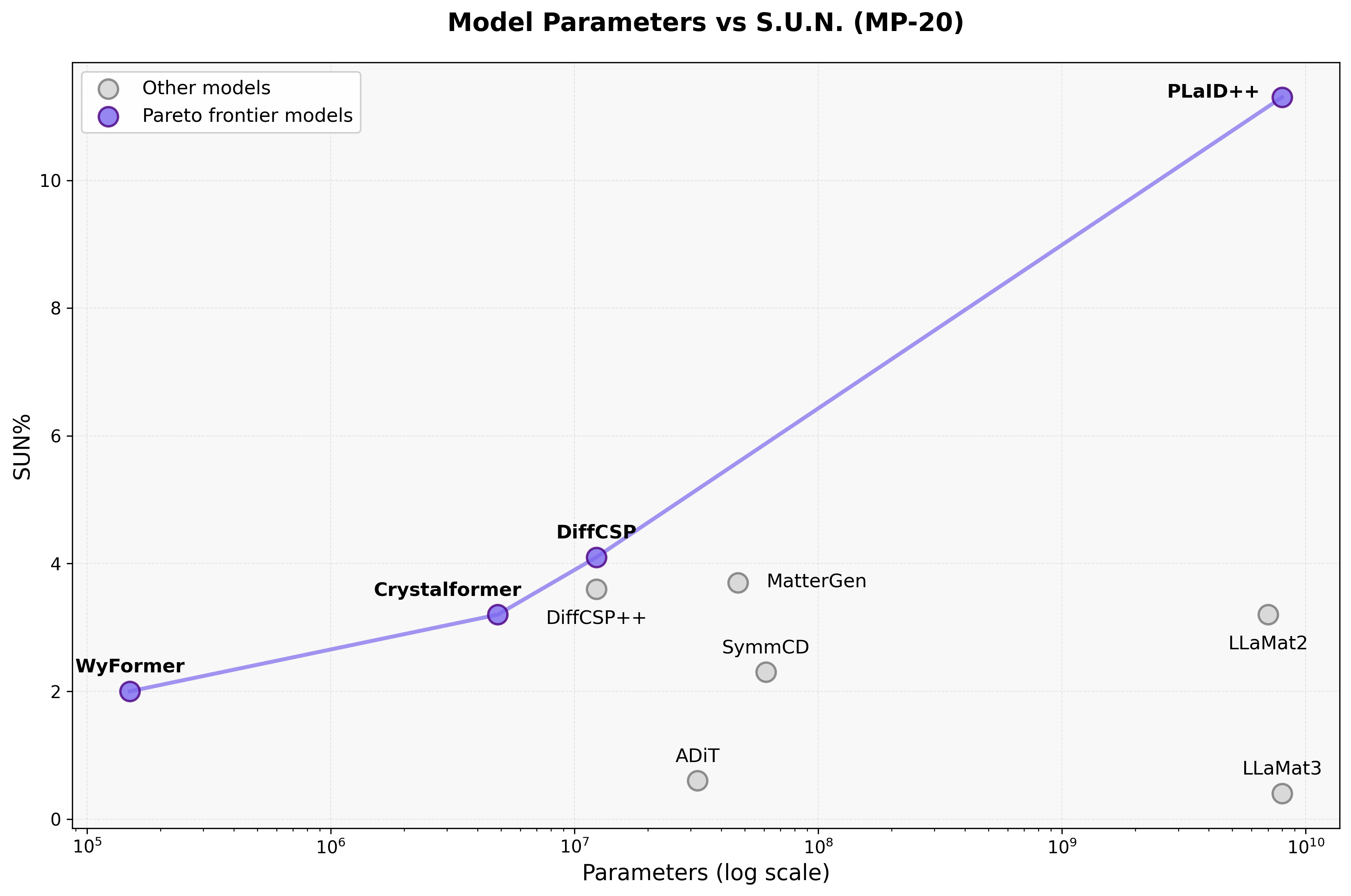}
    \caption{Model parameters vs SUN\%}
    \label{fig:params_sun_mp20}
\end{subfigure}
\caption{Performance-efficiency trade-offs for crystal structure generation models evaluated on MP-20 measured using SUN. (a) Inference time versus S.U.N.\% highlights the computational budget required to produce stable and unique structures. The Pareto frontier shows that WyFormer, Crystalformer, and PLaID++ jointly define the best speed–quality trade-offs, with PLaID++ achieving the highest S.U.N.\% at a moderate additional cost. (b) Model size versus S.U.N.\% illustrates how parameter count correlates with generation quality. With WyFormer, CrystalFormer, DiffCSP, and PLaID++ on the Pareto frontier, this demonstrates that larger models can achieve better performance when properly designed.}

\label{fig:efficiency_tradeoffs_sun_mp20}
\end{figure}

We report model efficiency metrics in~\Cref{fig:efficiency_tradeoffs_mp20} and \Cref{fig:efficiency_tradeoffs_sun_mp20}. The Pareto frontier in inference-time space is again spanned by WyFormer, Crystalformer, and PLaID++, which interpolate between low-latency, moderate-quality generation and slower, higher S.U.N.\% regimes. WyFormer remains at the extreme low-latency corner, whereas PLaID++ attains the highest S.U.N.\% at substantially increased runtime, with MatterGen, DiffCSP, and other approaches lying close to but generally below this curve. 

A similar picture appears when plotting S.U.N.\% versus parameter count. WyFormer and Crystalformer achieve competitive S.U.N.\% with relatively compact models, and
DiffCSP extends this frontier to a moderately larger parameter budget. PLaID++ pushes to the highest S.U.N.\% but does so with a substantially larger model, while SymmCD, ADiT, and the LLaMat variants sit well below the frontier—using one to two orders of magnitude more parameters than the most efficient models yet achieving comparable or worse S.U.N.\%. Since PLaID++ directly optimizes for S.U.N. with respect to MP-20, it is unsurprising that it has the highest S.U.N. rate. At the same time, its performance highlights the promise of reinforcement learning–based optimization, which could be applied to smaller models to improve material discovery rates across a range of model sizes.


\begin{table}[htbp]
\centering
\caption{Model evaluation metrics calculated using \textbf{MLIP ensemble} with \textbf{MP-20} as the reference set: Validity, Energy, Stability, and Metastability. All models provide 2500 structures except Crystalformer (1000 structures). Submissions are organized into five categories separated by horizontal lines: (1) models that incorporate relaxation as part of their generation pipeline (MatterGen, PLaID++, WyFormer, WyFormer-DFT), (2) other generative models, (3) samples from real-world datasets that contribute to LeMat-Bulk (Alexandria, OQMD) as baselines, (4) samples from the AFLOW dataset and Materials Project Experimental structures, (5) pre-relaxed samples from non-ML generative baselines. Best values are shown in \textbf{bold}, second-best values are \underline{underlined}. Arrows indicate whether higher ({\textuparrow}) or lower ({\textdownarrow}) values are better.}
\scriptsize
\begin{adjustbox}{max width=\textwidth}
\begin{tabular}{l cccc c c ccc ccc ccc}
\toprule
\textbf{Model} &
\multicolumn{4}{c}{\textbf{Validity}} &
\textbf{Unique\% {\textuparrow}} & \textbf{Novel\% {\textuparrow}} &
\multicolumn{3}{c}{\textbf{Energy-based}} &
\multicolumn{3}{c}{\textbf{Stability}} &
\multicolumn{3}{c}{\textbf{Metastability}} \\
\cmidrule(lr){2-5} \cmidrule(lr){8-10} \cmidrule(lr){11-13} \cmidrule(lr){14-16}
& \textbf{Valid\% {\textuparrow}} & \textbf{CN\% {\textuparrow}} & \textbf{MinDist\% {\textuparrow}} & \textbf{PhysPlau\% {\textuparrow}}
&  &  & \textbf{FormE (Std) {\textdownarrow}} & \textbf{$E_{\mathrm{hull}}$ (Std) {\textdownarrow}} & \textbf{RMSD (Std) {\textdownarrow}}
& \textbf{Stable\% {\textuparrow}} & \textbf{U-Stable\% {\textuparrow}} & \textbf{SUN\% {\textuparrow}}
& \textbf{Metastable\% {\textuparrow}} & \textbf{U-Meta\% {\textuparrow}} & \textbf{MSUN\% {\textuparrow}} \\
\midrule
MatterGen & \underline{95.7} & 95.9 & \underline{99.8} & \textbf{100.0} & \textbf{95.1} & \underline{83.1} & \meanstd{-0.71}{0.79} & \underline{\meanstd{0.16}{0.17}} & \meanstd{0.39}{0.50} & 4.5 {\scriptsize (113)} & 4.5 {\scriptsize (113)} & 3.7 {\scriptsize (92)} & 37.9 {\scriptsize (947)} & 37.3 {\scriptsize (933)} & \textbf{26.4} {\scriptsize (660)} \\
PLaID++ & \textbf{96.0} & 96.0 & 96.2 & 96.2 & 77.8 & 63.9 & \meanstd{-0.50}{0.44} & $\bm{\meanstd{0.06}{0.16}}$ & $\bm{\meanstd{0.13}{0.29}}$ & \textbf{27.2} {\scriptsize (681)} & \textbf{18.9} {\scriptsize (472)} & \textbf{11.3} {\scriptsize (282)} & \textbf{49.6} {\scriptsize (1239)} & \textbf{40.5} {\scriptsize (1012)} & \underline{25.2} {\scriptsize (630)} \\
WyFormer & 93.4 & 95.2 & 98.2 & \textbf{100.0} & 93.0 & 80.4 & $\meanstd{-0.45}{0.95}$ & $\meanstd{0.46}{0.52}$ & $\meanstd{0.81}{0.97}$ & 2.3 {\scriptsize (57)} & 2.2 {\scriptsize (56)} & 2.0 {\scriptsize (49)} & 16.9 {\scriptsize (422)} & 16.7 {\scriptsize (418)} & 7.1 {\scriptsize (177)} \\
WyFormer-DFT & 95.2 & 95.2 & \textbf{100.0} & \textbf{100.0} & \underline{95.0} & \textbf{83.3} & \meanstd{-0.67}{0.90} & \meanstd{0.24}{0.38} & \meanstd{0.42}{0.59} & \underline{6.8} {\scriptsize (170)} & \underline{6.8} {\scriptsize (169)} & 4.0 {\scriptsize (99)} & 26.4 {\scriptsize (660)} & 26.2 {\scriptsize (656)} & 17.4 {\scriptsize (436)} \\
\midrule
ADiT & 90.6 & \textbf{99.0} & 91.5 & \textbf{100.0} & 87.8 & 31.3 & \underline{\meanstd{-0.74}{0.93}} & \meanstd{0.31}{0.44} & \underline{\meanstd{0.38}{0.40}} & 0.8 {\scriptsize (20)} & 0.8 {\scriptsize (20)} & 0.6 {\scriptsize (16)} & \underline{40.3} {\scriptsize (1008)} & \underline{38.6} {\scriptsize (965)} & 3.4 {\scriptsize (84)} \\
Crystal-GFN & 51.7 & 58.3 & 91.1 & 96.3 & 51.7 & 51.7 & $\bm{\meanstd{-1.34}{2.24}}$ & $\meanstd{1.61}{2.06}$ & $\meanstd{1.86}{0.90}$ & 5.7 {\scriptsize (143)} & 5.7 {\scriptsize (143)} & \underline{5.7} {\scriptsize (143)} & 0.8 {\scriptsize (20)} & 0.8 {\scriptsize (20)} & 0.8 {\scriptsize (20)} \\
Crystalformer & 69.9 & 96.3 & 72.9 & \underline{99.9} & 69.4 & 44.2 & $\meanstd{-0.18}{1.47}$ & $\meanstd{0.67}{1.32}$ & $\meanstd{0.66}{1.03}$ & 4.4 {\scriptsize (44)} & 4.4 {\scriptsize (44)} & 3.2 {\scriptsize (32)} & 29.3 {\scriptsize (293)} & 29.1 {\scriptsize (291)} & 10.3 {\scriptsize (103)} \\
DiffCSP & \underline{95.7} & 96.2 & 99.5 & \textbf{100.0} & 94.8 & 81.9 & \meanstd{-0.65}{0.97} & \meanstd{0.24}{0.57} & \meanstd{0.59}{0.63} & 5.7 {\scriptsize (143)} & 5.6 {\scriptsize (140)} & 4.1 {\scriptsize (103)} & 31.4 {\scriptsize (785)} & 30.7 {\scriptsize (768)} & 19.4 {\scriptsize (485)} \\
DiffCSP++ & 95.3 & 95.5 & 99.7 & \textbf{100.0} & \textbf{95.1} & 81.1 & \meanstd{-0.53}{0.87} & \meanstd{0.37}{0.43} & \meanstd{0.70}{0.83} & 4.6 {\scriptsize (114)} & 4.6 {\scriptsize (114)} & 3.6 {\scriptsize (89)} & 27.4 {\scriptsize (686)} & 27.2 {\scriptsize (681)} & 14.8 {\scriptsize (371)} \\
LLaMat2 & 84.4 & \underline{97.0} & 87.1 & 99.6 & 81.4 & 53.0 & $\meanstd{-0.48}{1.00}$ & $\meanstd{0.41}{0.70}$ & $\meanstd{0.54}{0.71}$ & 3.9 {\scriptsize (98)} & 3.8 {\scriptsize (96)} & 3.2 {\scriptsize (81)} & 35.7 {\scriptsize (893)} & 33.5 {\scriptsize (838)} & 10.8 {\scriptsize (270)} \\
LLaMat3 & 15.4 & 82.5 & 16.8 & 84.2 & 15.2 & 13.8 & $\meanstd{0.69}{2.61}$ & $\meanstd{1.62}{2.39}$ & $\meanstd{1.01}{1.10}$ & 0.4 {\scriptsize (10)} & 0.4 {\scriptsize (10)} & 0.4 {\scriptsize (10)} & 2.3 {\scriptsize (57)} & 2.2 {\scriptsize (55)} & 1.0 {\scriptsize (25)} \\
SymmCD & 73.4 & 95.9 & 76.4 & \textbf{100.0} & 73.0 & 61.3 & $\meanstd{-0.04}{1.21}$ & $\meanstd{0.84}{1.06}$ & $\meanstd{0.87}{1.10}$ & 3.1 {\scriptsize (78)} & 3.1 {\scriptsize (77)} & 2.3 {\scriptsize (57)} & 19.5 {\scriptsize (488)} & 19.3 {\scriptsize (482)} & 9.4 {\scriptsize (236)} \\
\midrule
Alexandria & 93.4 & 93.4 & 99.0 & 99.0 & 93.4 & 92.4 & \meanstd{-0.19}{0.94} & \meanstd{0.40}{0.61} & \meanstd{0.15}{0.36} & 10.2 {\scriptsize (255)} & 10.2 {\scriptsize (255)} & 10.1 {\scriptsize (252)} & 25.8 {\scriptsize (645)} & 25.8 {\scriptsize (645)} & 25.0 {\scriptsize (626)} \\
OQMD & 96.8 & 96.8 & 99.4 & 99.5 & 96.4 & 94.2 & \meanstd{-0.24}{0.88} & \meanstd{0.35}{0.46} & \meanstd{0.14}{0.28} & 16.3 {\scriptsize (408)} & 16.3 {\scriptsize (407)} & 15.7 {\scriptsize (392)} & 21.4 {\scriptsize (535)} & 21.3 {\scriptsize (533)} & 19.4 {\scriptsize (486)} \\
\midrule
AFLOW & 91.4 & 91.4 & 100.0 & 100.0 & 87.1 & 57.0 & \meanstd{-0.19}{0.86} & \meanstd{0.32}{0.42} & \meanstd{0.12}{0.44} & 10.6 {\scriptsize (265)} & 10.0 {\scriptsize (249)} & 2.8 {\scriptsize (70)} & 30.8 {\scriptsize (771)} & 28.0 {\scriptsize (699)} & 4.5 {\scriptsize (113)} \\
MP-Exp & 96.9 & 97.1 & 98.8 & 98.9 & 96.8 & 66.8 & \meanstd{-1.10}{1.11} & \meanstd{0.15}{0.54} & \meanstd{0.20}{0.36} & 18.6 {\scriptsize (464)} & 18.6 {\scriptsize (464)} & 15.4 {\scriptsize (386)} & 50.4 {\scriptsize (1261)} & 50.4 {\scriptsize (1260)} & 24.7 {\scriptsize (618)} \\
\midrule
RandomEnum & 97.0 & 99.0 & 97.8 & 99.8 & 97.0 & 96.6 & \meanstd{-0.65}{0.84} & \meanstd{0.18}{0.18} & \meanstd{0.81}{0.66} & 5.4 {\scriptsize (136)} & 5.4 {\scriptsize (136)} & 5.4 {\scriptsize (135)} & 18.8 {\scriptsize (469)} & 18.8 {\scriptsize (469)} & 18.4 {\scriptsize (460)} \\
IonSub & 99.2 & 99.3 & 100.0 & 100.0 & 91.6 & 77.5 & \meanstd{-0.43}{0.40} & \meanstd{0.01}{0.10} & \meanstd{0.07}{0.19} & 42.4 {\scriptsize (1059)} & 39.8 {\scriptsize (994)} & 33.5 {\scriptsize (837)} & 46.4 {\scriptsize (1161)} & 42.4 {\scriptsize (1061)} & 29.8 {\scriptsize (746)} \\
\bottomrule
\end{tabular}
\end{adjustbox}
\end{table}

\begin{table}[htbp]
\centering
\caption{Model evaluation metrics calculated using \textbf{MLIP ensemble} with \textbf{MP-20} as the reference set: Distribution, Diversity, and HHI. All models provide 2500 structures except Crystalformer (1000 structures). Submissions are organized into five categories separated by horizontal lines: (1) models that incorporate relaxation as part of their generation pipeline (MatterGen, PLaID++, WyFormer, WyFormer-DFT), (2) other generative models, (3) samples from real-world datasets that contribute to LeMat-Bulk (Alexandria, OQMD) as baselines, (4) samples from the AFLOW dataset and MP-Exp structures, (5) pre-relaxed samples from non-ML generative baselines. Best values are shown in \textbf{bold}, second-best values are \underline{underlined}. Arrows indicate whether higher ({\textuparrow}) or lower ({\textdownarrow}) values are better.}
\scriptsize
\begin{adjustbox}{max width=\textwidth}
\begin{tabular}{l ccc cccc cc}
\toprule
\textbf{Model} &
\multicolumn{3}{c}{\textbf{Distribution}} &
\multicolumn{4}{c}{\textbf{Diversity}} &
\multicolumn{2}{c}{\textbf{HHI}} \\
\cmidrule(lr){2-4} \cmidrule(lr){5-8} \cmidrule(lr){9-10}
& \textbf{JS {\textdownarrow}} & \textbf{MMD {\textdownarrow}} & \textbf{FID {\textdownarrow}}
& \textbf{ElemDiv {\textuparrow}} & \textbf{SGDiv {\textuparrow}} & \textbf{SizeDiv {\textuparrow}} & \textbf{SiteDiv {\textuparrow}}
& \textbf{Prod {\textdownarrow}} & \textbf{Res {\textdownarrow}} \\
\midrule
MatterGen & 0.43 & \textbf{0.000} & \textbf{14.38} & 0.64 & 0.14 & 0.21 & 12.78 & 3.52 & 2.72 \\
PLaID++ & 0.48 & 0.049 & 15.48 & 0.68 & 0.26 & 0.21 & 6.01 & 5.23 & 3.36 \\
WyFormer & \underline{0.42} & \underline{0.001} & 15.03 & \underline{0.74} & \underline{0.50} & 0.27 & \underline{23.84} & 3.55 & 2.73 \\
WyFormer-DFT & \underline{0.42} & 0.002 & 21.23 & 0.73 & 0.49 & 0.26 & 23.56 & 3.49 & 2.74 \\
\midrule
ADiT & 0.43 & \underline{0.001} & \underline{14.41} & 0.71 & 0.03 & 0.24 & 14.18 & 3.52 & 2.78 \\
Crystal-GFN & 0.56 & 0.098 & 50.20 & \textbf{0.75} & 0.29 & \textbf{0.32} & \textbf{32.06} & \underline{3.48} & \textbf{2.34} \\
Crystalformer & \textbf{0.41} & 0.002 & 16.60 & 0.71 & 0.35 & \underline{0.31} & 18.59 & 3.78 & 2.82 \\
DiffCSP & 0.44 & 0.002 & 14.73 & 0.73 & 0.13 & 0.24 & 14.42 & \textbf{3.29} & \underline{2.60} \\
DiffCSP++ & \underline{0.42} & 0.003 & 14.97 & 0.72 & \textbf{0.51} & 0.27 & 20.84 & 3.56 & 2.77 \\
LLaMat2 & \textbf{0.41} & 0.003 & 14.91 & 0.71 & 0.25 & 0.24 & 9.86 & 3.94 & 2.86 \\
LLaMat3 & 0.45 & 0.012 & 27.67 & 0.70 & 0.10 & 0.30 & 13.15 & 3.83 & 2.81 \\
SymmCD & \textbf{0.41} & \underline{0.001} & 16.33 & 0.73 & 0.49 & 0.27 & 19.45 & 3.54 & 2.74 \\
\midrule
Alexandria & 0.38 & 0.004 & 14.82 & 0.71 & 0.50 & 0.25 & 13.50 & 4.02 & 3.33 \\
OQMD & 0.43 & 0.014 & 15.32 & 0.63 & 0.48 & 0.24 & 12.57 & 4.22 & 3.29 \\
\midrule
AFLOW & 0.44 & 0.003 & 15.34 & 0.69 & 0.46 & 0.25 & 10.07 & 4.12 & 3.22 \\
MP-Exp & 0.49 & 0.011 & 14.73 & 0.72 & 0.71 & 0.28 & 54.79 & 2.78 & 2.27 \\
\midrule
RandomEnum & 0.47 & 0.016 & 14.84 & 0.62 & 0.23 & 0.27 & 15.48 & 3.12 & 3.09 \\
IonSub & 0.46 & 0.010 & 15.39 & 0.67 & 0.36 & 0.26 & 10.82 & 4.15 & 3.86 \\
\bottomrule
\end{tabular}
\end{adjustbox}
\end{table}

\begin{table}[htbp]
\centering
\caption{Model evaluation metrics calculated using \textbf{uma} with \textbf{MP-20} as the reference set: Validity, Energy, Stability, and Metastability. All models provide 2500 structures except Crystalformer (1000 structures). Submissions are organized into five categories separated by horizontal lines: (1) models that incorporate relaxation as part of their generation pipeline (MatterGen, PLaID++, WyFormer, WyFormer-DFT), (2) other generative models, (3) samples from real-world datasets that contribute to LeMat-Bulk (Alexandria, OQMD) as baselines, (4) samples from the AFLOW dataset and MP-Exp structures, (5) pre-relaxed samples from non-ML generative baselines. Best values are shown in \textbf{bold}, second-best values are \underline{underlined}. Arrows indicate whether higher ({\textuparrow}) or lower ({\textdownarrow}) values are better.}
\scriptsize
\begin{adjustbox}{max width=\textwidth}
\begin{tabular}{l cccc c c ccc ccc ccc}
\toprule
\textbf{Model} &
\multicolumn{4}{c}{\textbf{Validity}} &
\textbf{Unique\% {\textuparrow}} & \textbf{Novel\% {\textuparrow}} &
\multicolumn{3}{c}{\textbf{Energy-based}} &
\multicolumn{3}{c}{\textbf{Stability}} &
\multicolumn{3}{c}{\textbf{Metastability}} \\
\cmidrule(lr){2-5} \cmidrule(lr){8-10} \cmidrule(lr){11-13} \cmidrule(lr){14-16}
& \textbf{Valid\% {\textuparrow}} & \textbf{CN\% {\textuparrow}} & \textbf{MinDist\% {\textuparrow}} & \textbf{PhysPlau\% {\textuparrow}}
&  &  & \textbf{FormE (Std) {\textdownarrow}} & \textbf{$E_{\mathrm{hull}}$ (Std) {\textdownarrow}} & \textbf{RMSD (Std) {\textdownarrow}}
& \textbf{Stable\% {\textuparrow}} & \textbf{U-Stable\% {\textuparrow}} & \textbf{SUN\% {\textuparrow}}
& \textbf{Metastable\% {\textuparrow}} & \textbf{U-Meta\% {\textuparrow}} & \textbf{MSUN\% {\textuparrow}} \\
\midrule
MatterGen & \underline{95.7} & 95.9 & \underline{99.8} & \textbf{100.0} & \textbf{95.1} & \underline{83.1} & \underline{\meanstd{-0.69}{0.80}} & \underline{\meanstd{0.17}{0.19}} & \meanstd{0.40}{0.60} & 5.4 {\scriptsize (136)} & 5.4 {\scriptsize (135)} & 3.5 {\scriptsize (88)} & 33.4 {\scriptsize (835)} & 33.0 {\scriptsize (824)} & \textbf{23.1} {\scriptsize (578)} \\
PLaID++ & \textbf{96.0} & 96.0 & 96.2 & 96.2 & 77.8 & 63.9 & \meanstd{-0.48}{0.48} & $\bm{\meanstd{0.07}{0.17}}$ & $\bm{\meanstd{0.12}{0.32}}$ & \textbf{29.6} {\scriptsize (741)} & \textbf{20.6} {\scriptsize (514)} & \textbf{12.0} {\scriptsize (299)} & \textbf{45.6} {\scriptsize (1141)} & \textbf{37.4} {\scriptsize (935)} & \textbf{23.1} {\scriptsize (578)} \\
WyFormer & 93.4 & 95.2 & 98.2 & \textbf{100.0} & 93.0 & 80.4 & \meanstd{-0.41}{0.97} & \meanstd{0.49}{0.54} & \meanstd{0.80}{1.05} & 2.4 {\scriptsize (61)} & 2.4 {\scriptsize (60)} & 1.9 {\scriptsize (47)} & 16.2 {\scriptsize (405)} & 16.0 {\scriptsize (401)} & 6.7 {\scriptsize (168)} \\
WyFormer-DFT & 95.2 & 95.2 & \textbf{100.0} & \textbf{100.0} & \underline{95.0} & \textbf{83.3} & \meanstd{-0.65}{1.07} & \meanstd{0.26}{0.66} & \meanstd{0.39}{0.60} & \underline{7.5} {\scriptsize (187)} & \underline{7.4} {\scriptsize (184)} & 3.8 {\scriptsize (94)} & 24.4 {\scriptsize (611)} & 24.4 {\scriptsize (609)} & 16.4 {\scriptsize (409)} \\
\midrule
ADiT & 90.6 & \textbf{99.0} & 91.5 & \textbf{100.0} & 87.8 & 31.3 & $\bm{\meanstd{-0.72}{0.95}}$ & \meanstd{0.32}{0.45} & \underline{\meanstd{0.38}{0.41}} & 1.0 {\scriptsize (26)} & 1.0 {\scriptsize (26)} & 0.6 {\scriptsize (16)} & \underline{38.2} {\scriptsize (954)} & \underline{36.6} {\scriptsize (915)} & 3.0 {\scriptsize (76)} \\
Crystal-GFN & 51.7 & 58.3 & 91.1 & 96.3 & 51.7 & 51.7 & \meanstd{-0.56}{4.46} & \meanstd{2.38}{4.28} & \meanstd{1.86}{1.10} & 5.4 {\scriptsize (134)} & 5.4 {\scriptsize (134)} & \underline{5.4} {\scriptsize (134)} & 0.8 {\scriptsize (19)} & 0.8 {\scriptsize (19)} & 0.8 {\scriptsize (19)} \\
Crystalformer & 69.9 & 96.3 & 72.9 & \underline{99.9} & 69.4 & 44.2 & \meanstd{-0.15}{1.48} & \meanstd{0.69}{1.33} & \meanstd{0.68}{1.24} & 5.1 {\scriptsize (51)} & 5.1 {\scriptsize (51)} & 3.2 {\scriptsize (32)} & 27.9 {\scriptsize (279)} & 27.8 {\scriptsize (278)} & 9.5 {\scriptsize (95)} \\
DiffCSP & \underline{95.7} & 96.2 & 99.5 & \textbf{100.0} & 94.8 & 81.9 & \meanstd{-0.62}{0.98} & \meanstd{0.26}{0.58} & \meanstd{0.59}{0.65} & 6.4 {\scriptsize (160)} & 6.2 {\scriptsize (155)} & 3.8 {\scriptsize (95)} & 28.8 {\scriptsize (719)} & 28.2 {\scriptsize (704)} & \underline{17.8} {\scriptsize (445)} \\
DiffCSP++ & 95.3 & 95.5 & 99.7 & \textbf{100.0} & \textbf{95.1} & 81.1 & \meanstd{-0.50}{0.89} & \meanstd{0.40}{0.45} & \meanstd{0.69}{0.88} & 4.7 {\scriptsize (118)} & 4.7 {\scriptsize (118)} & 3.2 {\scriptsize (79)} & 25.6 {\scriptsize (640)} & 25.4 {\scriptsize (635)} & 13.6 {\scriptsize (341)} \\
LLaMat2 & 84.4 & \underline{97.0} & 87.1 & 99.6 & 81.4 & 53.0 & \meanstd{-0.45}{1.03} & \meanstd{0.43}{0.72} & \meanstd{0.54}{0.75} & 4.6 {\scriptsize (116)} & 4.5 {\scriptsize (112)} & 3.2 {\scriptsize (80)} & 33.3 {\scriptsize (832)} & 31.3 {\scriptsize (782)} & 9.7 {\scriptsize (243)} \\
LLaMat3 & 15.4 & 82.5 & 16.8 & 84.2 & 15.2 & 13.8 & \meanstd{0.79}{2.73} & \meanstd{1.72}{2.52} & \meanstd{1.01}{1.26} & 0.4 {\scriptsize (10)} & 0.4 {\scriptsize (10)} & 0.4 {\scriptsize (9)} & 2.2 {\scriptsize (55)} & 2.2 {\scriptsize (54)} & 1.0 {\scriptsize (24)} \\
SymmCD & 73.4 & 95.9 & 76.4 & \textbf{100.0} & 73.0 & 61.3 & \meanstd{-0.00}{1.23} & \meanstd{0.87}{1.08} & \meanstd{0.85}{1.13} & 3.7 {\scriptsize (92)} & 3.6 {\scriptsize (91)} & 2.3 {\scriptsize (57)} & 18.2 {\scriptsize (456)} & 18.0 {\scriptsize (451)} & 8.8 {\scriptsize (219)} \\
\midrule
Alexandria & 93.4 & 93.4 & 99.0 & 99.0 & 93.4 & 92.4 & \meanstd{-0.16}{0.97} & \meanstd{0.42}{0.64} & \meanstd{0.12}{0.40} & 9.9 {\scriptsize (247)} & 9.9 {\scriptsize (247)} & 9.8 {\scriptsize (245)} & 24.5 {\scriptsize (612)} & 24.5 {\scriptsize (612)} & 23.7 {\scriptsize (592)} \\
OQMD & 96.8 & 96.8 & 99.4 & 99.5 & 96.4 & 94.2 & \meanstd{-0.21}{0.89} & \meanstd{0.37}{0.47} & \meanstd{0.13}{0.29} & 17.2 {\scriptsize (429)} & 17.1 {\scriptsize (428)} & 16.3 {\scriptsize (407)} & 19.6 {\scriptsize (491)} & 19.6 {\scriptsize (489)} & 18.0 {\scriptsize (449)} \\
\midrule
AFLOW & 91.4 & 91.4 & 100.0 & 100.0 & 87.1 & 57.0 & \meanstd{-0.18}{0.86} & \meanstd{0.33}{0.42} & \meanstd{0.11}{0.40} & 12.9 {\scriptsize (322)} & 12.1 {\scriptsize (302)} & 2.6 {\scriptsize (66)} & 28.0 {\scriptsize (699)} & 25.3 {\scriptsize (632)} & 4.5 {\scriptsize (112)} \\
MP-Exp & 96.9 & 97.1 & 98.8 & 98.9 & 96.8 & 66.8 & \meanstd{-1.07}{1.32} & \meanstd{0.18}{0.88} & \meanstd{0.22}{0.43} & 20.0 {\scriptsize (499)} & 20.0 {\scriptsize (499)} & 14.8 {\scriptsize (371)} & 47.5 {\scriptsize (1187)} & 47.4 {\scriptsize (1186)} & 23.9 {\scriptsize (597)} \\
\midrule
RandomEnum & 97.0 & 99.0 & 97.8 & 99.8 & 97.0 & 96.6 & \meanstd{-0.62}{0.86} & \meanstd{0.21}{0.20} & \meanstd{0.81}{0.78} & 5.0 {\scriptsize (126)} & 5.0 {\scriptsize (126)} & 5.0 {\scriptsize (125)} & 17.8 {\scriptsize (445)} & 17.8 {\scriptsize (445)} & 17.4 {\scriptsize (436)} \\
IonSub & 99.2 & 99.3 & 100.0 & 100.0 & 91.6 & 77.5 & \meanstd{-0.42}{0.43} & \meanstd{0.02}{0.11} & \meanstd{0.06}{0.16} & 46.0 {\scriptsize (1149)} & 43.2 {\scriptsize (1081)} & 35.7 {\scriptsize (893)} & 41.6 {\scriptsize (1040)} & 37.9 {\scriptsize (947)} & 26.6 {\scriptsize (664)} \\
\bottomrule
\end{tabular}
\end{adjustbox}
\end{table}

\begin{table}[htbp]
\centering
\caption{Model evaluation metrics calculated using \textbf{uma} with \textbf{MP-20} as the reference set: Distribution, Diversity, and HHI. All models provide 2500 structures except Crystalformer (1000 structures). Submissions are organized into five categories separated by horizontal lines: (1) models that incorporate relaxation as part of their generation pipeline (MatterGen, PLaID++, WyFormer, WyFormer-DFT), (2) other generative models, (3) samples from real-world datasets that contribute to LeMat-Bulk (Alexandria, OQMD) as baselines, (4) samples from the AFLOW dataset and MP-Exp structures, (5) pre-relaxed samples from non-ML generative baselines. Best values are shown in \textbf{bold}, second-best values are \underline{underlined}. Arrows indicate whether higher ({\textuparrow}) or lower ({\textdownarrow}) values are better.}
\scriptsize
\begin{adjustbox}{max width=\textwidth}
\begin{tabular}{l ccc cccc cc}
\toprule
\textbf{Model} &
\multicolumn{3}{c}{\textbf{Distribution}} &
\multicolumn{4}{c}{\textbf{Diversity}} &
\multicolumn{2}{c}{\textbf{HHI}} \\
\cmidrule(lr){2-4} \cmidrule(lr){5-8} \cmidrule(lr){9-10}
& \textbf{JS {\textdownarrow}} & \textbf{MMD {\textdownarrow}} & \textbf{FID {\textdownarrow}}
& \textbf{ElemDiv {\textuparrow}} & \textbf{SGDiv {\textuparrow}} & \textbf{SizeDiv {\textuparrow}} & \textbf{SiteDiv {\textuparrow}}
& \textbf{Prod {\textdownarrow}} & \textbf{Res {\textdownarrow}} \\
\midrule
MatterGen & 0.43 & \textbf{0.000} & \underline{0.18} & 0.64 & 0.14 & 0.24 & 12.78 & 3.52 & 2.72 \\
PLaID++ & 0.48 & 0.049 & 0.30 & 0.68 & 0.26 & 0.21 & 6.01 & 5.23 & 3.36 \\
WyFormer & \underline{0.42} & \underline{0.001} & 1.24 & \underline{0.74} & \underline{0.50} & 0.27 & \underline{23.84} & 3.55 & 2.73 \\
WyFormer-DFT & \underline{0.42} & 0.002 & \underline{0.18} & 0.73 & 0.49 & 0.26 & 23.56 & 3.49 & 2.74 \\
\midrule
ADiT & 0.43 & \underline{0.001} & 0.37 & 0.71 & 0.03 & 0.24 & 14.18 & 3.52 & 2.78 \\
Crystal-GFN & 0.56 & 0.098 & 1.70 & \textbf{0.75} & 0.29 & \textbf{0.32} & \textbf{32.06} & \underline{3.48} & \textbf{2.34} \\
Crystalformer & \textbf{0.41} & 0.002 & 2.24 & 0.71 & 0.35 & \underline{0.31} & 18.59 & 3.78 & 2.82 \\
DiffCSP & 0.44 & 0.002 & 0.60 & 0.73 & 0.13 & 0.24 & 14.42 & \textbf{3.29} & \underline{2.60} \\
DiffCSP++ & \underline{0.42} & 0.003 & \textbf{0.09} & 0.72 & \textbf{0.51} & 0.27 & 20.84 & 3.56 & 2.77 \\
LLaMat2 & \textbf{0.41} & 0.003 & 0.66 & 0.71 & 0.25 & 0.24 & 9.86 & 3.94 & 2.86 \\
LLaMat3 & 0.45 & 0.012 & 9.31 & 0.70 & 0.10 & 0.30 & 13.15 & 3.83 & 2.81 \\
SymmCD & \textbf{0.41} & \underline{0.001} & 3.34 & 0.73 & 0.49 & 0.27 & 19.45 & 3.54 & 2.74 \\
\midrule
Alexandria & 0.38 & 0.004 & 0.97 & 0.71 & 0.50 & 0.25 & 13.50 & 4.02 & 3.33 \\
OQMD & 0.43 & 0.014 & 0.55 & 0.63 & 0.48 & 0.24 & 12.57 & 4.22 & 3.29 \\
\midrule
AFLOW & 0.44 & 0.003 & 0.55 & 0.69 & 0.46 & 0.25 & 10.07 & 4.12 & 3.22 \\
MP-Exp & 0.49 & 0.011 & 0.16 & 0.72 & 0.71 & 0.28 & 54.79 & 2.78 & 2.27 \\
\midrule
RandomEnum & 0.47 & 0.016 & 0.83 & 0.62 & 0.23 & 0.27 & 15.48 & 3.12 & 3.09 \\
IonSub & 0.46 & 0.009 & 0.22 & 0.67 & 0.36 & 0.26 & 10.82 & 4.15 & 3.86 \\
\bottomrule
\end{tabular}
\end{adjustbox}
\end{table}

\begin{table}[htbp]
\centering
\caption{Model evaluation metrics calculated using \textbf{uma} with \textbf{MP-20} as the reference set \textbf{after relaxation}: Validity, Energy, Stability, and Metastability. All models provide 2500 structures except Crystalformer (1000 structures). Submissions are organized into five categories separated by horizontal lines: (1) models that incorporate relaxation as part of their generation pipeline (MatterGen, PLaID++, WyFormer, WyFormer-DFT), (2) other generative models, (3) samples from real-world datasets that contribute to LeMat-Bulk (Alexandria, OQMD) as baselines, (4) samples from the AFLOW dataset and MP-Exp structures, (5) pre-relaxed samples from non-ML generative baselines. Best values are shown in \textbf{bold}, second-best values are \underline{underlined}. Arrows indicate whether higher ({\textuparrow}) or lower ({\textdownarrow}) values are better.}
\scriptsize
\label{tab:table_15}
\begin{adjustbox}{max width=\textwidth}
\begin{tabular}{l cccc c c ccc ccc ccc}
\toprule
\textbf{Model} &
\multicolumn{4}{c}{\textbf{Validity}} &
\textbf{Unique\% {\textuparrow}} & \textbf{Novel\% {\textuparrow}} &
\multicolumn{3}{c}{\textbf{Energy-based}} &
\multicolumn{3}{c}{\textbf{Stability}} &
\multicolumn{3}{c}{\textbf{Metastability}} \\
\cmidrule(lr){2-5} \cmidrule(lr){8-10} \cmidrule(lr){11-13} \cmidrule(lr){14-16}
& \textbf{Valid\% {\textuparrow}} & \textbf{CN\% {\textuparrow}} & \textbf{MinDist\% {\textuparrow}} & \textbf{PhysPlau\% {\textuparrow}}
&  &  & \textbf{FormE (Std) {\textdownarrow}} & \textbf{$E_{\mathrm{hull}}$ (Std) {\textdownarrow}} & \textbf{RMSD (Std) {\textdownarrow}}
& \textbf{Stable\% {\textuparrow}} & \textbf{U-Stable\% {\textuparrow}} & \textbf{SUN\% {\textuparrow}}
& \textbf{Metastable\% {\textuparrow}} & \textbf{U-Meta\% {\textuparrow}} & \textbf{MSUN\% {\textuparrow}} \\
\midrule
MatterGen & 95.8 & 95.8 & \textbf{100.0} & \textbf{100.0} & \textbf{95.3} & \underline{83.0} & \meanstd{-0.74}{0.80} & \meanstd{0.12}{0.14} & $\bm{\meanstd{0.00}{0.05}}$ & 10.2 {\scriptsize (255)} & 10.1 {\scriptsize (252)} & 5.8 {\scriptsize (145)} & 40.2 {\scriptsize (1006)} & \underline{39.8} {\scriptsize (995)} & \textbf{31.9} {\scriptsize (797)} \\
PLaID++ & 96.0 & 96.0 & 96.2 & 96.2 & 77.7 & 63.8 & \meanstd{-0.49}{0.47} & \underline{\meanstd{0.06}{0.16}} & $\bm{\meanstd{0.00}{0.01}}$ & \textbf{34.3} {\scriptsize (857)} & \underline{23.4} {\scriptsize (586)} & \underline{12.8} {\scriptsize (320)} & \underline{42.3} {\scriptsize (1057)} & 35.4 {\scriptsize (886)} & 23.4 {\scriptsize (585)} \\
WyFormer & 95.0 & 95.1 & 99.8 & \underline{99.9} & 94.6 & 81.7 & \meanstd{-0.67}{0.90} & \meanstd{0.22}{0.32} & $\bm{\meanstd{0.00}{0.02}}$ & 9.6 {\scriptsize (239)} & 9.3 {\scriptsize (233)} & 4.8 {\scriptsize (120)} & 26.6 {\scriptsize (664)} & 26.3 {\scriptsize (658)} & 18.1 {\scriptsize (452)} \\
WyFormer-DFT & 95.2 & 95.2 & \textbf{100.0} & \textbf{100.0} & 94.9 & \textbf{83.1} & \meanstd{-0.71}{0.89} & \meanstd{0.20}{0.28} & \meanstd{0.04}{0.21} & 9.4 {\scriptsize (236)} & 9.3 {\scriptsize (232)} & 4.7 {\scriptsize (118)} & 26.6 {\scriptsize (666)} & 26.6 {\scriptsize (664)} & 19.4 {\scriptsize (486)} \\
\midrule
ADiT & \textbf{98.9} & \textbf{99.0} & \underline{99.9} & \textbf{100.0} & 95.1 & 31.9 & \underline{\meanstd{-1.04}{0.90}} & $\bm{\meanstd{0.04}{0.09}}$ & $\bm{\meanstd{0.00}{0.00}}$ & \underline{33.9} {\scriptsize (848)} & \textbf{32.2} {\scriptsize (805)} & 5.6 {\scriptsize (140)} & \textbf{50.0} {\scriptsize (1251)} & \textbf{48.3} {\scriptsize (1208)} & 13.4 {\scriptsize (335)} \\
Crystal-GFN & 55.3 & 57.4 & 98.0 & 95.7 & 55.3 & 55.3 & $\bm{\meanstd{-2.54}{2.08}}$ & \meanstd{0.41}{1.90} & \meanstd{0.72}{0.75} & 15.6 {\scriptsize (389)} & 15.6 {\scriptsize (389)} & \textbf{15.6} {\scriptsize (389)} & 3.4 {\scriptsize (84)} & 3.4 {\scriptsize (84)} & 3.4 {\scriptsize (84)} \\
Crystalformer & 92.0 & 93.8 & 95.6 & 97.2 & 91.5 & 64.2 & \meanstd{-0.74}{0.85} & \meanstd{0.15}{0.25} & $\bm{\meanstd{0.00}{0.02}}$ & 17.9 {\scriptsize (179)} & 17.6 {\scriptsize (176)} & 6.9 {\scriptsize (69)} & 34.7 {\scriptsize (347)} & 34.6 {\scriptsize (346)} & 18.3 {\scriptsize (183)} \\
DiffCSP & 96.2 & 96.2 & \textbf{100.0} & \textbf{100.0} & 94.8 & 81.8 & \meanstd{-0.76}{0.82} & \meanstd{0.12}{0.18} & \meanstd{0.07}{0.28} & 13.4 {\scriptsize (336)} & 12.8 {\scriptsize (321)} & 7.1 {\scriptsize (178)} & 38.6 {\scriptsize (965)} & 38.0 {\scriptsize (951)} & \underline{30.4 {\scriptsize (761)}} \\
DiffCSP++ & 95.5 & 95.5 & \textbf{100.0} & \textbf{100.0} & \underline{95.2} & 80.8 & \meanstd{-0.72}{0.86} & \meanstd{0.18}{0.23} & \meanstd{0.04}{0.21} & 12.2 {\scriptsize (304)} & 12.1 {\scriptsize (302)} & 6.8 {\scriptsize (171)} & 29.1 {\scriptsize (728)} & 29.0 {\scriptsize (725)} & 19.9 {\scriptsize (498)} \\
LLaMat2 & \underline{97.0} & \underline{97.0} & \underline{99.9} & \textbf{100.0} & 93.8 & 63.2 & \meanstd{-0.77}{0.85} & \meanstd{0.11}{0.21} & \underline{\meanstd{0.03}{0.20}} & 23.9 {\scriptsize (597)} & 22.6 {\scriptsize (566)} & 8.5 {\scriptsize (213)} & 39.2 {\scriptsize (980)} & 37.6 {\scriptsize (939)} & 20.4 {\scriptsize (510)} \\
LLaMat3 & 69.8 & 70.5 & 82.2 & 83.3 & 69.6 & 68.0 & \meanstd{-0.38}{0.71} & \meanstd{0.24}{0.27} & \meanstd{0.17}{0.46} & 2.5 {\scriptsize (63)} & 2.5 {\scriptsize (62)} & 1.7 {\scriptsize (43)} & 16.0 {\scriptsize (399)} & 15.9 {\scriptsize (398)} & 15.0 {\scriptsize (374)} \\
SymmCD & 95.6 & 95.9 & 99.7 & \textbf{100.0} & 95.1 & 79.8 & \meanstd{-0.67}{0.85} & \meanstd{0.21}{0.27} & $\bm{\meanstd{0.00}{0.03}}$ & 12.5 {\scriptsize (313)} & 12.3 {\scriptsize (308)} & 6.0 {\scriptsize (151)} & 27.4 {\scriptsize (686)} & 27.3 {\scriptsize (683)} & 18.4 {\scriptsize (461)} \\
\midrule
Alexandria & 93.4 & 93.4 & 99.0 & 99.0 & 93.4 & 92.4 & \meanstd{-0.18}{0.95} & \meanstd{0.40}{0.61} & \meanstd{0.00}{0.02} & 10.5 {\scriptsize (263)} & 10.5 {\scriptsize (263)} & 10.3 {\scriptsize (258)} & 24.8 {\scriptsize (621)} & 24.8 {\scriptsize (621)} & 24.2 {\scriptsize (604)} \\
OQMD & 96.8 & 96.8 & 99.5 & 99.5 & 96.4 & 94.3 & \meanstd{-0.22}{0.90} & \meanstd{0.35}{0.47} & \meanstd{0.00}{0.00} & 18.6 {\scriptsize (466)} & 18.5 {\scriptsize (463)} & 17.3 {\scriptsize (432)} & 19.9 {\scriptsize (497)} & 19.9 {\scriptsize (497)} & 18.6 {\scriptsize (466)} \\
\midrule
AFLOW & 91.4 & 91.4 & 100.0 & 100.0 & 87.1 & 56.9 & \meanstd{-0.18}{0.87} & \meanstd{0.32}{0.42} & \meanstd{0.00}{0.00} & 20.0 {\scriptsize (499)} & 18.5 {\scriptsize (462)} & 3.2 {\scriptsize (79)} & 22.0 {\scriptsize (549)} & 19.7 {\scriptsize (492)} & 4.3 {\scriptsize (108)} \\
MP-Exp & 97.0 & 97.1 & 98.9 & 98.9 & 97.0 & 67.2 & \meanstd{-1.12}{1.09} & \meanstd{0.13}{0.49} & \meanstd{0.00}{0.01} & 28.2 {\scriptsize (705)} & 28.2 {\scriptsize (705)} & 17.1 {\scriptsize (427)} & 42.2 {\scriptsize (1055)} & 42.2 {\scriptsize (1054)} & 24.5 {\scriptsize (613)} \\
\midrule
RandomEnum & 98.5 & 98.8 & 99.4 & 99.6 & 98.5 & 98.0 & \meanstd{-0.69}{0.83} & \meanstd{0.13}{0.18} & \meanstd{0.00}{0.04} & 9.8 {\scriptsize (245)} & 9.8 {\scriptsize (245)} & 9.6 {\scriptsize (240)} & 29.4 {\scriptsize (736)} & 29.4 {\scriptsize (736)} & 29.2 {\scriptsize (729)} \\
IonSub & 99.3 & 99.3 & 100.0 & 100.0 & 91.8 & 77.2 & \meanstd{-0.42}{0.43} & \meanstd{0.01}{0.11} & \meanstd{0.00}{0.00} & 49.6 {\scriptsize (1241)} & 46.2 {\scriptsize (1156)} & 37.2 {\scriptsize (930)} & 39.0 {\scriptsize (974)} & 35.5 {\scriptsize (888)} & 25.5 {\scriptsize (638)} \\
\bottomrule
\end{tabular}
\end{adjustbox}
\end{table}

\begin{table}[htbp]
\centering
\caption{Model evaluation metrics calculated using \textbf{uma} with \textbf{MP-20} as the reference set \textbf{after relaxation}: Distribution, Diversity, and HHI. All models provide 2500 structures except Crystalformer (1000 structures). Submissions are organized into five categories separated by horizontal lines: (1) models that incorporate relaxation as part of their generation pipeline (MatterGen, PLaID++, WyFormer, WyFormer-DFT), (2) other generative models, (3) samples from real-world datasets that contribute to LeMat-Bulk (Alexandria, OQMD) as baselines, (4) samples from the AFLOW dataset and MP-Exp structures, (5) pre-relaxed samples from non-ML generative baselines. Best values are shown in \textbf{bold}, second-best values are \underline{underlined}. Arrows indicate whether higher ({\textuparrow}) or lower ({\textdownarrow}) values are better.}
\scriptsize
\label{tab:table_16}
\begin{adjustbox}{max width=\textwidth}
\begin{tabular}{l ccc cccc cc}
\toprule
\textbf{Model} &
\multicolumn{3}{c}{\textbf{Distribution}} &
\multicolumn{4}{c}{\textbf{Diversity}} &
\multicolumn{2}{c}{\textbf{HHI}} \\
\cmidrule(lr){2-4} \cmidrule(lr){5-8} \cmidrule(lr){9-10}
& \textbf{JS {\textdownarrow}} & \textbf{MMD {\textdownarrow}} & \textbf{FID {\textdownarrow}}
& \textbf{ElemDiv {\textuparrow}} & \textbf{SGDiv {\textuparrow}} & \textbf{SizeDiv {\textuparrow}} & \textbf{SiteDiv {\textuparrow}}
& \textbf{Prod {\textdownarrow}} & \textbf{Res {\textdownarrow}} \\
\midrule
MatterGen & \underline{0.43} & \textbf{0.000} & 0.10 & 0.64 & 0.30 & 0.24 & 12.80 & 3.51 & 2.72 \\
PLaID++ & 0.48 & 0.049 & 0.31 & 0.68 & 0.28 & 0.21 & 6.01 & 5.23 & 3.36 \\
WyFormer & \textbf{0.42} & 0.002 & 0.35 & \textbf{0.74} & \textbf{0.50} & 0.27 & 24.27 & 3.52 & 2.71 \\
WyFormer-DFT & \textbf{0.42} & 0.002 & 0.30 & \underline{0.73} & \underline{0.49} & 0.27 & 23.57 & 3.49 & 2.74 \\
\midrule
ADiT & \underline{0.43} & \textbf{0.000} & \textbf{0.02} & 0.71 & 0.43 & 0.24 & 14.51 & \underline{3.41} & 2.70 \\
Crystal-GFN & 0.56 & 0.066 & 1.32 & \textbf{0.74} & 0.12 & \textbf{0.32} & \textbf{31.50} & 3.45 & \textbf{2.33} \\
Crystalformer & \textbf{0.42} & 0.002 & 0.13 & 0.71 & 0.34 & \underline{0.31} & 21.62 & 3.54 & 2.68 \\
DiffCSP & 0.44 & 0.003 & 0.26 & \underline{0.73} & 0.34 & 0.25 & 14.42 & \textbf{3.27} & \underline{2.59} \\
DiffCSP++ & \textbf{0.42} & \underline{0.001} & 0.09 & 0.72 & \textbf{0.50} & 0.27 & 20.90 & 3.55 & 2.77 \\
LLaMat2 & \textbf{0.42} & \underline{0.001} & \underline{0.05} & 0.71 & 0.38 & 0.24 & 11.91 & 3.84 & 2.79 \\
LLaMat3 & 0.44 & 0.007 & 0.68 & 0.72 & 0.17 & 0.27 & \underline{28.26} & 3.48 & 2.93 \\
SymmCD & \textbf{0.42} & 0.002 & 0.27 & \underline{0.73} & 0.48 & 0.27 & 20.84 & 3.51 & 2.68 \\
\midrule
Alexandria & 0.38 & 0.004 & 0.91 & 0.71 & 0.50 & 0.26 & 13.50 & 4.02 & 3.33 \\
OQMD & 0.43 & 0.012 & 0.51 & 0.63 & 0.46 & 0.24 & 12.57 & 4.22 & 3.29 \\
\midrule
AFLOW & 0.44 & 0.003 & 0.53 & 0.69 & 0.50 & 0.25 & 10.07 & 4.12 & 3.22 \\
MP-Exp & 0.49 & 0.013 & 0.12 & 0.72 & 0.68 & 0.28 & 54.79 & 2.78 & 2.27 \\
\midrule
RandomEnum & 0.47 & 0.018 & 0.60 & 0.62 & 0.34 & 0.27 & 15.48 & 3.13 & 3.10 \\
IonSub & 0.47 & 0.010 & 0.22 & 0.67 & 0.41 & 0.26 & 10.81 & 4.15 & 3.86 \\
\bottomrule
\end{tabular}
\end{adjustbox}
\end{table}

\begin{figure}[htbp]
    \centering
    \includegraphics[width=\textwidth]{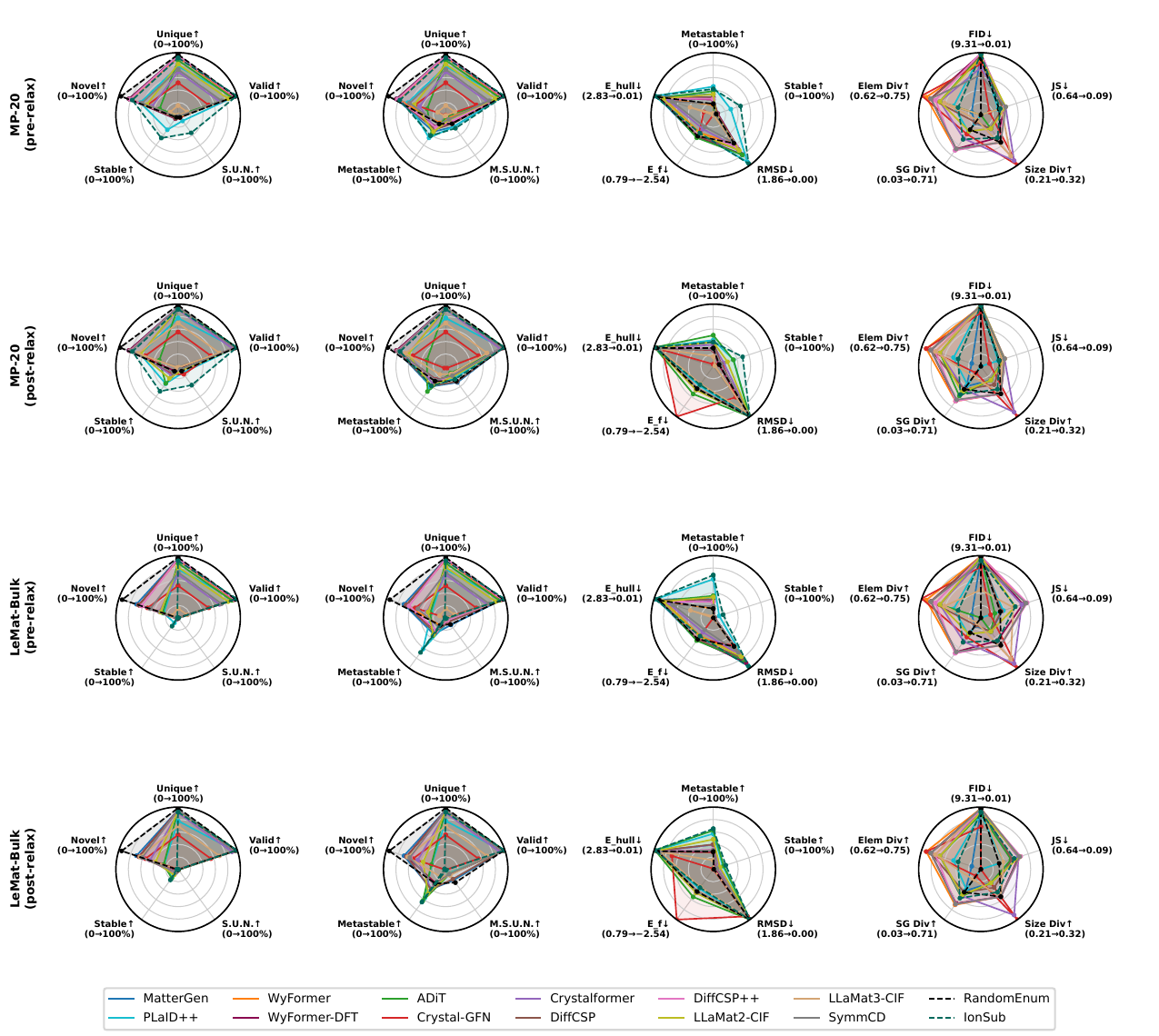}
    \caption{Generative models and non-ML baselines (RandomEnum, IonSub) evaluated across all metric categories with respect to UMA, both before and after relaxation. Rows from top to bottom: MP-20 pre-relaxation, MP-20 post-relaxation, LeMat-Bulk pre-relaxation, LeMat-Bulk post-relaxation. Columns from left to right: SUN Metrics, MSUN Metrics, Energy \& Stability, Distribution \& Diversity. All rows use consistent scaling for direct comparison.}
    \label{fig:appendix_generative_comparison}
\end{figure}

\begin{figure}[htbp]
    \centering
    \includegraphics[width=\textwidth]{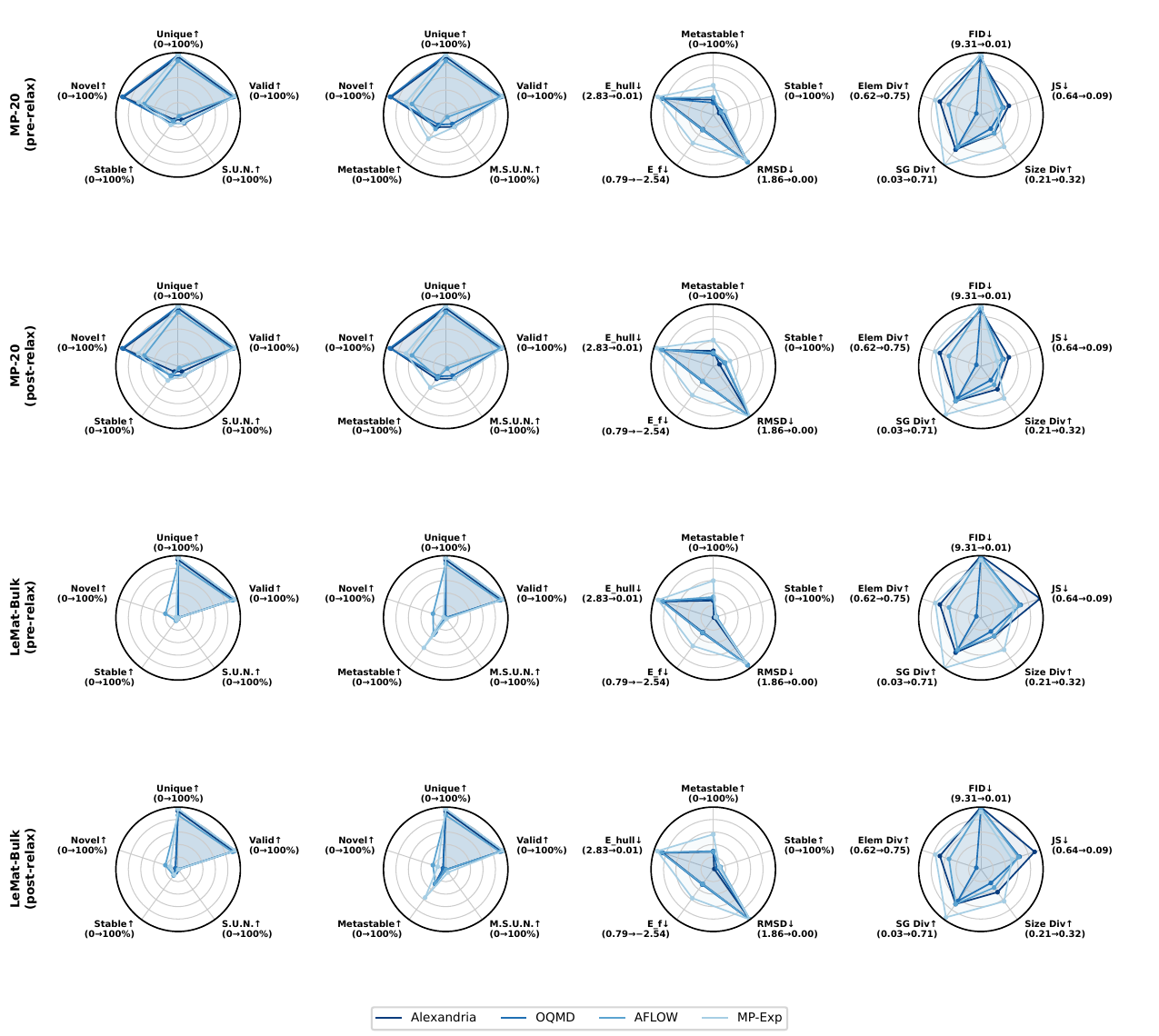}
    \caption{Dataset baselines (Alexandria, OQMD, AFLOW, and MP-Exp) evaluated across all metric categories with respect to UMA, both before and after relaxation. Rows from top to bottom: MP-20 pre-relaxation, MP-20 post-relaxation, LeMat-Bulk pre-relaxation, LeMat-Bulk post-relaxation. Columns from left to right: SUN Metrics, MSUN Metrics, Energy \& Stability, Distribution \& Diversity. All rows use consistent scaling for direct comparison.}
    \label{fig:appendix_datasets_comparison}
\end{figure}

\FloatBarrier
\section{Environmental and Sustainability Considerations}
\label{app:sec:sustainability}

The application of generative models to materials discovery presents significant opportunities for advancing environmental sustainability goals. As global challenges related to climate change, resource depletion, and environmental degradation intensify, the need for novel materials with reduced environmental footprints becomes increasingly urgent. Generative approaches can accelerate the discovery of sustainable alternatives by explicitly incorporating environmental criteria into the design process.

One promising direction involves the targeted generation of materials with reduced reliance on critical or environmentally problematic elements. By conditioning generative models on compositional constraints that exclude toxic, rare, or environmentally harmful elements, researchers can guide exploration toward more sustainable regions of chemical space. Similarly, models can be trained to prioritize earth-abundant elements and avoid those associated with problematic extraction practices or geopolitical supply risks.

Energy-related applications represent another frontier where generative models could significantly impact sustainability outcomes. The discovery of more efficient catalysts for renewable energy production, improved battery materials for energy storage, and novel photovoltaic materials could accelerate the transition away from fossil fuels. By specifically targeting properties relevant to these applications, generative models can focus computational and experimental resources on high-impact sustainability domains.

Life-cycle considerations present a more complex but equally important target for integration with generative approaches. Ideally, materials should be designed not only for performance but also for recyclability, biodegradability, or other end-of-life scenarios that minimize environmental impact. Incorporating such considerations into generative frameworks remains challenging due to the complex, multi-faceted nature of life-cycle assessment, but it represents a crucial direction for future research.

The computational efficiency of generative processes themselves also warrants consideration from a sustainability perspective. As models grow in complexity and scale, their energy consumption and carbon footprint increase accordingly. Developing more efficient architectures, training procedures, and sampling approaches could reduce the environmental impact of the discovery process itself, aligning computational means with environmental ends. This consideration becomes particularly important as generative approaches scale to industrial applications and high-throughput discovery platforms.

The ultimate success of generative approaches in advancing sustainability will depend not only on technical capabilities but also on intentional alignment with environmental objectives. By explicitly incorporating sustainability metrics into reward functions, objective functions, and evaluation criteria, the materials community can ensure that generative models contribute to addressing environmental challenges rather than merely accelerating traditional discovery paradigms without regard for sustainability implications.

\end{document}